\documentclass[runningheads]{llncs}

\usepackage{eccv}

\usepackage{float}
\usepackage{eccvabbrv}
\usepackage{caption}
\usepackage{graphicx}
\usepackage{booktabs}
\usepackage[accsupp]{axessibility}  
\usepackage{url}
\usepackage{threeparttable}
\usepackage{epsfig}
\usepackage{comment}
\usepackage{graphicx}
\usepackage{arydshln}
\usepackage{pifont}
\usepackage{color,colortbl}
\definecolor{Gray}{gray}{0.8}
\usepackage{multicol}
\usepackage{dcolumn}
\usepackage{amsmath, amssymb, caption, subcaption, multirow, overpic, textpos}
\usepackage{textcomp}
\usepackage{upgreek}
\usepackage{bm}
\usepackage{cases}

\usepackage{mathtools}
\usepackage{array}
\usepackage{bbm}
\usepackage{etoc}  
\usepackage{wrapfig}

\usepackage{rotating}
\usepackage{tabularx}
\usepackage{tabulary}
\usepackage{adjustbox}

\usepackage{enumitem}

\usepackage{algorithm}
\usepackage[noend]{algpseudocode}
\usepackage{tikz}
\usepackage{makecell}

\definecolor{bluebell}{rgb}{0.64, 0.64, 0.82}
\definecolor{cadet}{rgb}{0.33, 0.41, 0.47}
\definecolor{mydarkgreen}{RGB}{0, 150, 0}

\newcommand{\greencmark}{\color{mydarkgreen}{\ding{51}}}
\newcommand{\redxmark}{\color{gray!35}{\ding{55}}}
\usepackage[colorlinks]{hyperref}
\usepackage{orcidlink}
\begin{document}

\newcommand{\pc}{\textcolor{red}}

\newcommand{\judy}{\textcolor{teal}}

\newcommand{\ks}{\textcolor{black}}

\newcommand{\inc}{\textcolor{blue}}

\newcommand{\dec}{\textcolor{red}}

\newcommand*\rot{\multicolumn{1}{R{90}{1em}}}%
\newcommand*\rotb{\multicolumn{1}{R{45}{1em}}}%
\newcommand{\valid}{\textsc{VALID}\xspace}
\newcommand{\synth}{\textsc{SynthAer}\xspace}
\newcommand{\csim}{\textsc{SkyScenes}\xspace}
\newcommand{\crl}{\textsc{Carla}\xspace}
\newcommand{\syn}{\textsc{SynDrone}\xspace}
\newcommand{\icg}{\textsc{ICG Drone}\xspace}
\newcommand{\ua}{\textsc{UAV}\textnormal{id}\xspace}
\newcommand{\aero}{\textsc{AeroScapes}\xspace}
\newcommand{\pst}{\textsc{Pasta}\xspace}
\newcommand{\dl}{\textsc{DeepLabV$2$}\xspace}
\newcommand{\lightgray}[1]{\textcolor{gray!60}{#1}}

\newcommand{\da}{\textsc{DAFormer}\xspace}
\newcommand{\pnav}{\textsc{PointNav}\xspace}
\newcommand{\onav}{\textsc{ObjectNav}\xspace}
\newcommand{\rnav}{\textsc{RobustNav}\xspace}
\newcommand{\rthor}{\textsc{RoboTHOR}\xspace}
\newcommand{\thor}{\textsc{AI2-THOR}\xspace}
\newcommand{\degr}{$^{\circ}$\xspace}
\newcommand{\vcr}{{\texttt{vis}\xspace}}
\newcommand{\dcr}{{\texttt{dyn}\xspace}}
\newcommand{\vdcr}{{\texttt{vis}+\texttt{dyn}\xspace}}

\newcommand{\str}{sim-to-real\xspace}

\newcommand{\ttbf}[1]{\texttt{\textbf{#1}}}
\newcommand{\nocaps}[0]{\ttbf{nocaps}\xspace}

\newcommand{\mv}{\texttt{\textbf{move\_ahead}}\xspace}
\newcommand{\rlf}{\texttt{\textbf{rotate\_left}}\xspace}
\newcommand{\rrr}{\texttt{\textbf{rotate\_right}}\xspace}
\newcommand{\lu}{\texttt{\textbf{look\_up}}\xspace}
\newcommand{\ld}{\texttt{\textbf{look\_down}}\xspace}
\newcommand{\ed}{\texttt{\textbf{end}}\xspace}

\newcommand{\drop}[1]{\textcolor{gray}{\textsubscript{$-$#1}}}
\newcommand{\rise}[1]{\textcolor{gray}{\textsubscript{$+$#1}}}
\newcommand{\riseneg}[1]{\textcolor{gray}{\textsubscript{$-$#1}}}

\newcommand{\Drop}[1]{\textcolor{red}{\textsubscript{\bf $-$#1}}}
\newcommand{\Rise}[1]{\textcolor{green}{\textsubscript{\bf $+$#1}}}
\newcommand{\Riseneg}[1]{\textcolor{green}{\textsubscript{\bf $-$#1}}}

\newcommand{\graycell}{\cellcolor{gray!20}}

\makeatletter
\DeclareRobustCommand\onedot{\futurelet\@let@token\@onedot}
\def\@onedot{\ifx\@let@token.\else.\null\fi\xspace}

\def\eg{\emph{e.g}\onedot} \def\Eg{\emph{E.g}\onedot}
\def\ie{\emph{i.e}\onedot} \def\Ie{\emph{I.e}\onedot}
\def\st{s.t\onedot} \def\St{\emph{S.t}\onedot}
\def\cf{\emph{c.f}\onedot} \def\Cf{\emph{C.f}\onedot}
\def\etc{\emph{etc}\onedot} \def\vs{\emph{vs}\onedot}
\def\wrt{w.r.t\onedot} \def\dof{d.o.f\onedot}
\makeatother

\newcommand{\eqnref}[1]{(\ref{#1})}
\newcommand{\tableref}[1]{Table \ref{#1}}

\newcommand{\myvec}[1]{\mathbf{#1}}
\newcommand{\myvecsym}[1]{\boldsymbol{#1}}
\newcommand{\mytensor}[1]{\mathbf{\tilde{#1}}}

\newcommand{\eigmax}{\lambda_{\mathrm{max}}}
\newcommand{\eigmin}{\lambda_{\mathrm{min}}}

\newcommand{\valpha}{\myvecsym{\alpha}}
\newcommand{\vbeta}{\myvecsym{\beta}}
\newcommand{\vchi}{\myvecsym{\chi}}
\newcommand{\vdelta}{\myvecsym{\delta}}
\newcommand{\vDelta}{\myvecsym{\Delta}}
\newcommand{\vepsilon}{\myvecsym{\epsilon}}
\newcommand{\vzeta}{\myvecsym{\zeta}}
\newcommand{\vXi}{\myvecsym{\Xi}}
\newcommand{\vell}{\myvecsym{\ell}}
\newcommand{\veta}{\myvecsym{\eta}}
\newcommand{\vgamma}{\myvecsym{\gamma}}
\newcommand{\vGamma}{\myvecsym{\Gamma}}
\newcommand{\vmut}{\myvecsym{\tilde{\mu}}}
\newcommand{\vnu}{\myvecsym{\nu}}
\newcommand{\vkappa}{\myvecsym{\kappa}}
\newcommand{\vlambda}{\myvecsym{\lambda}}
\newcommand{\vLambda}{\myvecsym{\Lambda}}
\newcommand{\vLambdaBar}{\overline{\vLambda}}
\newcommand{\vomega}{\myvecsym{\omega}}
\newcommand{\vOmega}{\myvecsym{\Omega}}
\newcommand{\vphi}{\myvecsym{\phi}}
\newcommand{\vPhi}{\myvecsym{\Phi}}
\newcommand{\vpi}{\myvecsym{\pi}}
\newcommand{\vPi}{\myvecsym{\Pi}}
\newcommand{\vpsi}{\myvecsym{\psi}}
\newcommand{\vPsi}{\myvecsym{\Psi}}
\newcommand{\vrho}{\myvecsym{\rho}}
\newcommand{\vthetat}{\myvecsym{\tilde{\sigma}}}
\newcommand{\vTheta}{\myvecsym{\sigma}}
\newcommand{\vsigma}{\myvecsym{\sigma}}
\newcommand{\vSigma}{\myvecsym{\Sigma}}
\newcommand{\vSigmat}{\myvecsym{\tilde{\Sigma}}}
\newcommand{\vsigmoid}{\vsigma}
\newcommand{\vtau}{\myvecsym{\tau}}
\newcommand{\vxi}{\myvecsym{\xi}}

\newcommand{\vBt}{\myvec{\tilde{B}}}
\newcommand{\vws}{\vw_s}
\newcommand{\vwt}{\myvec{\tilde{w}}}
\newcommand{\vWt}{\myvec{\tilde{W}}}
\newcommand{\vwh}{\hat{\vw}}
\newcommand{\vxt}{\myvec{\tilde{x}}}
\newcommand{\vyt}{\myvec{\tilde{y}}}

\newcommand{\vA}{\myvec{A}}
\newcommand{\vB}{\myvec{B}}
\newcommand{\vC}{\myvec{C}}
\newcommand{\vD}{\myvec{D}}
\newcommand{\vE}{\myvec{E}}
\newcommand{\vF}{\myvec{F}}
\newcommand{\vG}{\myvec{G}}
\newcommand{\vH}{\myvec{H}}
\newcommand{\vI}{\myvec{I}}
\newcommand{\vJ}{\myvec{J}}
\newcommand{\vK}{\myvec{K}}
\newcommand{\vL}{\myvec{L}}
\newcommand{\vM}{\myvec{M}}
\newcommand{\vMt}{\myvec{\tilde{M}}}
\newcommand{\vN}{\myvec{N}}
\newcommand{\vO}{\myvec{O}}
\newcommand{\vP}{\myvec{P}}
\newcommand{\vQ}{\myvec{Q}}
\newcommand{\vR}{\myvec{R}}
\newcommand{\vS}{\myvec{S}}
\newcommand{\vT}{\myvec{T}}
\newcommand{\vU}{\myvec{U}}
\newcommand{\vV}{\myvec{V}}
\newcommand{\vW}{\myvec{W}}
\newcommand{\vX}{\myvec{X}}
\newcommand{\vXs}{\vX_{s}}
\newcommand{\vXt}{\myvec{\tilde{X}}}
\newcommand{\vY}{\myvec{Y}}
\newcommand{\vZ}{\myvec{Z}}
\newcommand{\vZt}{\myvec{\tilde{Z}}}
\newcommand{\vzt}{\myvec{\tilde{z}}}

\newcommand{\bbI}{\mathbb{I}}
\newcommand{\bbL}{\mathbb{L}}
\newcommand{\bbM}{\mathbb{M}}
\newcommand{\bbS}{\mathbb{S}}

\newcommand{\pmvar}[1]{\scriptsize{$\pm$#1}}
\newcommand{\mymathcal}[1]{\mathcal{#1}}

\newcommand{\calA}{\mymathcal{A}}
\newcommand{\calB}{\mymathcal{B}}
\newcommand{\calC}{\mymathcal{C}}
\newcommand{\calD}{{\mymathcal{D}}}
\newcommand{\calDx}{\calD_x}
\newcommand{\calE}{\mymathcal{E}}
\newcommand{\cale}{{\cal e}}
\newcommand{\calF}{\mymathcal{F}}
\newcommand{\calG}{\mymathcal{G}}
\newcommand{\calH}{\mymathcal{H}}
\newcommand{\calHX}{{\calH}_X}
\newcommand{\calHy}{{\calH}_y}
\newcommand{\calI}{\mymathcal{I}}
\newcommand{\calK}{\mymathcal{K}}
\newcommand{\calL}{\mymathcal{L}}
\newcommand{\calM}{\mymathcal{M}}
\newcommand{\calN}{\mymathcal{N}}
\newcommand{\caln}{{\cal n}}
\newcommand{\calNP}{\mymathcal{NP}}
\newcommand{\calO}{\mymathcal{O}}
\newcommand{\calMp}{\calM^+}
\newcommand{\calMm}{\calM^-}
\newcommand{\calMo}{\calM^o}
\newcommand{\Ctest}{C_*}
\newcommand{\calP}{\mymathcal{P}}
\newcommand{\calq}{{\cal q}}
\newcommand{\calQ}{\mymathcal{Q}}
\newcommand{\calR}{\mymathcal{R}}
\newcommand{\calS}{\mymathcal{S}}
\newcommand{\calSstar}{\calS_*}
\newcommand{\calT}{\mymathcal{T}}
\newcommand{\calU}{\mymathcal{U}}
\newcommand{\calV}{\mymathcal{V}}
\newcommand{\calW}{\mymathcal{W}}
\newcommand{\calv}{{\cal v}}
\newcommand{\calX}{\mymathcal{X}}
\newcommand{\calY}{\mymathcal{Y}}
\newcommand{\calZ}{\mymathcal{Z}}

\newcommand{\expectAngle}[1]{\langle #1 \rangle}
\newcommand{\expect}[1]{\mathbb{E}\left[{#1}\right]}
\newcommand{\expectQ}[2]{\mathbb{E}_{{#2}}\left[ {#1} \right]}
\newcommand{\var}[1]{\mathbb{V}\left[ {#1}\right]}
\newcommand{\varQ}[2]{\mathbb{V}_{{#2}}\left[ {#1}\right]}

\newcommand{\entropy}{\mathbb{H}}
\newcommand{\JSpq}[2]{\mathbb{JS}\left({#1}||{#2}\right)}
\newcommand{\KLpq}[2]{\mathbb{KL}\left({#1}||{#2}\right)}
\newcommand{\MI}{\mathbb{I}}
\newcommand{\MIxy}[2]{\mathbb{I}\left({#1};{#2}\right)}
\newcommand{\JS}{\mathrm{JS}}

\newcommand{\const}{\mathrm{const}}

\newcommand{\yobs}{\vy_{\calO}}
\newcommand{\ymiss}{\vy_{\calM}}

\newcommand{\union}{\cup}
\newcommand{\intersect}{\cap}

\newcommand{\eat}[1]{}
\newcommand{\replace}[2]{} %

\setlength\marginparsep{5pt}
\setlength\marginparwidth{\dimexpr1in+\hoffset+\oddsidemargin-\marginparsep*2}

\newcommand{\elbo}{\mathrm{elbo}}
\newcommand{\elboo}[2]{\elbo(#1 ; \; #2)}
\newcommand{\elbooo}[3]{\elbo(#1; \; #2; \; #3)}
\newcommand{\vqabayes}{Prob-NMN}
\newcommand{\data}{\mathcal{D}}
\newcommand{\datatrain}{\data^{\mathrm{train}}}
\newcommand{\datatest}{\data^{\mathrm{test}}}
\newcommand{\datassl}{\data^{\mathrm{ssl}}}
\newcommand{\dataretrieval}{\data^{\mathrm{retrieval}}}
\newcommand{\gauss}{\mathcal{N}}
\newcommand{\proj}{\mathrm{proj}}
\newcommand{\ssim}{\mathrm{sim}}
\newcommand{\sort}{\mathrm{rank}}
\newcommand{\mean}{\mathrm{mean}}
\newcommand{\dist}{\mathrm{dist}}
\newcommand{\correctness}{correctness\xspace}
\newcommand{\Correctness}{Correctness\xspace}
\newcommand{\coverage}{coverage\xspace}
\newcommand{\leaves}{\mathrm{leaves}}
\newcommand{\level}{\mathrm{level}}
\newcommand{\error}{\mathrm{error}}
\newcommand{\obs}{{\mathcal O}}
\newcommand{\missing}{{\mathcal M}}
\newcommand{\attributes}{{\mathcal A}}
\newcommand{\mask}[2]{#1_{#2}}
\newcommand{\concept}{\mask{\vy}{\obs}}
\newcommand{\extension}{{\mathcal S}}
\newcommand{\ptrue}{p^*}
\newcommand{\pone}{p_1}
\newcommand{\ptwo}{p_2}
\newcommand{\super}[2]{#1^{(#2)}} %

\newcommand{\bivcca}{BiVCCA\xspace}
\newcommand{\Bivcca}{\bivcca}
\newcommand{\jmvae}{JMVAE\xspace}
\newcommand{\JMVAE}{\jmvae}
\newcommand{\telbo}{TELBO\xspace}
\newcommand{\CelebA}{CelebA\xspace}
\newcommand{\CELEBA}{CelebA\xspace}
\newcommand{\MNISTa}{MNIST-A\xspace}
\newcommand{\MNISTA}{\MNISTa}
\newcommand{\mnistaffine}{\MNISTa}
\newcommand{\mnistbit}{MNIST-2bit\xspace}
\newcommand{\comp}{comp\xspace}

\newcommand{\unimodalx}{\lambda_x^x}
\newcommand{\bimodalx}{\lambda_x^{xy}}
\newcommand{\unimodaly}{\lambda_y^y}
\newcommand{\bimodaly}{\lambda_y^{yx}}

\newcommand{\pp}{p_{\vtheta}}
\newcommand{\px}{p_{\vtheta_x}}
\newcommand{\py}{p_{\vtheta_y}}
\newcommand{\qq}{q_{\vphi}}
\newcommand{\qx}{q_{\vphi_x}}
\newcommand{\qy}{q_{\vphi_y}}
\newcommand{\qqavg}{q_{\vphi}^{\mathrm{avg}}}

\newenvironment{packed_itemize}{
	\begin{list}{\labelitemi}{\leftmargin=1em}
		\setlength{\itemsep}{0pt}
		\setlength{\parskip}{0pt}
		\setlength{\parsep}{0pt}
	}{\end{list}}

\newenvironment{packed_enumerate}{
	\begin{list}{\labelenumi}{\leftmargin=1em}
		\setlength{\itemsep}{0pt}
		\setlength{\parskip}{0pt}
		\setlength{\parsep}{0pt}
	}{\end{list}}


\title{\textsc{SkyScenes}: A Synthetic Dataset for Aerial Scene Understanding} 
\author{
  Sahil Khose\thanks{Equal Contribution \\ The 18th European Conference on Computer Vision ECCV 2024}\orcidlink{0000-0003-2194-3115} \qquad Anisha Pal$^*$\orcidlink{0000-0002-7850-5936} \qquad Aayushi Agarwal$^*$\orcidlink{0009-0009-3933-3893} \qquad Deepanshi$^*$\orcidlink{0000-0001-5210-0722}\\
  \qquad Judy Hoffman\orcidlink{0000-0003-1971-1606} \qquad Prithvijit Chattopadhyay\orcidlink{0000-0001-7555-3644}\\[0.02in]}
\institute{Georgia Institute of Technology
\email{\{sahil.khose,apal72,judy,prithvijit3\}@gatech.edu} \email{\{aayushi.agarwal007,deepanshi.asr.21\}@gmail.com}
\url{https://huggingface.co/datasets/hoffman-lab/SkyScenes}}

\titlerunning{\textsc{SkyScenes}}
\authorrunning{S.Khose et al.}

\maketitle
\begin{figure}
    \centering
\includegraphics[width=\textwidth]{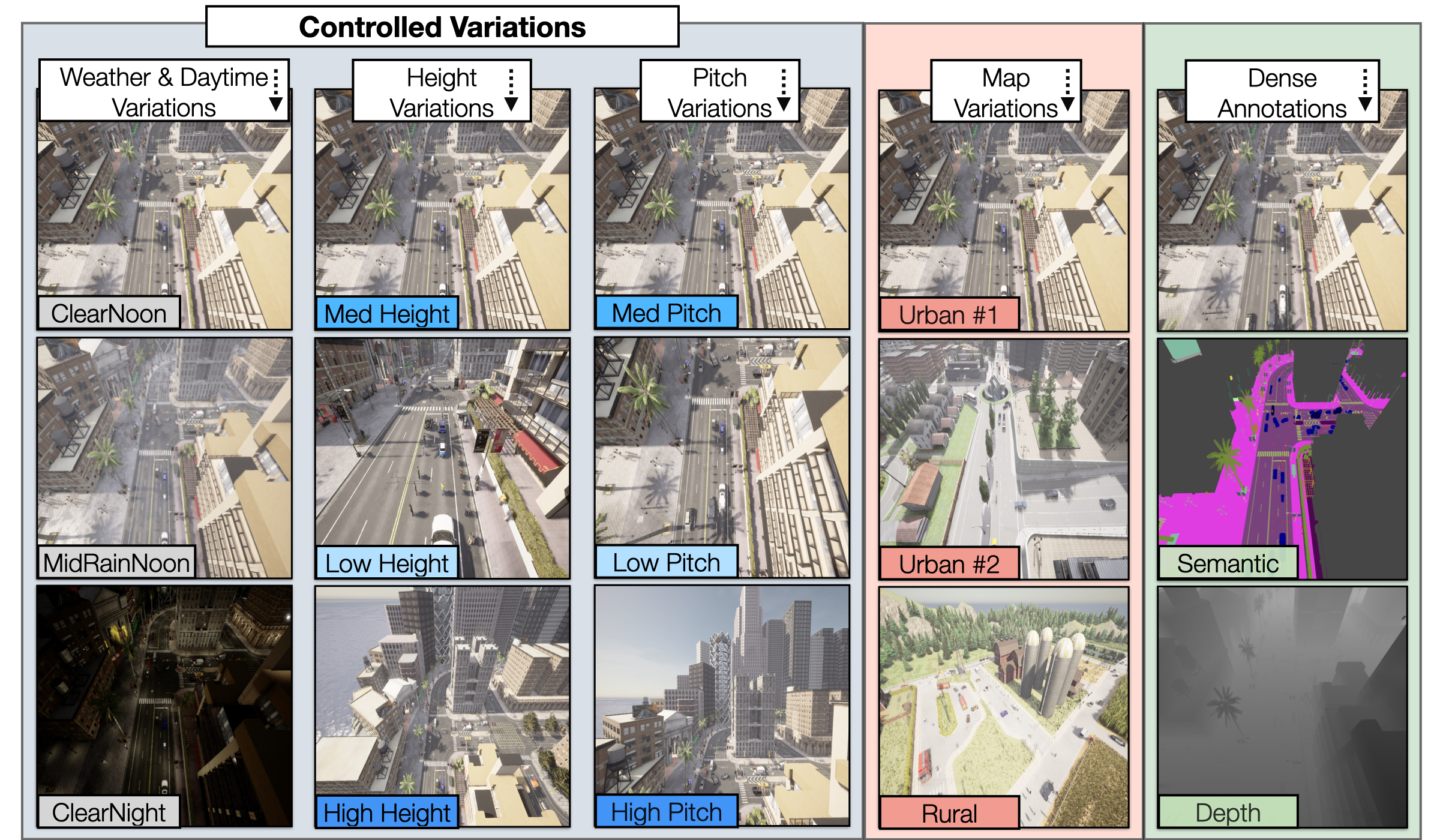}
\captionof{figure}{
\csim comprises of $33.6$k aerial images curated from aerial oblique viewpoints with \textit{\textbf{controlled variations}} facilitating reproducibility of viewpoints across different weather and daytime conditions (col $1$), different flying altitudes (col $2$) and different viewpoint pitch angles (col $3$), across different map layouts (rural and urban, col $4$) with dense pixel-level semantic, instance and depth annotations (col $5$). 
}

\label{fig:teaser}
\end{figure}

\def\paperID{10113} %

\begin{abstract}
\par\noindent
Real-world aerial scene understanding is limited by a lack of datasets that contain densely annotated images curated under a diverse set of conditions. Due to inherent challenges in obtaining such images in controlled real-world settings, we present \csim, a synthetic dataset of densely annotated aerial images captured from Unmanned Aerial Vehicle (UAV) perspectives. We carefully curate \csim images from \crl to comprehensively capture diversity across layouts (urban and rural maps), weather conditions, times of day, pitch angles and altitudes with corresponding semantic, instance and depth annotations.
Through our experiments using \csim, we show that (1) models trained on \csim generalize well to different real-world scenarios, (2) augmenting training on real images with \csim data can improve real-world performance, (3) controlled variations in \csim can offer insights into how models respond to changes in viewpoint conditions (height and pitch), weather and time of day, and (4) incorporating additional sensor modalities (depth) can improve aerial scene understanding. Our dataset and associated generation code are publicly available at: \href{https://hoffman-group.github.io/SkyScenes/}{https://hoffman-group.github.io/SkyScenes/}


\end{abstract}

\section{Introduction}
\label{sec:intro}
Aerial imagery provides a unique perspective that is invaluable for a wide range of applications, including surveillance~\cite{nguyen2022state,Prokaj_2014_CVPR}, mapping~\cite{otal2024harnessing, CHAUHAN2024197}, urban planning~\cite{LIU201542, 1335266}, environmental monitoring~\cite{9151141, ijgi11020100}, and disaster response~\cite{su13147547, KEDYS2024129}. These applications rely on accurate and detailed analysis of aerial images to make informed decisions and effectively address various challenges. Naturally, training effective aerial scene-understanding models requires access to large-scale annotated exemplar data that have been \textit{carefully curated} under diverse conditions. Capturing such images not only allows training models that can be robust to anticipated test-time variations but also allows assessing model susceptibility to changing conditions.  
However, carefully curating and annotating such images in the real-world can be prohibitively expensive due to various reasons. First, densely annotating every pixel of high-resolution real-world aerial images is expensive -- for instance, densely annotating a single $4$K image in UAVid~\cite{lyu2020uavid} can take up to $2$ hours! Second, although diversity in training data is vital for developing robust generalization algorithms and for model sensitivity assessment, expanding the real set to include widespread variations (weather, time of day, pitch, altitude) would be uncontrolled (\textit{i.e.,} we can't guarantee the same viewpoint under different conditions as the real world is not static), and hence would require re-annotating newly captured frames. Synthetic data curated from simulators can help counter both these issues as (1) labels are automatic and cheap to obtain and (2) it is possible to recreate same viewpoint (with same scene layout and actor instances, \texttt{vehicles}, \texttt{humans}, etc. in the scene) under differing conditions.\\ 

\par\noindent
The unique challenges introduced by outdoor aerial imagery setting such as variability in altitude and angle of image capture, skewed representation for classes with smaller object sizes (\texttt{humans}, \texttt{vehicles}), size and occlusion variations of object classes in the same image, etc., make it a relatively difficult problem compared to ground-view imagery setting. This is illustrated by the observation that the current developments in syn-to-real generalization methods ~\cite{wei2024stronger,hoyer2022daformer} have resulted in a significant reduction in the syn-to-real generalization gap in ground-view settings to as minimal as 20\%\footnote{metric of choice: mIoU.}, compared to the comparatively higher margin observed in aerial imagery scenarios, reaching up to 50\% (see Sec.~\ref{sec:syn2real}, ~\ref{sec:syn2real add}, ~\ref{sec:real+syn} in appendix).
Unlike synthetic ground plane view datasets (especially for autonomous driving~\cite{sankaranarayanan2018GTA,cordts2016cityscapes,yu2020bdd100k,mapillary,ros2016synthia}), synthetic datasets for aerial imagery (see Table.~\ref{table:syn_dataset_comparison}, rows $4$-$10$) have received relatively less attention \cite{midAir,valid, espada,wang2020tartanairdatasetpushlimits,UrbanScene3D,synthaer,rizzoli2023syndrone}. While existing synthetic aerial imagery datasets have tried to close this gap, they are often found lacking in a few different aspects -- complementary metadata to reproduce existing frame viewpoints under different conditions, limited diversity, availability of dense annotations for a wide vocabulary of classes and image capture (height, pitch) conditions (see Table.~\ref{table:syn_dataset_comparison} for an exhaustive summary). These synthetic aerial datasets rarely allow for reproducing the exact viewpoint with different variations based on detailed scene metadata (see Table.\ref{table:syn_dataset_comparison} Controlled Variations column), a key aspect for evaluating deep learning models' responses to changing conditions and assessing sensitivity to single-variable alterations (e.g., weather, time of day, sensor angle).

\par\noindent
We cover all these aspects by introducing \csim, a synthetic dataset containing $33.6$k densely annotated aerial scenes, leveraging the \crl~\cite{dosovitskiy2017carla} simulator to capture diverse \textit{layout} (urban and rural), \textit{weather}, \textit{daytime}, \textit{pitch} and \textit{altitude} conditions. Our pipeline meticulously generates scenes with precise actor locations and orientations, ensuring each scene's reproducibility with a comprehensive metadata store and rigorous consistency checks. 
\par\noindent
\csim encompasses detailed semantic, instance segmentation ($28$ classes), and depth annotations across $8$ distinct map layouts across $5$ different weather and daytime conditions each over a  combination of $3$ altitude and $4$ pitch variations (see Fig.~\ref{fig:teaser} for examples). 
While doing so, we keep several important desiderata in mind. First, we ensure that stored snapshots are not correlated, to promote diverse viewpoints within a town and facilitate model training. Second, we store all metadata associated with the position of actors, camera, and other scene elements to be able to reproduce the same viewpoints under different weather and daytime conditions. Thirdly, we ensure that the generated data mimics real-world imperfections by introducing variations in sensor locations, such as adding jitter to specified height and pitch values.\footnote{Moreover, through rigorous validations, we ensure this process is consistent and yields error-free re-generations. See Sec.~\ref{sec:desiderata}.} 
Finally, since \crl~\cite{dosovitskiy2017carla} by default does not spawn a lot of pedestrians in a scene, we propose an algorithm to ensure adequate representation of \texttt{humans} in the scene while curating images (see Sec.~\ref{sec:desiderata} and Sec.~\ref{sec:dset_algo} for a detailed discussion).

\par\noindent
Our experiments across $3$ different real datasets and $3$ increasingly competitive semantic segmentation architectures consistently demonstrate that \csim outperforms its closest synthetic counterpart dataset, \syn\cite{rizzoli2023syndrone}. However, despite these dataset-level improvements, the syn-to-real generalization gap persists, indicating that algorithmic improvements developed for ground-view imagery fail to translate effectively to aerial imagery. This underscores the urgent need for specialized algorithmic development in this area.

\par\noindent
Empirically, 
we demonstrate the utility of \csim in several different ways.
First, we show that \csim is a valuable pre-training dataset for real-world aerial scene understanding by, (1) demonstrating that models trained on \csim generalize well to multiple real-world datasets and (2) demonstrating that \csim pretraining improves real-world performance in low-shot regimes.
Second, we show that controlled variations in \csim can serve as a diagnostic test-bed to assess model sensitivity to weather, daytime, pitch, altitude, and layout conditions -- by testing \csim trained models in unseen \csim conditions. Finally, we show that \csim can enable developing multi-modal segmentation models with improved aerial-scene understanding capabilities when additional sensors, such as Depth, are available. 
To summarize, we make the following contributions:
\begin{packed_itemize}
    \item We introduce, \csim, a densely-annotated dataset of $33.6$k synthetic aerial images. \csim contains images from different altitude and pitch settings, encompassing different layouts, weather, and daytime conditions with corresponding dense annotations and viewpoint metadata.
    
    \item We demonstrate that \csim pre-trained models generalize well to real-world scenes and that \csim data can effectively augment real-world training data for improved performance. We also bring attention to the point that while the synthetic-to-real gap has considerably narrowed for ground-view datasets, the same algorithms are unable to bridge this gap in aerial imagery. 
    \item We show that our unique ability to generate controlled variations enables \csim to serve as a diagnostic test-bed to assess model sensitivity to changing weather, daytime, pitch, altitude, and layout conditions. 
    \item Finally, we show that incorporating additional modalities (depth) while training aerial scene-understanding models can improve aerial scene recognition, enabling further development of multi-modal segmentation models.
\end{packed_itemize}

\section{Related Work}

\begin{table*}[ht!]
\centering
\setlength{\tabcolsep}{1pt}
\resizebox{\linewidth}{!}{
\begin{tabular}{lccccccccccccc}
\toprule
\multirow{2}{*}{\textbf{Dataset}} & \multirow{2}{*}{\textbf{Controlled Variations}} && \multicolumn{3}{c}{\textbf{Diversity}} && \multirow{2}{*}{\textbf{Annotation Diversity}} & \multirow{2}{*}{\textbf{Altitude}} & \multirow{2}{*}{\textbf{Perspective}} & \multirow{2}{*}{\textbf{Resolution}} && \multirow{2}{*}{\textbf{Scale}} \\
\cmidrule{4-6}
&&&Town & Daytime & Weather &&&&&&& \\
\midrule
\multicolumn{13}{l}{\textbf{Real}}\\
\midrule
\texttt{1} UAVid~\cite{lyu2020uavid} & \redxmark && \redxmark & \redxmark & \redxmark && S & Med & Obl. & $3840\times2160$ && $0.42$k  \\
\texttt{2} AeroScapes~\cite{aeroscapes} & \redxmark && \redxmark & \redxmark & \redxmark && S & (Low, Med) & (Obl., Nad.) & $1280\times720$ && $3.27$k \\
\texttt{3} ICG Drone~\cite{drone-dataset} & \redxmark && \redxmark & \redxmark & \redxmark && S & Low & Nad. & $6000\times4000$ && $0.6$k  \\
\midrule
\multicolumn{13}{l}{\textbf{Synthetic}}\\
\midrule
\texttt{4} Espada~\cite{espada} & \redxmark && \greencmark & \redxmark & \redxmark && D & (Med, High) & Nad. & $640\times480$ && $80$k  \\
\texttt{5} UrbanScene3D~\cite{UrbanScene3D} & \redxmark && \greencmark & \redxmark & \redxmark && - & Med & Obl. & $6000\times4000$ && $128$k  \\
\texttt{6} SynthAer~\cite{synthaer} & \greencmark && \redxmark & \greencmark & \redxmark && S & (Low, Med) & Obl. & $1280\times720$ && $\sim0.77$k  \\
\texttt{7} MidAir~\cite{midAir} & \redxmark && \greencmark & \greencmark & \greencmark && (S,D) & Low & (Obl., Nad.) & $1024\times1024$ && $119$k   \\
\texttt{8} TartanAir~\cite{wang2020tartanairdatasetpushlimits} & \redxmark && \greencmark & \greencmark & \greencmark && (S,D) & Low & (Fwd., Obl.) & $640\times480$ && $\sim1$M \\
\texttt{9} VALID~\cite{valid} & \redxmark && \greencmark & \greencmark & \greencmark && (S,I,D) & (Low, Med, High) & Nad. & $1024\times1024$ && $6.7$k \\
\texttt{10} SynDrone~\cite{rizzoli2023syndrone} & \redxmark && \greencmark & \redxmark & \redxmark && (S,D) & (Low, Med, High) & (Obl., Nad.) & $1920\times1080$ && $72$k \\
\rowcolor{blue!12}
\texttt{11} \csim & \greencmark && \greencmark & \greencmark & \greencmark && (S,I,D) & (Low, Med, High) & (Fwd., Obl., Nad.) & $2160\times1440$ && $33.6$k \\
\bottomrule
\end{tabular}}
\caption{\textbf{\csim compared with other Real and Synthetic Datasets.} We compare \csim (row $11$) with other real (rows $1-3$) and synthetic (rows $4-10$) aerial datasets across several axes: (i) \textbf{Controlled Variations} -- the ability to reproduce the exact viewpoint under different variations from fine-grained scene metadata, (ii) \textbf{Diversity} -- diversity of map layouts (rural, urban), weather and daytime conditions in the provided images, (iii) \textbf{Annotation Diversity} -- supporting dense annotations across depth (D), semantic(S) and instance segmentation (I) tasks, (iv) \textbf{Altitude} -- altitude of image capture; Low is $<30$m, Med is $\in [30, 50]$m and High is $>50$m, (v) \textbf{Perspective} -- UAV pitch angle during image capture; Fwd. is forward view with $\theta = 0^{\circ}$, Obl. is oblique view with $\theta \in (0^{\circ}, 90^{\circ})$ and Nad. is nadir view with $\theta=90^{\circ}$ ($\theta$ is pitch), (vi) \textbf{Resolution} -- resolution of the images, (vii) \textbf{Scale} -- number of images. We see that while existing datasets might be lacking in a subset of criteria, \csim fulfills all of these.}
\label{table:syn_dataset_comparison}
\end{table*}

\label{sec:related_work}
\par \noindent
\textbf{Ground-view Synthetic Datasets.}
Real-world ground-view scene-understanding datasets (Cityscapes~\cite{cordts2016cityscapes}, Mapillary~\cite{mapillary}, BDD-100K~\cite{yu2020bdd100k}, Dark Zurich~\cite{SDV19}) fail to capture the full range of variations that exist in the world. Synthetic data is a popular alternative for generating diverse and bountiful views. GTAV~\cite{sankaranarayanan2018GTA}, Synthia~\cite{ros2016synthia}, and VisDA-C~\cite{visdac} are some of the widely-used synthetic datasets. These datasets can be curated using underlying simulators, such as GTAV~\cite{sankaranarayanan2018GTA} game engine and \crl ~\cite{dosovitskiy2017carla} simulator and offer a cost-effective and scalable way to generate large amounts of labeled data under diverse conditions. 
Similar to SELMA~\cite{testolina2022selma} and SHIFT~\cite{sun2022shift}, we use \crl~\cite{dosovitskiy2017carla} as the underlying simulator for \csim.
\par\noindent
\textbf{Real-World Aerial Datasets.}
To support remote sensing applications, it is crucial to have access to datasets that offer aerial-specific views. Datasets such as GID~\cite{GID2020}, DeepGlobe~\cite{deepglobe}, ISPRS2D~\cite{isprs}, and FloodNet~\cite{rahnemoonfar2020floodnet} primarily provide nadir perspectives and are designed for scene-recognition and understanding tasks. However, this study specifically focuses on lower altitudes, which are more relevant to UAVs, enabling object identification. Unfortunately, there is a scarcity of high-resolution real-world datasets based on UAV imagery emphasizing object identification. Existing urban scene datasets, like Aeroscapes~\cite{aeroscapes}, UAVid~\cite{lyu2020uavid}, VDD~\cite{cai2023vdd}, UDD~\cite{chen2018large}, UAVDT~\cite{du2018unmanned}, VisDrone~\cite{VisDrone}, Semantic Drones~\cite{drone-dataset} and others, suffer from limited sizes and a lack of diverse images under different conditions. This limitation raises concerns regarding model robustness and generalization.
\par \noindent
\textbf{Synthetic Aerial Datasets.}
Simulators can facilitate affordable, reliable, and quick collection of large synthetic aerial datasets, which aids in fast prototyping, improves real-world performance by enhancing robustness, and enables controlled studies on varied conditions. One such high-fidelity simulator, AirSim~\cite{airsim}, used for development and testing of autonomous systems (in particular, aerial vehicles), is the foundation of several synthetic UAV-based datasets -- MidAir~\cite{midAir}, Espada~\cite{espada}, Tartan Air~\cite{wang2020tartanairdatasetpushlimits}, UrbanScene3D~\cite{UrbanScene3D} and VALID~\cite{valid}. \crl ~\cite{dosovitskiy2017carla} is another such open-source simulator that is the foundation of datasets like SynDrone\cite{rizzoli2023syndrone}. However, these datasets fall short in capturing real-world irregularities, lack deterministic re-generation capabilities, controlled diversity in weather and daytime conditions, and exhibit skewed representation for certain classes (differences summarized in Table. \ref{table:syn_dataset_comparison}). This restricts their ability to generalize well to real-world datasets and their usage as a diagnostic tool for studying the controlled effect of diversity on the performance of computer vision perception tasks. To enable such studies, \csim offers images featuring varied scenes, diverse weather, daytime, altitude, and pitch variations while incorporating real-world irregularities and addressing skewed class representation along with simultaneous depth, semantic, and instance segmentation annotations.

\begin{figure}[t]
\centering
\begin{subfigure}{0.58\linewidth}
    \includegraphics[width=\columnwidth]{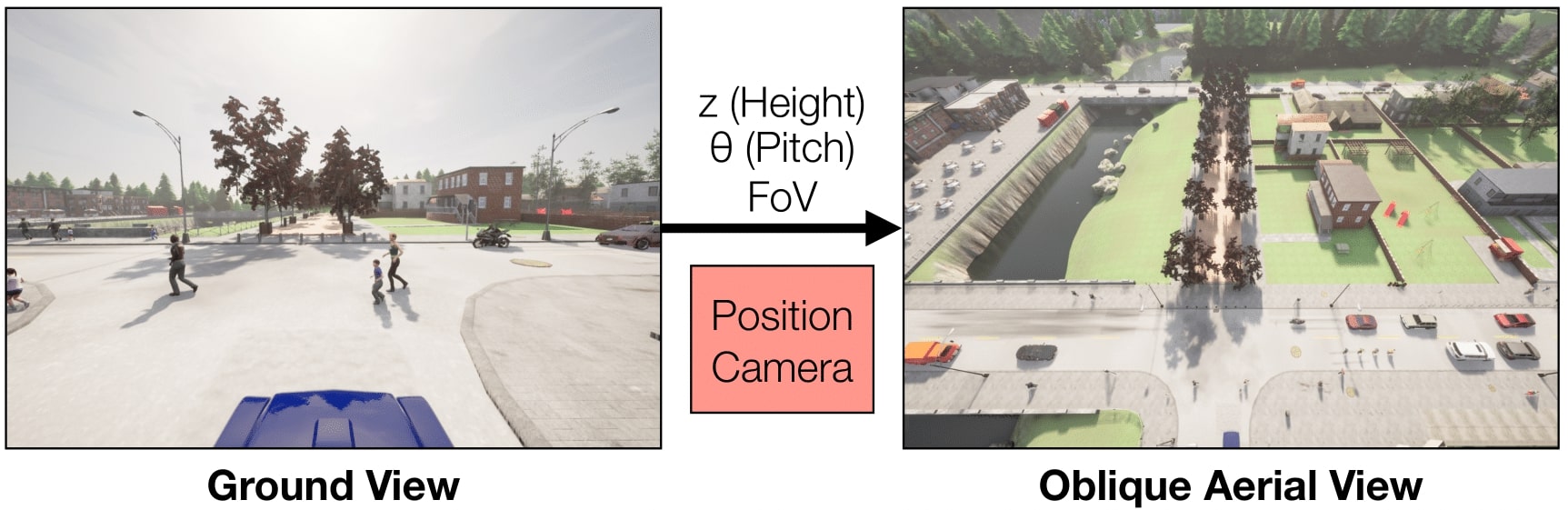}
    \caption{\csim Ground View $\rightarrow$ (Oblique) Aerial View. }
    \label{fig:cam_pos}
\end{subfigure}%
\begin{subfigure}{0.38\linewidth}
    \includegraphics[width=\linewidth]{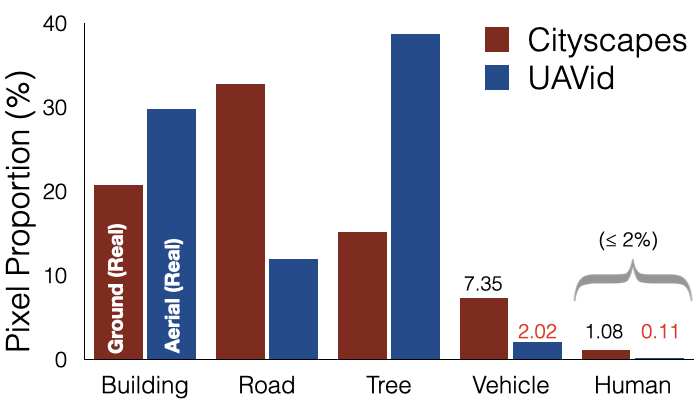}
    \caption{Pixel proportions }
    \label{fig:miou_gva}
\end{subfigure}%
    \caption{\textbf{Ground View $\rightarrow$ (Oblique) Aerial View.} (a)  The same scene viewed in Ground View vs Aerial View exhibits a significant difference in pixel proportion especially across the tail classes (\texttt{vehicle}, \texttt{human}) (b) For a subset of commonly annotated classes across CityScapes~\cite{cordts2016cityscapes} (red), \ua~\cite{lyu2020uavid} (dark blue) , we show the percentage of pixels occupied by different classes. Aerial scenes (in \ua) have significant under-representation of tail classes (\texttt{vehicle}, \texttt{human}). }
\label{fig:pixel_prop_aerial_ground}

\end{figure}

\section{\csim}
\label{sec:methodology}
\par\noindent
We curate \csim using, \crl~\cite{dosovitskiy2017carla}\footnote{\href{https://carla.org/}{https://carla.org/}} $0.9.14$, which is a flexible and realistic open-source autonomous vehicle simulator. 
The simulator offers a wide range of sensors, environmental configurations, and varying rendering configurations. As noted earlier, we take several important considerations into account while curating \csim images. These include strategies for obtaining diverse synthetic data and embedding real-world irregularities, avoiding correlated images, addressing skewed class representations, and more. In this section, we first discuss such desiderata and then describe our procedural image curation algorithm. Finally, we describe different aspects of the curated dataset.

\subsection{(Synthetic) Aerial Image Desiderata}
\label{sec:desiderata}

Before investigating the image curation pipeline, we first outline a set of
desiderata taken into account while curating synthetic aerial images in \csim.

\begin{enumerate}[wide, labelwidth=0pt, labelindent=0pt]

    \item[1.] \textbf{Viewpoint Reproducibility:}
    Critical to understanding how models respond to changing conditions is the ability to evaluate them under scenarios where only one variable is altered. However, any effort to do so in the real-world would be uncontrolled, due to its dynamic (constantly changing) nature. In contrast, simulated data allows us to do so by providing control over image generation conditions. Unlike certain existing aerial datasets that do not support this feature (see Table.~\ref{table:syn_dataset_comparison}), we do so in \csim by additionally storing comprehensive metadata for each viewpoint (and image), including details about camera world coordinates, orientation, and all movable/immovable actors and objects in the scene. We couple this with rigorous consistency checks for image generation that verify the number of actors, their location, sensor height, pitch, etc. This meticulous approach enables us to reproduce the same viewpoint under multiple conditions effortlessly.

    \item[2 .] \textbf{Adequate Representation of Tail Classes:} Unlike ground-view datasets, pixel distribution of classes in aerial images is substantially more long-tailed (see Fig.~\ref{fig:pixel_prop_aerial_ground} (a); classes with smaller object size, \texttt{humans}). This substantial difference in class proportions severely affects the performance of tail classes in aerial datasets when compared to ground-view datasets (see Fig.~\ref{fig:pixel_prop_aerial_ground} (b)), thus making visual recognition tasks harder.  
    To counter this, we consider structured spawning of \texttt{human}s to ensure adequate representation (see Sec. ~\ref{sec:dset_algo}).

    \item [3. ] \textbf{Adequate Height Variations:} Aerial images are captured at different altitudes to meet specific needs. Lower altitudes ($5$-$15$m) are optimal for high-resolution photography and detailed inspections. Altitudes ranging from $30$m-$50$m strike a balance between fine-grained detail and a broader perspective, making them ideal for surveillance. Altitudes above $50$m are suitable for capturing extensive areas, making them ideal for surveying and mapping. Existing datasets (synthetic or real) often focus on ``specific'' altitude ranges (see Table.~\ref{table:syn_dataset_comparison}, Image Capture columns), limiting their adaptability to different 
    scenarios. With \csim, our aim is to provide flexibility in altitude sampling, thus accommodating various real-world requirements.
    We curate \csim images at heights of $15$m, $35$m, and $60$m. Additionally, recognizing imperfections in real-world actuation, we induce slight jitter in the height values ($\Delta h\sim\mathcal{N}(1, 2.5$m$)$) to simulate realistic data sampling.

    \item [4. ] \textbf{Adequate Pitch Variations:} Similar to height, aerial images can be captured from $3$ primary perspectives or pitch angles ($\theta$): nadir ($\theta = 90^{\circ}$), oblique ($\theta \in (0^{\circ}, 90^{\circ})$), or forward ($\theta = 0^{\circ}$) views (see Table.~\ref{table:syn_dataset_comparison}, Image Capture columns). The nadir view (directly perpendicular to the ground plane), preserves object scale while forward views are well-suited for tasks like UAV navigation and obstacle detection. Oblique views, on the other hand, capture objects from a side profile, aiding object recognition and providing valuable context and depth perspective often lost in nadir and forward views. To ensure widespread utility, \csim data generation process is designed to support all these viewing angles, with a particular emphasis on oblique views (the most common one). Similar to height, pitch variations allow models trained on \csim to generalize to different viewpoint variations. We use $\theta = 45^{\circ} \text{ and }60^{\circ}$ for oblique-views and introduce jitter ($\Delta \theta\sim\mathcal{N}(1, 5^{\circ})$) to mimic real-world data sampling.

    \item[5.] \textbf{Adequate Map Variations:} In addition to sensor locations, it is equally important to curate aerial images across diverse scene layouts. To ensure adequate map variations, we gather images from $8$ different \crl~\cite{dosovitskiy2017carla} towns (can be categorized as \textit{urban} or \textit{rural}), which provide substantial variations in the observed scene. These towns differ in layouts, size, road map design, building design, and vegetation cover. Fig.~\ref{fig:carla_uavid_pixel_prop} illustrates how images curated from different towns in \crl~\cite{dosovitskiy2017carla} differ in class distributions.

    \item[6.] \textbf{Adequate Weather \& Daytime Variations:}
    Training robust perception models using \csim that generalize to unforeseen environmental conditions necessitates the curation of annotated images encompassing various weather and daytime scenarios.
    To accomplish this, we generate \csim images from identical viewpoints under $5$ different variations -- ClearNoon, ClearSunset, MidRainNoon, ClearNight, and CloudyNoon.\footnote{Note that \crl~\cite{dosovitskiy2017carla} provides $14$ such conditions but we use only $5$ such conditions in this preliminary version of \csim.}
    Generating images in different conditions from the same perspectives allows us to (1) leverage diverse data for 
    improved
    generalization and (2) systematically investigate the susceptibility of trained models to variations in daytime and weather conditions.
    \item[7.] \textbf{Fine-grained Annotations:}
    To support a host of different computer vision tasks (segmentation, detection, multimodal recognition), we curate all \csim images with dense semantic, instance segmentation and depth annotations. We provide semantic annotations for a wide vocabulary of $28$ classes to support broad applicability (see Fig.~\ref{fig:teaser} column $4$ for an example).

\end{enumerate}

\subsection{\textbf{\csim} Image Generation}

\label{sec:dset_algo}
We generate \csim images from \crl~\cite{dosovitskiy2017carla} by taking the previously mentioned considerations into account. Curating images from \crl~\cite{dosovitskiy2017carla} broadly consists of two key steps: (1) positioning the agent camera in an aerial perspective and (2) procedurally guiding the agent within the scene to capture images. We accomplish the first by mimicking a UAV perspective in \crl~\cite{dosovitskiy2017carla} by positioning the ego vehicle (with RGB, semantic and depth sensors) based on specified (high) altitude ($h$) and pitch ($\theta$) values to generate aerial views (see Fig.~\ref{fig:cam_pos}).\footnote{This also requires setting other scenes -- weather, daytime, etc. -- and camera (notably the $\texttt{FoV}=110^{\circ}$ (field of view) and image resolution $ = 2160$x$1440$) parameters.} Once positioned, the agent is translated by fixed amounts to traverse the scene and capture images from various viewpoints (detailed in Sec.~\ref{sec:dataset} in the appendix). Initially, we generate $70$ data points for each of the $8$ town variations under ClearNoon conditions using the baseline $h=35$m$, \theta=45^{\circ}$ setting. 
Subsequently, following 
the traversal algorithm (see Sec.~\ref{sec:dataset} in the appendix),
we re-generate these datapoints across $5$ weather conditions and $12$ height/pitch variations, 
resulting in $70\times8\times5\times12=33,600$ images.\\

\begin{figure}[t]
\centering
    \begin{subfigure}{0.3\linewidth}
    
        \includegraphics[width=\linewidth]{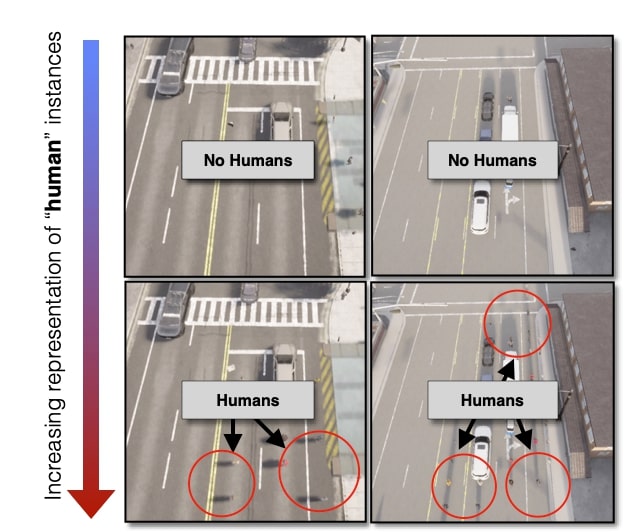}
        \caption{ HumanSpawn() (HS) }
        \label{fig:hum_spawn1}
    \end{subfigure}
    \begin{subfigure}{0.3\linewidth}
        \includegraphics[width=\linewidth]{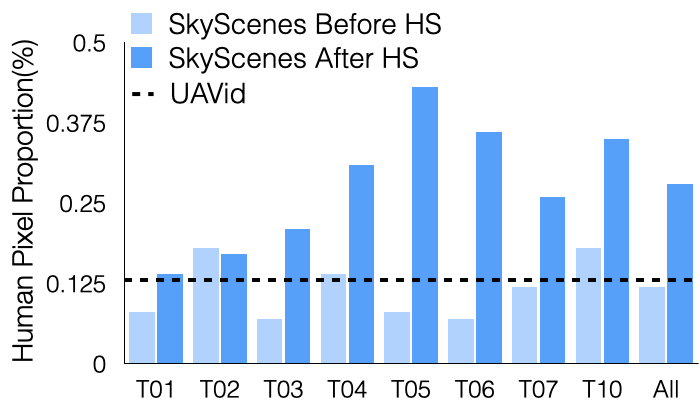}
        \caption{ Human Pixel Proportion}
        \label{fig:hum_spawn2}
    \end{subfigure}%
    \begin{subfigure}{0.3\linewidth}
            \centering
        \resizebox{\linewidth}{!}{
            \small
            \begin{tabular}{lcc}
                \toprule
                &\multicolumn{2}{c}{\textbf{mIoU$(\uparrow)$}} \\
                \cmidrule{2-3}
                \textbf{Eval Data} & \texttt{human} & All \\
                \midrule
                \csim\\
                \midrule
                \texttt{1} Before HS & $43.03$ & $80.87$ \\
                \texttt{2} After HS & $\textbf{61.79}$ & $\textbf{84.07}$ \\
                \midrule
                \csim $\rightarrow$ \ua\\
                \midrule
                \texttt{3} Before HS & $4.71$ & $45.11$ \\
                \texttt{4} After HS & $\textbf{10.21}$ & $\textbf{47.09}$ \\
                \bottomrule
            \end{tabular}
        }
        \caption{ mIoU improvement w/ HS}
        \label{table:human_spawn_tab}

    \end{subfigure}
    
    \caption{\textbf{\csim w/ HumanSpawn() increases representation of \texttt{humans} and improves \csim$\rightarrow$\ua(S$\rightarrow$U) performance.} (a) Incorporating HumanSpawn() in the image generation pipeline for \csim increases the proportion of \texttt{humans} in snapshots (\textbf{[Top]}$\rightarrow$\textbf{[Bottom]}). (b) Increased representation of \texttt{humans} across all the layout variations in \csim after HumanSpawn(), with the dotted line representing the proportion of \texttt{humans} in UAVid (c)  Training on HumanSpawn (HS) \csim images improves the model's ability to recognize \texttt{humans} (improved mIoU). T = Town.}
    \label{fig:human_spawn_fig}

\end{figure}

\par \noindent
\textbf{Checks and Balances.}
Additionally, we ensure the following checks and balances while curating \csim images.\\
\par \noindent
\textbf{$\triangleright$ Avoiding Overly Correlated Frames for Viewpoints.} \crl~\cite{dosovitskiy2017carla} uses a traffic manager with a PID controller to control the egocentric vehicle based on current pose, speed, and a list of waypoints at every pre-defined time step. 
Curating images at every time step (or tick) results in highly correlated frames with little change in object position. 
Since overly correlated frames are not very useful when training models for static scene understanding, we move the camera by a fixed distance multiple times before saving a frame. This also helps with moving dynamic actors by a considerable amount in the scene. Additionally, pedestrian objects are regenerated before saving an image, which adds randomness to the spawning and placement of pedestrians,  reducing the correlation between frames.

\par \noindent
\textbf{$\triangleright$ Adequate Representation of \texttt{humans}.} 
Real-world scenes often exhibit a long-tailed distribution in pixel proportions, particularly in aerial images where variations in object sizes and camera positions contribute to significant under-representation of the tail classes (in Fig.~\ref{fig:pixel_prop_aerial_ground}, for the shared set of classes across UAVid~\cite{lyu2020uavid} (aerial) and Cityscapes~\cite{cordts2016cityscapes} (ground), we can see that the class distributions are different and aerial images are significantly more heavy-tailed). As a result, naively spawning \texttt{humans} (rarest class) in \crl~\cite{dosovitskiy2017carla} is detrimental for eventual task performance -- for the \texttt{human} class, a \csim trained DAFormer~\cite{hoyer2022daformer} (with HRDA~\cite{hoyer2022hrda} source training; MiT-B5~\cite{xie2021segformer} backbone) model leads to an in-distribution performance of $43.03$ mIoU and out-of-distribution (\csim$\rightarrow$UAVid~\cite{lyu2020uavid}) performance of $4.71$ mIoU. To counter this under-representation issue, we design an algorithm, HumanSpawn() (see Sec.~\ref{sec:dataset} in appendix), to explicitly spawn more \texttt{human} instances while curating \csim images. HumanSpawn() increases \texttt{human} instances by $40-200$ per snapshot, improving the proportion of densely annotated \texttt{humans} in \csim by approximately $10$ times (see Fig. ~\ref{fig:human_spawn_fig} (a) \& Fig.~\ref{fig:human_spawn_fig} (b)). This improvement in \texttt{human} representation is also evident in eventual task performance, with in-distribution and out-of-distribution mIoUs for \texttt{humans} increasing from $43.03$ to $61.79$ ($+18.76$) and $4.71$ to $10.21$ ($+5.50$) respectively (see Table.~\ref{fig:human_spawn_fig} (c)).

\subsection{\textbf{\csim: Dataset Details}}
\label{sec:dset_stats}
\begin{figure*}[t]
\centering
\includegraphics[width=\textwidth]{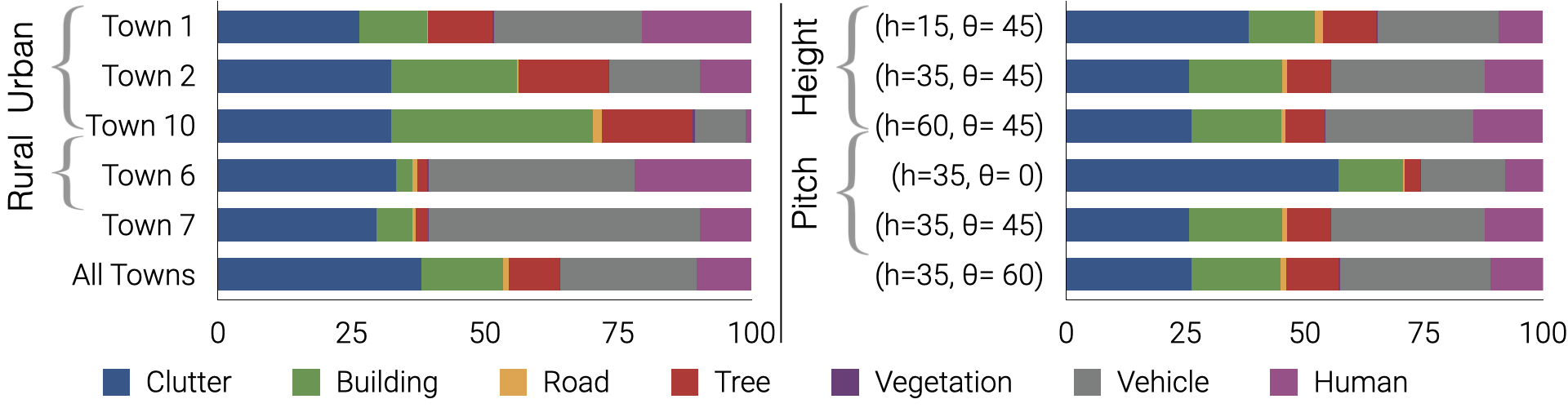}
\caption{\textbf{Class-distribution Diversity in \csim.} We show how the distribution of densely-annotated pixels varies across different \csim conditions. \textbf{[Left]} Class distribution varies substantially within and across urban and rural map layouts. \textbf{[Right]} Similarly, for the same \csim layouts (and viewpoints) class distribution varies substantially across different height and pitch values.}
\label{fig:carla_uavid_pixel_prop}

\end{figure*}
\par \noindent
\textbf{Annotations.} We provide semantic, instance and depth annotations for every image in \csim. Semantic annotations in \csim by default are across $28$ classes. These are building, fence, pedestrian, pole, roadline (markings on road), road, sidewalk, vegetation, cars, wall, traffic sign, sky, bridge, railtrack, guardrail, traffic light, water, terrain, rider, bicycle, motorcycle, bus, truck and others (see Fig.~\ref{fig:teaser} for an example and  Sec.~\ref{sec:class} in appendix for definitions)

\par \noindent
\textbf{Training, Validation and Test Splits.} \csim
has $70$ images per town (across $8$ towns) for each of the $5$ weather and daytime conditions, and $12$ height \& pitch combinations, resulting in a total of $33,600$ images. We use $80$\% ($26,880$ images) of the dataset for training models, with $10$\% ($3,360$ images) each for validation and testing (see Sec.~\ref{sec:splits} in appendix).

\par\noindent
\textbf{Class Distribution(s).} 
In Fig.~\ref{fig:carla_uavid_pixel_prop}, we highlight how the distribution of classes changes across variations within \csim -- rural and urban map layouts and height and pitch specifications. \csim exhibits substantial diversity in class distributions across such conditions, allowing these individual conditions to serve as diagnostic splits to assess model sensitivity (see Sec.~\ref{sec:diagnostic}).

\section{Experiments}
\label{sec:experiments}
\begin{table}[ht]
\centering
\begin{minipage}[t]{0.4\textwidth}
\centering
\resizebox{\linewidth}{!}{
\begin{tabular}{lccccc}
\toprule
\multirow{2}{*}{\textbf{Source}} & \multicolumn{3}{c}{\textbf{(Target) Real-World mIoU ($\uparrow$)}}\\ 
& \ua & \aero & \icg \\ 
\midrule
DeepLabv2 (R-101)~\cite{chen2017deeplab} \\
\midrule
\texttt{1} \syn & $39.86$ & $24.50$ & $8.20$ \\
\texttt{2} \csim & $\textbf{41.82}$ & $\textbf{26.94}$ & $\textbf{15.14}$ \\
\midrule
DAFormer (MiT-B5)~\cite{hoyer2022hrda} \\
\midrule
\texttt{3} \syn & $42.31$ &$30.53$ & $15.92$\\
\texttt{4} \csim & $\textbf{47.09}$ &$\textbf{40.72}$ & $\textbf{25.91}$ \\
\midrule
Rein (DINOv2) \\
\midrule
\texttt{5} \syn & $\textbf{54.92}$ &$40.28$ & $20.01$\\
\texttt{6} \csim & $54.19$ &$\textbf{43.96}$ & $\textbf{28.10}$ \\
\bottomrule
\end{tabular}
}
\caption{\textbf{Models trained on \csim generalize well to the real-world.} We train semantic segmentation models (DeepLabv2~\cite{chen2017deeplab}, DAFormer~\cite{hoyer2022daformer}, Rein~\cite{wei2024stronger}) on \csim, \syn~\cite{rizzoli2023syndrone}, and real datasets and show how training models on \csim provides better out-of-the-box generalization to multiple real-world datasets.}
\label{table:sim2real_semseg}
\end{minipage}
\hspace{0.5em} 
\begin{minipage}[t]{0.57\textwidth}
\centering
\resizebox{\linewidth}{!}{
\begin{tabular}{lccccccccc}
\toprule
\multirow{2}{*}{\textbf{Source}} & \multicolumn{8}{c}{\textbf{(Target) Real-World IoU ($\uparrow$)}}\\ 
& \multicolumn{2}{c}{\ua} && \multicolumn{2}{c}{\aero} && \multicolumn{2}{c}{\icg} \\
\cmidrule{2-3} \cmidrule{5-6} \cmidrule{8-9}
& vehicle & human && vehicle & person && vehicle & person \\ 
\midrule
DAFormer (MiT-B5)~\cite{hoyer2022hrda} \\
\midrule
\texttt{1} \syn & $42.52$ & $8.27$ && $49.77$ & $0.77$ && $0.24$ & $0.38$\\
\texttt{2} \csim & $\textbf{63.64}$ & $\textbf{10.21}$ && $\textbf{80.99}$ & $\textbf{3.09}$ && $\textbf{39.71}$ & $\textbf{45.89}$\\
\midrule
Rein (DINOv2)~\cite{wei2024stronger} \\
\midrule
\texttt{3} \syn & $68.68$ & $21.6$ && $84.2$ & $10.29$ && $7.91$ & $0$\\
\texttt{4} \csim & $\textbf{75.14}$ & $\textbf{25.52}$ && $\textbf{87.71}$ & $\textbf{21.67}$ && $\textbf{50.91}$ & $\textbf{77.93}$\\
\bottomrule
\end{tabular}
}
\caption{\textbf{\csim training exhibits strong real-world generalization for tail classes.} We show how DAFormer~\cite{hoyer2022daformer} and Rein~\cite{wei2024stronger} models trained on \csim exhibit improved real-world generalization compared to those trained on \syn~\cite{rizzoli2023syndrone} for under-represented tail classes (\texttt{vehicles} and \texttt{humans}). \csim training facilitates better recognition of tail class instances.}
\label{table:sim2real_semseg_tail}
\end{minipage}
\end{table}

We conduct semantic segmentation experiments with \csim to assess a few different factors. First, we check if training on \csim is beneficial for real-world transfer. Second, we check if \csim can augment real-world training data in low and full shot regimes. Third, we check if variations in \csim can be used to assess the sensitivity of trained models to changing conditions. Finally, we check if using additional modality information (depth) can help improve aerial scene understanding.

\par \noindent
\textbf{Synthetic and Real Datasets.} We compare real-world generalization performance of training on \csim with \syn~\cite{rizzoli2023syndrone}, a recently proposed synthetic aerial dataset also curated from \crl~\cite{dosovitskiy2017carla} featuring $3$ different  $(h, \theta)$ conditions across $8$ different map layouts. We assess performance on $3$ real-world aerial datasets -- \ua~\cite{lyu2020uavid}, \aero~\cite{aeroscapes}, \icg ~\cite{drone-dataset}. Since different datasets have different class vocabularies and definitions, for our experiments, we adapt the class vocabulary of the synthetic source dataset to that of the target real-world datasets (see Sec.~\ref{sec:Merging} in appendix for class merging and assignment schemes). Additionally, since different real aerial datasets have been captured from different heights and pitch angles, we train models on $(h, \theta)$ subsets of synthetic datasets that are aligned with corresponding real data $(h, \theta)$ conditions. We provide additional details for the real aligned synthetic data selection and model evaluation in Sec.~\ref{sec:syn2real} in the appendix. 
\par \noindent
\textbf{Models.} We use (1) CNN -- DeepLabv2~\cite{chen2017deeplab} (ResNet-101~\cite{he2016deep}), (2) transformer  -- DAFormer~\cite{hoyer2022daformer} (with HRDA~\cite{hoyer2022hrda} source training; MiT-B5~\cite{xie2021segformer} backbone) and (3) Vision Foundation Model -- Rein~\cite{wei2024stronger} (LoRA~\cite{hu2021lora} fine-tuned Dino-V2~\cite{oquab2024dinov2} backbone) based semantic segmentation architectures for our experiments. We provide implementation details for our experiments in Sec.~\ref{sec:experiment} in appendix.

\begin{table}
\centering
\resizebox{\textwidth}{!}{
\begin{tabular}{cc}
\begin{minipage}[t]{0.48\textwidth}
\centering
\setlength{\tabcolsep}{5pt}
\resizebox{\linewidth}{!}{
\begin{tabular}{lccccc}
\toprule
\multirow{2}{*}{\textbf{Source}} & \multicolumn{5}{c}{\textbf{(Target) Real World mIoU ($\uparrow$)}}\\ 
& 5\% & 10\% & 25\% &50\% & 100\%\\ 
\midrule
DeepLabv2 (R-101)~\cite{chen2017deeplab} \\
\midrule
\texttt{1} {Only Real} & {$48.25$} & {$55.29$}& {$62.86$} & {$66.81$} & {$68.53$} \\
\texttt{2} \csim + Real (JT) & $\textbf{59.27}$ &$\textbf{64.15}$ & $\textbf{68.11}$ &$\textbf{70.18}$ & $69.51$ \\
\texttt{3} \csim + Real (FT) &$53.67$ &$60.61$ &$65.57$ & $68.54$ & $\textbf{69.70}$ \\
\midrule
DAFormer (MiT-B5)~\cite{hoyer2022hrda} \\
\midrule
\texttt{4} {Only Real} & {$60.59$} & {$65.63$}& {$70.31$} & {$72.16$} & {$72.47$}\\
\texttt{5} \csim + Real (JT) &$\textbf{62.97}$& $\textbf{67.58}$&$70.20$ & $71.83$ & $72.25$\\
\texttt{6} \csim + Real (FT) & $60.90$&$66.79$& $\textbf{70.41}$ & $\textbf{72.63}$ & $\textbf{73.02}$ \\
\midrule
Rein (DINOv2)~\cite{wei2024stronger} \\
\midrule
\texttt{7} {Only Real} & {$64.04$} & {$71.87$}& {$73.87$} & {$76.05$} & $76.55$\\
\texttt{8} \csim + Real (JT) & $69.15$ &$73.54$ & $\textbf{75.07}$ & $76.08$ & $76.54$\\
\texttt{9} \csim + Real (FT) &$\textbf{70.07}$ &$\textbf{73.99}$ &$75.01$ & $\textbf{76.44}$ & $\textbf{76.89}$ \\
\bottomrule
\end{tabular}
}
\caption{\textbf{\csim augmented real data improves performance in low shot regimes.} We compare DeepLabv2~\cite{chen2017deeplab}, DAFormer~\cite{hoyer2022daformer}, and Rein~\cite{wei2024stronger} models trained using varying percentages of labeled UAVid~\cite{lyu2020uavid} images. Models are either trained jointly on \csim and UAVid (JT) or pretrained on \csim and finetuned on UAVid (FT). Augmenting real data with \csim enhances real-world generalization in low-shot scenarios.}
\label{table:sim+real_semseg}
\end{minipage}
&
\begin{minipage}[t]{0.48\textwidth}
\centering
\includegraphics[width=\linewidth]{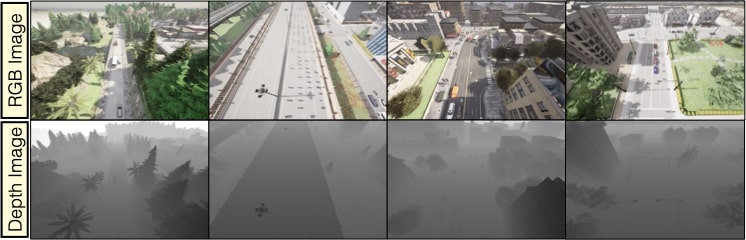}
\captionof{figure}{\csim RGB Images and corresponding depth images generated using depth sensor for $h=35$, $\theta=45^{\circ}$, and ClearNoon setting across four different town layouts.}
\label{fig:example}
\setlength{\tabcolsep}{3pt}
\resizebox{\linewidth}{!}{
\begin{tabular}{lccccccccc}
\toprule
\multirow{2}{*}{\textbf{Sensors}}& \multicolumn{8}{c}{\textbf{\csim Test IoU ($\uparrow$)}} \\ \cmidrule{2-9}
& clutter & building & road & tree & low-veg. & vehicle & human & Avg\\
\midrule
\texttt{1} RGB & $87.80$	&$94.54$	&$94.07$	&$88.03$	&$69.37$	&$82.89$	&$43.35$	&$80.01$ \\
\texttt{2} RGB+D & $\textbf{90.64}$	&$\textbf{95.97}$	&$\textbf{94.87}$	&$\textbf{89.41}$	&$\textbf{74.36}$	&$\textbf{86.87}$	&$\textbf{50.47}$	&$\textbf{83.22}$ \\
\bottomrule
\end{tabular}
}
\captionof{table}{\textbf{Multi-modal Segmentation in \csim.} We evaluate M3L~\cite{maheshwari2023m3l} multimodal segmentation architectures with MiT-B5~\cite{xie2021segformer} backbones using RGB and RGB+D data in \csim. Additional sensors improve aerial scene understanding significantly across various classes in UAVid~\cite{lyu2020uavid}.}
\label{table:s2s_multimod_depth}
\end{minipage}
\end{tabular}
}
\end{table}
\subsection{\csim$\to$ Real Transfer}

\par \noindent
\textbf{$\triangleright$ \csim trained models generalize well to real-settings.} In Table.~\ref{table:sim2real_semseg}, we show how models trained on \csim exhibit strong out-of-the box generalization performance on multiple real world datasets. We find that \csim pretraining exhibits stronger generalization compared to \syn\\ ~\cite{rizzoli2023syndrone} across both CNN and transformer segmentation backbones. In Table.~\ref{table:sim2real_semseg_tail}, we show how generalization improvements are more pronounced for under-represented tail classes (\texttt{vehicles} and \texttt{humans}). Comparison across all classes is provided in Tables~\ref{table:uavid_expert}, \ref{table:aero_expert} and~\ref{table:icg_expert} in appendix. 

\par \noindent
\textbf{$\triangleright$ \csim can augment real training data.} In addition to zero-shot real-world generalization, akin to other synthetic aerial datasets, we also show how \csim is useful as additional training data when labeled real-world data is available. In Table.~\ref{table:sim+real_semseg}, for \csim$\to$\ua~\cite{lyu2020uavid}, we compare models trained only using $5\%, 10\%, 25\%$, $50\%$, $100\%$ of the $200$ \ua~\cite{lyu2020uavid} training images with counterparts that were either pretrained using \csim data or additionally supplemented with \csim data at training time. We find that in low-shot regimes (when little ``real'' world data is available), \csim data (either explicitly via joint training or implicitly via finetuning) is beneficial in improving recognition performance (see Sec.~\ref{sec:real+syn} of appendix).

\begin{table*}[ht!]
\centering
\setlength{\tabcolsep}{2pt}
\resizebox{\linewidth}{!}{
\begin{tabular}{c c c c c}
\parbox{.37\linewidth}{
\centering
\begin{tabular}{lccc}
\toprule
\multirow{2}{*}{\textbf{Train}} & \multicolumn{3}{c}{\textbf{Test mIoU ($\uparrow$)}} \\
& Clear & Cloudy & Rainy \\
\midrule
\texttt{1} Clear & \textcolor{blue}{$73.91$} & \textcolor{blue}{$73.59$} & $69.95$\\
\texttt{2} Cloudy & $69.60$ & \textcolor{blue}{$74.02$} & $69.14$\\
\texttt{3} Rainy & $69.00$ & \textcolor{blue}{$73.36$} & \textcolor{blue}{$72.62$}\\
\bottomrule
\end{tabular}
\caption*{\small \textbf{(a) Weather Variation}}
\label{table:s2s_weather}
}
& 
\parbox{.36\linewidth}{
\centering
\begin{tabular}{lccc}
\toprule
\multirow{2}{*}{\textbf{Train}} & \multicolumn{3}{c}{\textbf{Test mIoU ($\uparrow$)}} \\
& Noon & Sunset & Night \\
\midrule
\texttt{1} Noon & \textcolor{blue}{$73.91$} & \textcolor{blue}{$71.16$} & $35.60$\\
\texttt{2} Sunset & \textcolor{blue}{$63.16$} & \textcolor{blue}{$66.53$} & $39.36$\\
\texttt{3} Night & $52.00$ & $57.35$ & \textcolor{blue}{$70.36$}\\
\bottomrule
\end{tabular}
\caption*{\small \textbf{(b) Daytime Variation}}
\label{table:s2s_daytime}
}
& 
\parbox{.32\linewidth}{
\centering
\begin{tabular}{lcc}
\toprule
\multirow{2}{*}{\textbf{Train}} & \multicolumn{2}{c}{\textbf{Test mIoU ($\uparrow$)}} \\
& Rural & Urban\\
\midrule
\texttt{1} Rural & \textcolor{blue}{$58.00$} & $35.90$\\
\texttt{2} Urban & $38.99$ & \textcolor{blue}{$73.16$}\\
\bottomrule
\end{tabular}
\caption*{\small \textbf{(c) Map Variation}}
\label{table:s2s_map}
}
& 
\parbox{.52\linewidth}{
\centering
\begin{tabular}{lcccc}
\toprule
& \multicolumn{4}{c}{\textbf{Test mIoU ($\uparrow$)}} \\
\multirow{2}{*}{\textbf{Height}} & \multicolumn{4}{c}{\textbf{Pitch}} \\
& $\theta=0^{\circ}$ & $\theta=45^{\circ}$ & $\theta=60^{\circ}$ & $\theta=90^{\circ}$\\
\midrule
\texttt{1} $h=15$m & \textcolor{blue}{48.50} & \textcolor{blue}{$50.71$} & $45.22$ & $42.21$\\
\texttt{2} $h=35$m & $50.49$ & \textcolor{blue}{$55.74$} & \textcolor{blue}{$57.11$} & $52.19$\\
\texttt{3} $h=60$m & $45.33$ & \textcolor{blue}{$49.79$} & \textcolor{blue}{$50.37$} & $44.62$\\
\bottomrule
\end{tabular}
\caption*{\small \textbf{(d) Height \& Pitch Variation}}
\label{table:s2s_hp_2}
}
\end{tabular}}
\caption{\textbf{Model Sensitivity to Changing Conditions.} We show how changing conditions (weather, daytime, map, viewpoint) in \csim can serve as diagnostic test splits to assess the sensitivity of trained DAFormer~\cite{hoyer2022daformer} semantic segmentation models. In (a) and (b), we evaluate models trained under different weather and daytime conditions across the same conditions. In (c), we evaluate models trained on rural and urban scenes across the same layouts. In (d), we evaluate a model trained on moderate height, pitch settings ($h=35$m$, \theta=45^{\circ}$) across different $h,\theta$ variations. 
Best numbers across each row condition is highlighted in \textcolor{blue}{blue.}} 
\label{table:s2s_all}
\end{table*}

\subsection{\csim as a Diagnostic Framework}
\label{sec:diagnostic}
As noted earlier, the images we curate in \csim contain several variations -- ranging from $5$ different weather and daytime conditions, rural and urban map layouts, and $12$ different height and pitch combinations (see Fig.~\ref{fig:carla_uavid_pixel_prop} for variations in class distributions). We curate images under such diverse conditions in a \textit{controlled manner} -- ensuring the same spatial coordinates for $(h, \theta)$ variations, same spatial coordinates and $(h, \theta)$ settings across different weather and daytime conditions, the same number of images across layouts. 
\par\noindent
This allows us to assess the sensitivity of trained models to one factor of variation ($h$, $\theta$, daytime, weather, map layout) by changing that specific aspect. We summarize some takeaways from such experiments in Table.~\ref{table:s2s_all}.
\par\noindent
In Table.~\ref{table:s2s_all} (a), we show how models trained in a certain weather condition are best at generalizing to the same condition at test-time. We make similar observations for daytime variations in Table.~\ref{table:s2s_all} (b). In Table.~\ref{table:s2s_all} (c), we show how models trained in rural conditions fail to perform well in urban test-time conditions and vice-versa. In Table.~\ref{table:s2s_all} (d), we evaluate a model trained under moderate $(h=35$m$, \theta=45^{\circ})$ conditions under different $(h, \theta)$ variations. We find that as altitudes increase, trained models are better at recognizing objects from oblique ($\theta\in(0^{\circ}, 90^{\circ})$) viewpoints. We provide exhaustive quantitative comparisons in  Sec.~\ref{sec:diagnostic_exp} in the appendix.

\begin{figure}
    \centering
    \includegraphics[width=\linewidth]{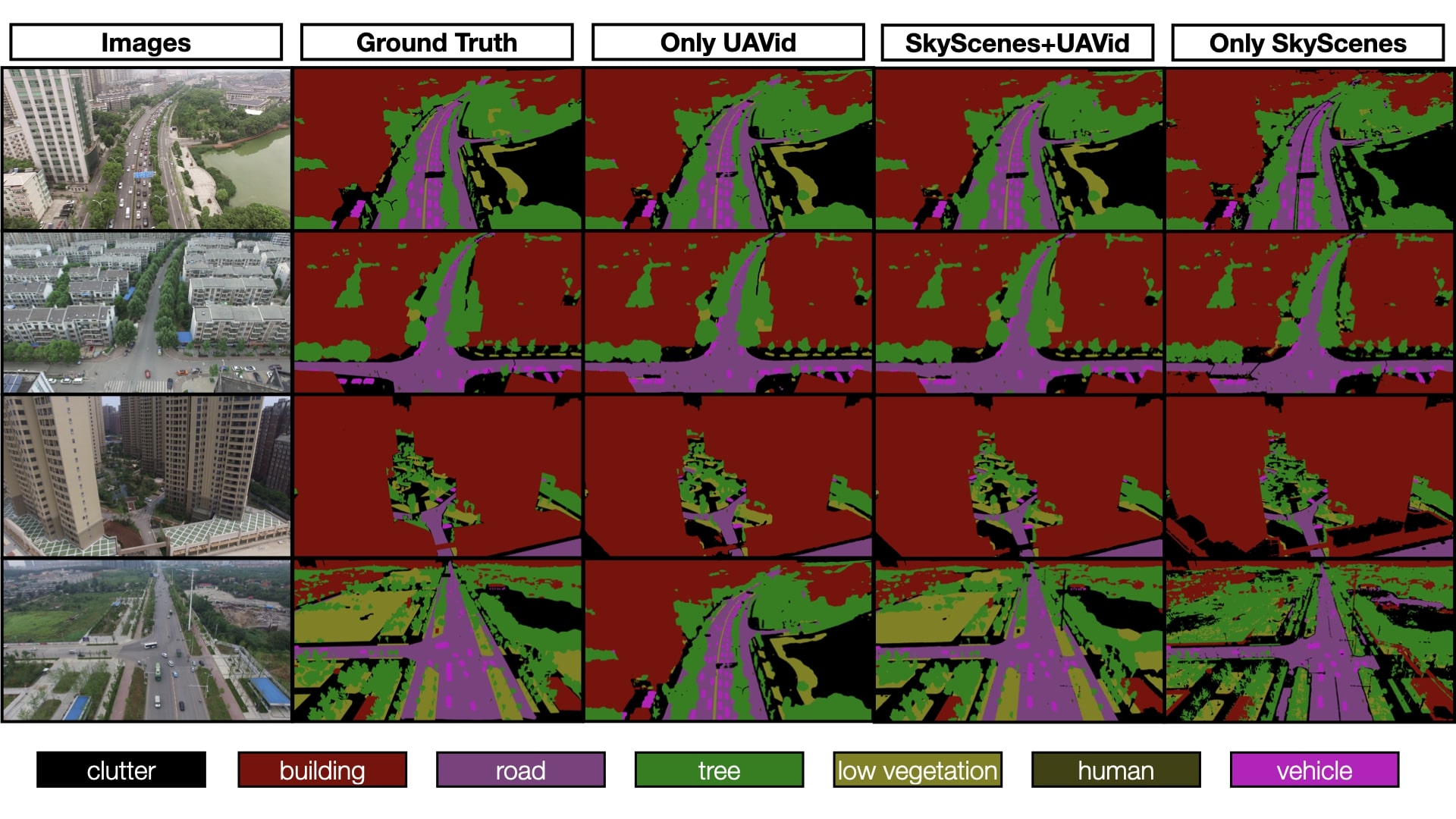}
    \caption{\textbf{\ua, \csim + \ua and \csim $\rightarrow$ \ua semantic segmentation predictions} Predictions on randomly selected UAVid~\cite{lyu2020uavid} validation images by a Rein~\cite{wei2024stronger} model trained on UAVid and \csim. Columns 1 and 2 show the original image and its ground truth. Columns 3, 4, and 5 display predictions from models trained exclusively on UAVid, jointly on \csim and UAVid, and exclusively on \csim, respectively.}
    \label{fig:lidar}

\end{figure}
\subsection{\csim Enables Multi-modal Dense Prediction}
Sensors on UAVs in deployable settings often include modalities beyond RGB cameras, such as depth sensors. These additional modalities can significantly enhance aerial scene understanding. In Table.~\ref{table:s2s_multimod_depth}, we investigate the impact of augmenting RGB data with depth observations from \csim viewpoints on aerial semantic segmentation using M3L~\cite{maheshwari2023m3l}, a multimodal segmentation model. Similar to our DAFormer~\cite{hoyer2022daformer} experiments, we consider a SegFormer equivalent version of M3L~\cite{maheshwari2023m3l} (with an MiT-B5~\cite{xie2021segformer} backbone). We test RGB and RGB+D models trained under  $(h=35, \theta=45^{\circ})$ (moderate viewpoint) conditions on \csim and find that incorporating additional Depth observations can substantially improve recognition performance. This demonstrates that images in \csim can be used to train multimodal scene-recognition models.
\section{Conclusion}
\label{sec:conclusion}
\par\noindent
We introduce \csim, a large-scale densely-annotated dataset of synthetic aerial scene images curated from unmanned aerial vehicle (UAV) perspectives. \csim images are generated using \crl by situating an agent aerially and procedurally tele-operating it through the scene to capture frames with semantic, instance, and depth annotations. Our careful curation process ensures that \csim images span across diverse weather, daytime, map, height, and pitch conditions, with accompanying metadata that enables reproducing the same viewpoint (spatial coordinates and perspective) under differing conditions.
\par\noindent
Through our experiments, we demonstrate that: (1) \csim-trained models generalize well to real-world settings, (2) \csim augments labeled real-world data in low-shot scenarios, (3) \csim serves as a diagnostic tool for assessing model sensitivity to varied conditions, and (4) incorporating additional sensors like depth enhances multi-modal aerial scene understanding.
\par\noindent
We aim to enhance \csim with improved realism, additional anticipated edge cases, and support for 3D perception tasks aligning with advancements in our simulator (additional details in Sec.~\ref{sec:future work} of appendix) 
We have publicly released the dataset and associated generation code and hope that our experimental findings encourage further research using \csim for aerial scenes. 
\par \noindent
\textbf{Acknowledgements.} We would like to thank Sean Foley for his contributions to the early efforts and discussions of this project. This work has been partially sponsored by NASA University Leadership Initiative (ULI) \#80NSSC20M0161, ARL, and NSF \#2144194.

\bibliographystyle{splncs04}
\bibliography{main}
\newpage
\appendix
\section{Overview}
The appendix is organized as follows. In Sec.~\ref{sec:dataset}, we provide more details on different aspects of dataset -- including the procedural image curation algorithm, algorithm to ensure appropriate representation of \texttt{human} instances, brief descriptions of all the classes in \csim and comparisons between data distribution of \csim and real datasets. Then, we describe experimental details in Sec.~\ref{sec:experiment} -- class-merging schemes used for our \csim$\rightarrow$Real transfer experiments, train / val / test splits and experimental details for \csim diagnostic setup to probe model vulnerabilities. Sec.~\ref{sec:results} provides more quantitative and qualitative experimental results.

\section{\csim Details}

\begin{algorithm}[t]
\caption{\csim ImgGen ($z$, $\theta$, \texttt{FoV}, $H$, $W$)}
\label{algo:image_gen_algo}
\begin{algorithmic}[1]
\State \textcolor{blue}{\# Initialize key \crl parameters}
\State \textbf{Input:} $z$ (height), $\theta$ (pitch), \texttt{FoV}, $H$, $W$
\State \textcolor{blue}{\# Initialize auxiliary \crl parameters}
\State \textbf{Initialize:} MB$\gets$Off \Comment{\textcolor{blue}{Turn off motion blur}}
\State \textbf{Initialize:} Post-process RGB$\gets$True \Comment{\textcolor{blue}{Turn on RGB post processing}}
\State \textbf{Dataset:} $D = \{\cdot\}$
\State \textcolor{blue}{\# Town and variation vocabulary}
\State $\mathbb{T}=\{T_i\}_{i=1}^M$ (Towns), $\mathbb{V}=\{V_i\}_{i=1}^N$ (Vars)
\For{$T_i\in\mathbb{T}$}
    \For{$V_j\in\mathbb{V}$}
        \State \textcolor{blue}{\# Initialize \crl scene}
        \State $E\gets$\crl$_{\text{init}}(T_i, V_j)$
        \State \textcolor{blue}{\# Position Camera}
        \State Init\_Sensor ($E$, $z$, $\theta$, \texttt{FoV}, $H$, $W$, MB)
        \State \textcolor{blue}{\# Spawn pedestrians, vehicles, etc.}
        \State Spawn\_Actors ($E$)
        \State \textcolor{blue}{\# Initialize Movement Steps}
        \State $\Delta_{\text{step}} \gets \Delta$, $N_{\text{steps}}\gets N$
        \For{$k\in N_{\text{steps}}$}
            \State \textcolor{blue}{\# Sample Frame}
            \State $I\gets$Sample\_Frame ($E$)
            \State \textcolor{blue}{\# Get pixel-level annotations}
            \State $I_{\text{anno}}\gets$Get\_Anno ($E$,$I$)
            \State \textcolor{blue}{\# Get metadata}
            \State $I_{\text{meta}}\gets$Get\_Meta ($E$,$I$)
            \State \textcolor{blue}{\# Append to dataset}
            \State $D\gets D \cup \{ (I, I_{\text{anno}}, I_{\text{meta}}) \}$
            \State \textcolor{blue}{\# Move camera by a fixed distance}
            \State Move\_Camera ($E$, $\Delta_{\text{step}}$)
        \EndFor
    \EndFor
\EndFor
\State \textbf{Return:} $D$ (\csim data) \Comment{\textcolor{blue}{Gathered Images}}
\end{algorithmic}
\end{algorithm}

\begin{algorithm}[t]
\caption{HumanSpawn ($x,y$, $d$, $p_{\text{gen}}$)}
\label{algo:human_spawn_algo}
\begin{algorithmic}[1]
\State \textcolor{blue}{\# Initialize parameters}
\State \textbf{Input:} $x,y$ (camera position), $d$ (distance between spawned instances), $p_{\text{gen}}$ (spawn probability)
\State \textcolor{blue}{\# Spawn locations}
\State $D_{\text{spawn}} = \{ \cdot\}$
\State \textcolor{blue}{\# Get candidate positions in front of the camera}
\State $\{(x, y)\}_{\text{front}}\gets$get\_loc($x$, $y$, $d$) \Comment{\textcolor{blue}{Current Lane}}
\State \textcolor{blue}{\# Get candidate positions left of camera}
\State $\{(x, y)\}_{\text{left}}\gets$get\_loc($x-\Delta_{\text{left}}$, $y$, $d$) \Comment{\textcolor{blue}{Left Lane}}
\State \textcolor{blue}{\# Get candidate positions right of camera}
\State $\{(x, y)\}_{\text{right}}\gets$get\_loc($x+\Delta_{\text{right}}$, $y$, $d$) \Comment{\textcolor{blue}{Right Lane}}
\State $D_{\text{spawn}}\gets \{(x, y)\}_{\text{front}}\cup\{(x, y)\}_{\text{left}}\cup\{(x, y)\}_{\text{right}}$
\For{$(x,y)\in D_{\text{spawn}}$}
    \If{\texttt{random()}$\leq p_{\text{gen}}$}
        \State \textcolor{blue}{\# Spawn \texttt{human}}
        \State spawn\_human()
    \EndIf
\EndFor
\end{algorithmic}
\end{algorithm}

\subsection{Image Generation Algorithms}
\label{sec:dataset}

We curate \csim using \crl~\cite{dosovitskiy2017carla} $0.9.14$ simulator. The generation process broadly consists of two key steps: (1) positioning the agent camera in an aerial perspective and (2) procedurally guiding the agent within the scene to capture images. We accomplish the first by mimicking a UAV perspective in \crl~\cite{dosovitskiy2017carla} by positioning the ego vehicle (with RGB, semantic, depth and instance segmentation sensors) based on specified altitude ($h$) and pitch ($\theta$) values to generate aerial views. This also requires setting other scene information like -- town, weather, and daytime. The camera $\texttt{FoV}=110^{\circ}$ (field of view) and $H\times W$ = $2160 \times 1440$ (image resolution) are also set. 
\par\noindent
To maintain the adequate representation of tail classes, especially for \texttt{humans} we incorporate structured spawning of \texttt{humans} using Algo.~\ref{algo:human_spawn_algo}. \crl \cite{dosovitskiy2017carla} has a limit on the number of actors that can be spawned in a scene, which depends on factors such as the type and size of the town, number of lanes to spawn vehicles, and amount of sidewalk area. To overcome this limitation, we decided to bring the actors into the field of view of the camera instead of having them spread out in the scene. We developed an algorithm to find all the points to spawn pedestrians in the field of view using the camera location and spawn them like vehicles on roads. After taking a snapshot, we destroy the spawned pedestrians and repeat the process. Manual spawning not only increases the number of \texttt{human} instances and their proportion but also aligns their placement with real-world settings. 
\par\noindent
As \csim images are curated by teleoperating over entire maps (rural or urban) across multiple layouts that differ substantially in class distributions (Fig. 4 left in the paper). Since these frames are stored with corresponding geographical (layout identifier) and positional (spatial location) metadata, filtering data splits to avoid overlap across physical layout regions is always possible.
\par\noindent
The steps involved in manual spawning instances of \texttt{humans} are summarized below:
\begin{packed_enumerate}
    \item[1] Specify maximum number of \texttt{human}s to be spawned
    $N_{\text{max}}$
    \item[2] Get camera position $(x,y,z)$, set a pre-defined distance $d$ to check for spawnable locations and execute the subroutine in Algo.~\ref{algo:human_spawn_algo}. This will place the actors in the field of view till a junction or the next driving lane.
    \item[3] If at a junction, obtain the left and the right waypoints for every retrieved location at a distance of $d$ to get the list of waypoints. \looseness=-1
    \item[4] Using the waypoint from the end of the current lane, generate waypoints for the new main, left and right lanes by repeating the previous steps.
    \item[5] Repeat the above steps till $N_{\text{humans}}\leq N_{\text{max}}$
\end{packed_enumerate} 
\noindent
Once positioned, the agent is translated by fixed amounts to traverse the scene and capture images from various viewpoints. Initially, we generate $70$ datapoints for each of the $8$ town variations under ClearNoon conditions using the baseline $h=35, \theta=45^{\circ}$ setting. Subsequently, following 
the traversal algorithm (Algo.~\ref{algo:image_gen_algo}),
we re-generate these datapoints across $5$ weather conditions and $12$ height/pitch variations, 
resulting in $70\times8\times5\times12=33,600$ images.

\subsection{Class Descriptions}
\label{sec:class}
We provide semantic, instance, and depth annotations for every image in \csim. \csim provides dense semantic annotations for $28$ classes. These are:
\begin{packed_itemize}
\item \texttt{unlabeled}: elements/objects in the scene that have not been categorized in \crl
\item \texttt{other}:  uncategorized elements
\item \texttt{building}: includes houses, skyscrapers, and the elements attached to them.
\item \texttt{fence}:  wood or wire assemblies that enclose an area of ground
\item \texttt{pedestrian}: humans that walk 
\item \texttt{pole}: vertically oriented pole and its horizontal components if any
\item \texttt{roadline}: markings on road.
\item \texttt{road}: lanes, streets, paved areas on which cars drive
\item \texttt{sidewalk}: parts of ground designated for pedestrians or cyclists
\item \texttt{vegetation}: trees, hedges, all kinds of vertical vegetation (ground-level vegetation is not included here).
\item \texttt{cars}: cars  \looseness=-1
\item \texttt{wall}: individual standing walls, not part of buildings 
\item \texttt{traffic sign}: signs installed by the 
state/city authority, usually for traffic regulation
\item \texttt{sky}: open sky, including clouds and sun
\item \texttt{ground}: any horizontal ground-level structures that do not match any other category
\item \texttt{bridge}: the structure of the bridge
\item \texttt{railtrack}: rail tracks that are non-drivable by cars
\item \texttt{guardrail}: guard rails / crash barriers
\item \texttt{traffic light}: traffic light boxes without their poles.
\item \texttt{static}: elements in the scene and props that are immovable.
\item \texttt{dynamic}: elements whose position is susceptible to change over time.
\item \texttt{water}: horizontal water surfaces
\item \texttt{terrain}: grass, ground-level vegetation, soil or sand
\item \texttt{rider}: humans that ride/drive any kind of vehicle or mobility system
\item \texttt{bicycle}: bicycles in scenes
\item \texttt{motorcycle}: motorcycles in scene
\item \texttt{bus}: buses in scenes
\item \texttt{truck}: trucks in scenes
\end{packed_itemize}

\subsection{Depth \& Instance Segmentation}
We also provide depth and instance segmentation annotations along with semantic segmentation annotations. Both the depth and instance segmentation sensors are mounted alongside the RGB camera and semantic segmentation sensors. Depth is stored in the $LogarithmicDepth$ format which provides better results for closer objects. We also provide depth-aided semantic segmentation results (training details in Sec. \ref{sec:training_details}).

\subsection{\csim vs Real Characteristics}
\begin{figure}[htbp]
    \begin{minipage}[b]{0.48\textwidth}
        \centering
        \footnotesize
        \setlength{\tabcolsep}{2.5pt}
        \resizebox{\textwidth}{!}{
            \begin{tabular}{lccc}
                \toprule
                \multicolumn{4}{c}{\textbf{FID Score ($\downarrow$)}} \\
                & \ua & \aero & \icg \\
                \midrule
                \texttt{1} \syn & $23.07$ & $19.56$ & $9.14$ \\
                \texttt{2} \csim & $\textbf{8.78}$ & $\textbf{7.66}$ & $\textbf{5.77}$ \\
                \bottomrule
            \end{tabular}
        }

        \captionsetup[sub]{labelformat=parens} 
        \subcaption{\textbf{Syn vs. Real Perceptual Similarity}}
        \label{table:fid}
    \end{minipage}%
    \hfill
    \begin{minipage}[b]{0.48\textwidth}
        \centering
        \includegraphics[width=\linewidth]{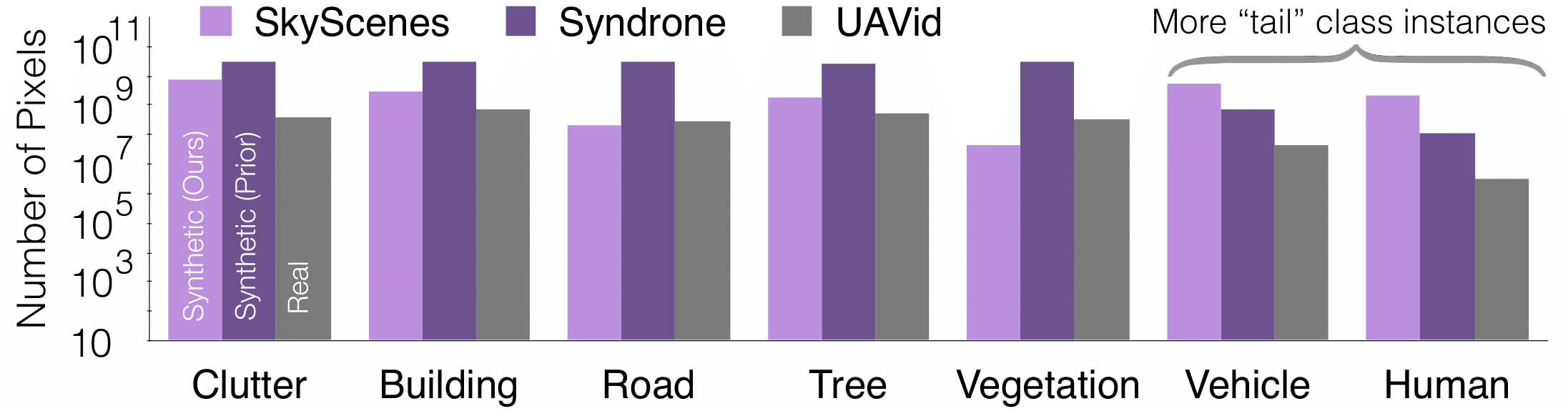}
        \captionsetup[sub]{labelformat=parens} 
        \subcaption{\textbf{\csim Per-Class Pixel Counts.}}
        \label{fig:raw_pixel_counts}
    \end{minipage}
    \caption{\textbf{\csim vs Real Characteristics.} (a) Comparing perceptual similarity using FID scores calculated between real datasets (\ua, \aero, \icg) and synthetic datasets (\syn, \csim). (b) We compare the number of densely annotated pixels per-class for \csim (ours), Syndrone (another synthetic aerial dataset), and UAVid (real aerial dataset). We can see how compared to both synthetic and real counterparts, \csim provides better representation of tail classes (\texttt{vehicles}, \texttt{humans}).}
    \label{fig:comparison}
\end{figure}

\noindent
In Fig.~\ref{table:fid}, we compare perceptual similarity (in terms of FID~\cite{heusel2018ganstrainedtimescaleupdate}) of synthetic images from \csim and \syn~\cite{rizzoli2023syndrone} with real datasets (\ua~\cite{lyu2020uavid}, \aero~\cite{aeroscapes}, \icg~\cite{drone-dataset}). We find \csim images are closer to real data distributions than \syn~\cite{rizzoli2023syndrone}. For semantic comparisons, we show class-distribution comparisons (per-class pixel frequencies) between \csim, \syn~\cite{rizzoli2023syndrone} and \ua~\cite{lyu2020uavid} in Fig~\ref{fig:raw_pixel_counts}. \csim images display higher distribution of tail classes(\texttt{vehicles, humans}) compared to \syn~\cite{rizzoli2023syndrone}, and shows better alignment with the real data distribution (\ua~\cite{lyu2020uavid}). 
\section{Experiment Details}
\label{sec:experiment}
\subsection{Class Merging Details}
\label{sec:Merging}

\begin{minipage}[t]{0.5\textwidth}
\resizebox{1.0\columnwidth}{!}{
\begin{tabular}{l@{\hskip -5pt}cccccc}
\toprule

\textbf{Syn $\rightarrow$ \ua} & \textbf{\ua}  & \textbf{\csim} & \textbf{\syn} & \textbf{\valid} & \textbf{\synth} \\
\midrule

\multirow{24}{*}{\texttt{1} \textbf{clutter}} & clutter & unlabelled & unlabelled & background & sky \\
&&sidewalk&sidewalk&pavement&footpath\\
&&fence&fence&fence&\\
&&bridge&bridge&bridge&\\
&&water&water&water\\
&&traffic light & traffic light& traffic light &\\
&&other&other&sign&  \\
&& traffic sign & traffic sign&land& \\
&&rail track&rail track&tunnel &\\
&&guard rail & guard rail & pool &\\
&&static&static&stones&\\
&&dynamic&dynamic&pierruble\\
&&ground&ground&chair&\\
&&sky&sky&ice&\\
&&pole&pole&ship&\\
&&&&plane&\\
&&&&harbor&\\
&&&&lamp&\\
&&&&bus stop&\\
&&&&powerline&\\
&&&&garbage bin&\\
&&&&other low obstacle&\\
&&&&other high obstacle&\\

\midrule
\multirow{2}{*}{\texttt{2} \textbf{building}} & building & building & building & building & building \\
&&wall&wall&&wall\\
\midrule
\multirow{2}{*}{\texttt{3} \textbf{road}} & road & road & road & road& road  \\
&&roadline&roadline & &\\
\midrule
\texttt{4} \textbf{vegetation} &vegetation&vegetation&vegetation&tree & tree \\
\midrule
\texttt{5} \textbf{low vegetation} &terrain&terrain&terrain & other plant & vegetation \\
\midrule
\texttt{6} \textbf{person} & person & pedestrian & pedestrian & person & \\
&&rider&rider \\
&&motorcycle&motorcycle\\
&&bicycle&bicycle\\
\midrule
\multirow{4}{*}{\texttt{7} \textbf{vehicle}} & car & car & car &small vehicle & car \\
&&truck&truck & large vehicle\\
&&bus&bus\\
&&&train\\
\bottomrule
\end{tabular}
}

\captionof{table}{\small\textbf{Class merging scheme for evaluating Syn $\rightarrow$ \ua experiments} The first column is the final set of merged classes we use for Syn $\rightarrow$\ua evaluation, the second column is the original \ua~\cite{lyu2020uavid}  classes, the third column is the original \csim classes, the fourth and fifth column are original \syn~\cite{rizzoli2023syndrone} and \valid~\cite{valid} classes respectively and the last column is the original \synth~\cite{synthaer} classes. Each row indicates all the classes from \ua, \csim, \syn~\cite{rizzoli2023syndrone}, \valid~\cite{valid}, and \synth~\cite{synthaer} that were merged and correspond to the final Syn$\rightarrow$\ua class in the first column }
\label{table:uavid_sky_class}

\end{minipage}
\begin{minipage}[t]{0.5\textwidth}
\resizebox{1.0\columnwidth}{!}{
\begin{tabular}{lccccc}
\toprule

\textbf{Common Scheme} & \textbf{\ua}  & \textbf{\aero} & \textbf{\icg} & \textbf{\csim}\\
\midrule

\multirow{13}{*}{\texttt{1} \textbf{clutter}} & clutter & background & unlabeled & unlabeled \\
&&sky&dirt&sky\\
&&&grass&pole\\
&&&gravel&sidewalk\\
&&&water&traffic sign\\
&&&pool&other\\
&&&rock&ground\\
&&&fence-pole&guard rail\\
&&&dog&traffic light\\
&&&ar-marker&static\\
&&&obstacle&dynamic\\
&&&conflicting&water\\
&&&&terrain\\
\midrule
\multirow{2}{*}{\texttt{2} \textbf{road}} & road & road & paved area & road \\
&&&&roadline\\
&&&&rail track\\
\midrule
\multirow{3}{*}{\texttt{3} \textbf{nature}} & vegetation & vegetation & vegetation & vegetation \\
&tree&&tree&\\
&&&bald tree&\\
\midrule
\multirow{4}{*}{\texttt{4} \textbf{person}} & human & person & person & pedestrian \\
&&bicycle&bicycle&bicycle\\
&&&&rider\\
&&&&motorcycle\\
\midrule
\multirow{4}{*}{\texttt{5} \textbf{vehicle}} & car & vehicle & car & car \\
&&&&truck\\
&&&&bus\\
\midrule
\multirow{4}{*}{\texttt{5} \textbf{construction}} & building & building & roof & building \\
&&&wall&wall\\
&&&fence&fence\\
&&&window&bridge\\
&&&door&\\
\bottomrule
\end{tabular}
}

\captionof{table}{\small\textbf{Common class merging scheme across all real datasets and \csim } The first column is the final set of merged classes we use for evaluations, the second column is the original \ua~\cite{drone-dataset}  classes, the third column is the original \aero~\cite{aeroscapes} classes, the fourth column is the original \icg~\cite{drone-dataset} classes and the last column is the original \csim classes. Each row indicates all the classes from \ua, \aero, \icg, and \csim, that were merged and correspond to the final common class in the first column }
\label{table:common_sky_class}
    
\end{minipage}

\par\noindent
\begin{minipage}[t]{0.5\textwidth}
\resizebox{1.0\columnwidth}{!}{
\begin{tabular}{lcccc}
\toprule
\textbf{Syn $\rightarrow$ \aero} & \textbf{\aero}  & \textbf{\csim} & \textbf{\syn} \\
\midrule
\multirow{13}{*}{\texttt{1} \textbf{background}} & background & unlabelled & unlabelled \\
& drone&other&other \\
&boat & traffic sign & traffic sign \\
&animal&rail track&rail track\\
&obstacle&guard rail & guard rail\\
&&traffic light & traffic light\\
&&static&static\\
&&dynamic&dynamic\\
&&ground&ground\\
&&sidewalk&sidewalk\\
&&terrain&terrain\\
&&water&water\\
&&pole&pole\\
\midrule
\multirow{2}{*}{\texttt{2} \textbf{bicycle}} & bike & bicycle & bicycle  \\
&&motorcycle&motorcycle\\
\midrule
\texttt{3} \textbf{person} & person & pedestrian & pedestrian \\
&&rider&rider \\
\midrule
\multirow{4}{*}{\texttt{4} \textbf{vehicle}} & car & car & car \\
&&truck&truck\\
&&bus&bus\\
&&&train\\
\midrule
\texttt{5} \textbf{vegetation} &vegetation&vegetation&vegetation \\
\midrule
\multirow{4}{*}{\texttt{6} \textbf{building}} & construction & building & building \\
&&wall&wall\\
&&fence&fence\\
&&bridge&bridge\\
\midrule
\multirow{2}{*}{\texttt{7} \textbf{road}} & road & road & road  \\
&&roadline&roadline\\
\midrule
\multirow{1}{*}{\texttt{8} \textbf{sky}}&sky & sky & sky  \\
\bottomrule
\end{tabular}
}

\captionof{table}{\small\textbf{Class merging scheme for evaluating Syn $\rightarrow$ \aero experiments} The first column is the final set of merged classes we use for Syn $\rightarrow$\aero evaluation, the second column is the original \aero~\cite{aeroscapes}  classes, the third column is the original \csim classes and the last column is the original \syn~\cite{rizzoli2023syndrone} classes. Each row indicates all the classes from \aero, \csim, and \syn that were merged and correspond to the final Syn$\rightarrow$\aero class in the first column }
\label{table:aero_sky_class}
\end{minipage}
\begin{minipage}[t]{0.5\textwidth} 
    \centering

\resizebox{1.0\columnwidth}{!}{
\begin{tabular}{lcccccc}
\toprule

\textbf{Syn $\rightarrow$ \icg} & \textbf{\icg}  & \textbf{\csim} & \textbf{\syn} & \textbf{\valid} &\textbf{\synth} \\
\midrule

\multirow{22}{*}{\texttt{1} \textbf{other}} & obstacle & unlabelled & unlabelled & background & sky \\
& dog&other&other & animal &  \\
&conflicting & traffic sign & traffic sign & sign &  \\
&ar-marker&sky&sky&ice\\
&unlabelled & bridge & bridge & bridge\\
&&gurad rail & guard rail & stones\\
&&traffic light & traffic light & traffic light\\
&&static&static & pierruble\\
&&dynamic&dynamic& garbage bin\\
&&&&plane&\\
&&&&harbor&\\
&&&&land&\\
&&&&chair&\\
&&&&ship&\\
&&&&lamp&\\
&&&&tunnel&\\
&&&&bus stop&\\
&&&&powerline&\\
&&&&garbage bin&\\
&&&&other low obstacle&\\
&&&&other high obstacle&\\
\midrule
\texttt{2} \textbf{fence} & fence & fence & fence& fence  \\
\midrule
\texttt{3} \textbf{pole} & fence pole & pole & pole &&\\
\midrule
\multirow{2}{*}{\texttt{4} \textbf{vegetation}} & tree & vegetation & vegetation & tree & tree \\
&bald-tree&&&other plant & vegetation\\
\midrule
\multirow{4}{*}{\texttt{5} \textbf{building}}&wall&wall&wall&&wall \\
& roof&building&building&building&\\
&door&&&&\\
&window&&&&\\
\midrule
\multirow{2}{*}{\texttt{6} \textbf{water}} & water & water & water&water& \\
&pool&&pool&\\
\midrule
\multirow{2}{*}{\texttt{7} \textbf{bicycle}} & bicycle & bicycle & bicycle&&  \\
&&motorcycle&motorcycle&&\\
\midrule
\multirow{4}{*}{\texttt{8} \textbf{vehicle}}&car & car & car& small vehicle & vehicle  \\
&&truck&truck&large vehicle& \\
&&bus&bus\\
&&&train\\
\midrule
\multirow{2}{*}{\texttt{9} \textbf{person}}&person& pedestrian&pedestrian&person& \\
&&rider&rider\\
\midrule
\multirow{5}{*}{\texttt{10} \textbf{paved area}}&paved area & road&road&road&road \\
&&sidewalk&sidewalk& pavement&footpath\\
&&ground & ground\\
&&roadline&roadline\\
&&rail track & rail track\\
\midrule
\multirow{5}{*}{\texttt{11} \textbf{terrain}}&rocks&terrain&terrain&land  \\
&gravel&&\\
&dirt&&\\
&vegetation&&\\
&grass&&\\
\bottomrule
\end{tabular}
}

\captionof{table}{\small\textbf{Class merging scheme for evaluating Syn $\rightarrow$ \icg experiments} The first column is the final set of merged classes we use for Syn $\rightarrow$\icg evaluation, the second column is the original \icg~\cite{drone-dataset}  classes, the third column is the original \csim classes, the fourth and fifth column are original \syn~\cite{rizzoli2023syndrone} and \valid~\cite{valid} classes respectively and the last column is the original \synth~\cite{synthaer} classes. Each row indicates all the classes from \icg, \csim, \syn, \valid, and \synth that were merged and correspond to the final Syn$\rightarrow$\icg class in the first column }
\label{table:icg_sky_class}

\end{minipage}
\textbf{$\triangleright$ Synthetic$\rightarrow$Real. }As noted in Sec.$4$ of the main paper, since different real-world datasets have different class vocabularies and definitions, for our Synthetic\\$\to$Real semantic segmentation experiments, we adapt the class-vocabulary of the synthetic source dataset (\csim, \syn, \valid, \synth) to that of the target real dataset (\ua, \aero, \icg). This is done using a class-merging scheme based on the class-vocabularies and after visually inspecting dataset annotations. We provide the class-merging schemes used for the synthetic datasets (\csim, \syn~\cite{rizzoli2023syndrone}, \valid~\cite{valid} and \synth~\cite{synthaer}) across real counterparts \ua~\cite{lyu2020uavid} (in Table.~\ref{table:uavid_sky_class}), \aero~\cite{aeroscapes} (in Table.~\ref{table:aero_sky_class}) and \icg~\cite{drone-dataset} (in Table.~\ref{table:icg_sky_class}). We also include a coarser common merging scheme for \csim and all the real datasets \ua, \aero, and \icg in Table.~\ref{table:common_sky_class}. 
\par \noindent
\textbf{$\triangleright$ \csim Diagnostic Experiments.} To assess the sensitivity of trained models to different factors -- weather, time of day, height, pitch, \etc -- we train models on different \csim variations and evaluate them on held-out-variati-ons. For these experiments, we reduce the \csim vocabulary to a reasonable subset of $20$ classes (consistent with the widely used Cityscapes~\cite{cordts2016cityscapes} palette) -- \texttt{road}, \texttt{sidewalk}, \texttt{building}, \texttt{wall}, \texttt{fence}, \texttt{pole}, \texttt{traffic light}, \texttt{traffic sign}, \texttt{vegetation}, \texttt{water}, \texttt{sky}, \texttt{pedestrian}, \texttt{rider}, \texttt{cars}, \texttt{truck}, \texttt{bus}, \texttt{roadline}, \texttt{motorcycle}, \texttt{bicycle} and an \texttt{ignore} class.

\subsection{Training, Validation and Test Splits}
\label{sec:splits}
\textbf{\csim.} For each 
$(h, \theta)$ combination,
\csim has a total of $2800$ datapoints (frames) which are distributed evenly across each of the $8$ town layouts and $5$ weather and daytime conditions. We use $80$\% ($2240$ images) of these data points for training models, 
and remaining
$10$\% ($280$ images) each for validation and testing. While creating train, val and test splits, we collect equal number of samples from each town by dividing each town-specific traversal sequence into $3$ segments: the initial $80$\% for training, the next $10$\% for testing, and the final $10$\% of the segment for validation. Moreover, within each split, we ensure that every viewpoint is accompanied by its $60$ different variations across weather, daytime, height, and pitch settings. This safeguards against any potential cross-contamination across different splits while ensuring fair representation and equal distributions of all variations.
 \\
\textbf{\syn.} \syn has $3000$ images per town (across $8$ \crl towns) for each of the $3$ $(h, \theta)$ combinations,
resulting in a total of  $3000\times8=24,000$ images per $(h, \theta)$ combination. We use $20,000$ of these data points for training models, with $4000$ kept aside for testing. The data points selected for training and testing are kept consistent with the one reported in \syn~\cite{rizzoli2023syndrone}. \\ 
\textbf{\valid.}\valid has a total of $6690$ images spread across $3$ different height variations and $6$ different layout and daytime conditions all from the nadir perspective. For the $3$ $(h, \theta)$ combination, $(h=100m,\theta = 90^{\circ}),(h=50m,\theta = 90^{\circ})$ and $(h=20m,\theta = 90^{\circ})$, the total images are $1734$ , $2158$ and $2798$ respectively which are split into $80\%$ for training and $10\%$ each for validation and testing. The data points selected for training and testing are kept consistent with the one reported in \valid~\cite{valid}  \\
\textbf{\synth.} The split of the dataset and the data points selected for training and testing are consistent with the one reported in \synth~\cite{synthaer} with $435$ images for training, $132$ for validation and $198$ for testing. 
\subsection{\csim Diagnostic Experiments}
\textbf{Weather \& Daytime Variations.} For each weather variation we sample evenly across the $8$ towns and $9$  $(h, \theta)$ combinations (excluding $\theta=0^{\circ}$),
resulting in a total of $70\times9\times8=5040$ images. We use $80$\% ($4032$ images) of these  for training models, and remaining $10$\% ($504$ images) each for validation and testing. We evaluate the model on the same weather variation it was trained on and select the model with the best mIoU score for further evaluations on other weather variations. \\
\textbf{Town Variations.} For each variation in town we sample evenly across $5$ weather and daytime conditions and $9$ height and pitch variations(excluding pitch=$0^{\circ}$ variations), resulting in a total of $70\times5\times9=3150$ images per town variation. Out of these, $80$\% ($2520$ images) is allocated for training models, with $10$\% ($315$ images) each for validation and testing. We evaluate the model on the same town variation it was trained on and select the model with the best mIoU score for further evaluations on other town variations.\\
\textbf{Height \& Pitch Variations.} For each $(h, \theta)$ variation, we evenly sample across $8$ towns and $5$ weather and daytime conditions, resulting in a total of $70\times8\times5=2800$ images per height and pitch variation. Out of these, $80$\% ($2240$ images) are allocated for training models, and $10$\% ($280$ images) each for validation and testing. We evaluate the model on the same height \& pitch setting it was trained on and select the model with the best mIoU score for further evaluations on other height \& pitch settings.

\subsection{Training Details}
\label{sec:training details}
\textbf{Semantic Segmentation.} For our semantic segmentation experiments we use both CNN -- DeepLabv2~\cite{chen2017deeplab} (ResNet-101~\cite{he2016deep} backbone) -- and vision transformer  -- 1. DAFormer~\cite{hoyer2022daformer} (with HRDA~\cite{hoyer2022hrda} source training; MiT-B5~\cite{xie2021segformer} backbone) and 2. Rein DINOv2~\cite{wei2024stronger} -- based semantic segmentation architectures. Note that we utilize only the DAFormer architecture to perform source-only training for our experiments. Following~\cite{hoyer2023mic}, we enable rare class sampling ~\cite{hoyer2022daformer} and use Imagenet feature-distance for our thing classes during training. 
DeepLabv2~\cite{chen2017deeplab} and DAFormer~\cite{hoyer2022daformer} are trained using the AdamW~\cite{loshchilov2019decoupled} optimizer coupled with a polynomial learning rate scheduler with an initial learning rate of $6\times10^{-5}$. For our fine-tuning experiments, we use an initial learning rate of $6\times10^{-6}$. For Rein~\cite{wei2024stronger}, we use initial learning rate of $3\times10^{-5}$ and for fine-tuning we use $3\times10^{-6}$. Each model is trained for $40$k iterations with a batch size of $4$.

\par \noindent
\textbf{Depth-Aided Semantic Segmentation.}
For our depth-aided semantic segmentation experiments in Sec. 4.3 of the main paper, similar to DAFormer~\cite{hoyer2022daformer}, we employ a SegFormer~\cite{xie2021segformer} equivalent version of M3L~\cite{maheshwari2023m3l} (multimodal segmentation network) with an MiT-B5~\cite{xie2021segformer} backbone. We initialize the network with ImageNet-1k pre-trained checkpoints. For M3L Linear Fusion, we use $\alpha=0.8$. We use AdamW~\cite{kingma2014adam} optimizer and train on a batch size of $4$ for $50$ epochs. We use a learning rate of $10^{-4}$ for the encoder and $3\times10^{-4}$ for the decoder with a momentum of $0.9$ and weight decay of $10^{-4}$. We set the polynomial decay of power $0.9$. We train both RGB and RGB+D models with 
complete supervision for $(h=35m, \theta=45^{\circ})$ (moderate viewpoint) conditions on \csim.
\label{sec:training_details}
\subsection{Evaluation Details}
\par\noindent
Due to memory constraints, in addition to heavily parameterized models, our GPUs were unable to fit images larger than $1280\times 720$. Hence for high-resolution datasets like \ua, we use the trained model to make separate predictions on $4$  equally sized slightly-overlapping crops (overlap of $20$ pixels) of the  of the real image 
and stitch crop predictions to obtain the overall image prediction. Similarly, for \icg, we obtain overall image prediction using such crop predictions.
\section{Results}

\label{sec:results}
\subsection{Ground vs Aerial Performance Gap}
Recent advancements in synthetic-to-real (syn-to-real) generalization have significantly closed the performance gap in ground imagery to about \textbf{20\%}, unlike in aerial imagery where the gap remains as high as \textbf{50\%}. In our study, we evaluated Rein DINOv2~\cite{wei2024stronger} across ground-view settings using GTAV$\to$Cityscapes and aerial-view settings using \csim$\to$ICG. For ground views, the gap between GTAV$\to$Cityscapes and Cityscapes~\cite{cordts2016cityscapes} mIoU was \textbf{16.29\%}, demonstrating a mIoU of $66.7$ compared to $82.99$. Aerial views showed a larger gap, with \csim$\to$ICG mIoU at $25.91$ versus ICG mIoU at $76.44$, resulting in a \textbf{50.53\%} gap. This highlights the need for focused efforts to address the larger performance discrepancies observed in aerial imagery.
\begin{table}[htbp]
\footnotesize
\centering
\setlength{\tabcolsep}{2.5pt}
\begin{center}
\resizebox{\columnwidth}{!}{
\begin{tabular}{lcccccccccc}
\toprule
{\textbf{\textit{h}(m)}} &\textit{$\bm{\theta(^{\circ})}$} & \multicolumn{8}{c}{\textbf{Synthetic$\rightarrow\ua$ mIoU ($\uparrow$)}}\\ 
& &Clutter &Building &Road &Tree &Low Vegetation &Human &Vehicle & Avg  \\ 
\midrule
\textbf{\csim}\\
\midrule
\texttt{1}  $15$ &$0$& $30.23$ &$70.24$ &$43.91$ &$52.20$ &$6.24$ &$10.33$ &$43.45$ &$36.66$\\
\texttt{2}  $15$ &$45$& $23.36$ &$61.88$ &$43.10$ &$38.18$ &$11.79$ &$0.32$ &$3.76$ &$26.05$\\
\texttt{3}  $15$ &$60$& $22.15$ &$57.85$ &$39.31$ &$38.43$ &$5.35$ &$0.27$ &$3.72$ &$23.35$\\
\texttt{4}  $15$ &$90$& $27.01$ &$68.40$ &$41.92$ &$53.79$ &$17.95$ &$17.26$ &$42.80$ &$38.45$  \\
\midrule
\texttt{5}  $35$ &$0$& $31.66$ &$72.37$ &$38.65$ &$45.73$ &$12.97$ &$0.45$ &$23.63$ &$32.21$ \\
\rowcolor{blue!12}
\texttt{6}  $35$ &$45$& $36.44$ &$81.3$&$52.09$&$60.00$&$25.96$	&$10.21$& $63.64$&	$\bm{47.09}$  \\
\texttt{7}  $35$ &$60$& $28.17$ &$68.31$ &$44.96$ &$44.84$ &$15.32$ &$0.05$ &$8.81$&$30.06$  \\
\texttt{8}  $35$ &$90$& $28.88$&$76.49$ &$48.11$ &$57.88$ &$13.61$ &$7.98$ &$49.32$ &$40.07$ \\ 
\midrule
\texttt{9}  $60$ &$0$&$24.83$ &$66.37$ &$26.18$ &$39.58$ &$11.43$ &$0.01$ &$4.76$ &$24.74$\\ 
\texttt{10} $60$ &$45$& $27.32$ &$66.02$ &$38.29$ &$41.45$ &$11.72$ &$0.0$ &$5.25$ &$27.15$\\ 
\texttt{11} $60$ & $60$&$23.98$ &$62.62$&$32.72$&$41.00$&$17.55$&$0.00$&$6.34$&$26.32$ \\ 
\texttt{12} $60$ & $90$& $28.84$ &$75.03$ &$40.72$ &$54.27$ &$13.02$ &$1.16$ &$49.48$ &$37.50$ \\ 
\midrule
\textbf{\textbf{\syn}}\\
\midrule
\rowcolor{blue!12}
\texttt{13} $20$ &$30$&$36.20$ &$75.74$ &$48.71$ &$55.95$ &$28.75$ &$8.27$ &$42.52$ &$\underline{42.31}$\\ 
\texttt{14} $50$ & $60$& $31.13$ &$69.93$ &$48.87$ &$54.49$ &$27.71$ &$1.32$ &$36.06$ &$38.50$  \\ 
\texttt{15} $80$ & $90$& $28.89$ &$65.66$ &$42.05$ &$51.51$ &$32.16$ &$0.13$ &$28.39$ &$35.54$  \\ 
\bottomrule
\end{tabular}
}
\caption{\footnotesize\textbf{Models trained on \ua aligned \textit{(h,$\bm{\theta}$)}} display better generalization performance . We have trained both \csim and \syn on every subset of \textit{(h,$\theta$)} provided by the respective datasets}
\label{table:uavid_expert}
\end{center}
\end{table}
\begin{table}[htbp]
\footnotesize
\centering
\setlength{\tabcolsep}{2.5pt}
\begin{center}
\resizebox{\columnwidth}{!}{
\begin{tabular}{lccccccccccc}
\toprule
{\textbf{\textit{h}(m)}} &\textit{$\bm{\theta(^{\circ})}$} & \multicolumn{9}{c}{\textbf{Synthetic$\rightarrow\aero$ mIoU ($\uparrow$)}}\\ 
& &Background & Bicycle &Person&Vehicle &Vegetation &Building &Road & Sky & Avg  \\ 
\midrule
\textbf{\csim}\\
\midrule
\texttt{1}  $15$ &$0$& $32.39$ &$0.00$ &$3.45$ &$55.42$ &$56.08$ &$30.11$ &$15.4$ &$69.75$ &$32.81$\\
\texttt{2}  $15$ &$45$& $25.26$ &$0.0$ &$4.12$ &$5.72$ &$31.9$ &$11.56$ &$26.43$ &$7.23$ & $14.03$ \\
\texttt{3}  $15$ &$60$& $27.16$ &$0.00$ &$1.69$ &$11.81$ &$32.27$ &$22.48$ &$31.22$ &$0.21$ &$15.86$ \\
\texttt{4}  $15$ &$90$ &$28.53$ &$1.32$ &$30.35$ &$77.22$ &$53.09$ &$12.09$ &$11.43$ &$0.00$ &$26.75$ \\
\midrule
\texttt{5}  $35$ &$0$& $32.45$ &$0.00$ &$0.00$ &$17.65$ &$51.00$ &$42.03$ &$7.14$ &$75.03$ &$28.16$\\
\rowcolor{blue!12}
\texttt{6}  $35$ &$45$& $32.07$ & $0.8$	& $3.09$ & $80.99$ &$51.34$ &	$45.54$ &$23.64$ &$88.29$ &$\bm{40.72}$ \\
\texttt{7}  $35$ &$60$& $29.72$&$0.00$ &$0.00$ &$30.00$ &$45.54$ &$24.01$&$23.26$&$0.00$ &$19.07$ \\
\texttt{8}  $35$ &$90$& $30.62$ &$1.9$ &$2.62$ &$72.77$ &$55.56$ &$26.85$ &$17.34$ &$1.94$ &$26.2$ \\ 
\midrule
\texttt{9}  $60$ &$0$&$29.99$ &$0.0$ &$0.0$ &$1.05$ &$34.68$ &$42.11$ &$10.53$ &$49.63$ &$21.00$\\ 
\texttt{10} $60$ &$45$& $28.05$ &$0.0$ &$0.0$ &$7.14$ &$40.60$ &$22.38$ &$17.84$ &$5.16$ &$15.15$\\ 
\texttt{11} $60$ & $60$& $26.71$ &$0.0$ &$0.0$ &$0.53$ &$37.93$ &$19.95$ &$18.63$ &$0.80$ &$13.07$  \\ 
\texttt{12} $60$ & $90$& $31.08$ &$0.00$ &$0.11$ &$30.83$ &$58.10$ &$31.61$ &$17.76$ &$0.06$ &$21.19$ \\ 
\midrule

\textbf{\textbf{\syn}}\\
\midrule
\rowcolor{blue!12}
\texttt{13} $20$ &$30$&$32.32$ &$0.92$ &$0.77$ &$49.77$ &$54.42$ &$35.71$ &$6.89$ &$63.45$ &$\underline{30.53}$\\ 
\texttt{14} $50$ & $60$& $32.29$ &$0.99$ &$0.05$ &$29.41$ &$56.47$ &$39.59$ &$22.15$ &$5.36$ &$23.29$  \\ 
\texttt{15} $80$ & $90$&$30.40$ &$0.09$&$0.01$&$27.09$&$51.04$&$39.20$&$27.32$&$0.17$&$21.92$ \\ 
\bottomrule
\end{tabular}}

\caption{\footnotesize\textbf{Models trained on \aero aligned \textit{(h,$\bm{\theta}$)}} display better generalization performance . We have trained both \csim and \syn on every subset of \textit{(h,$\theta$)} provided by the respective datasets}
\label{table:aero_expert}
\end{center}

\end{table}
\begin{table}[htbp]
\footnotesize
\centering
\setlength{\tabcolsep}{2.5pt}
\begin{center}
\resizebox{\linewidth}{!}{
\begin{tabular}{lcccccccccccccc}
\toprule
{\textbf{\textit{h}(m)}} &\textit{$\bm{\theta(^{\circ})}$} & \multicolumn{12}{c}{\textbf{Synthetic$\rightarrow\icg$ mIoU ($\uparrow$)}}\\ 
& &other & fence &pole	&vegetation	&building&water	&bicycle	&vehicle &person	 &paved area &terrain& Avg & Avg$^{*}$	 \\ 
\midrule
\textbf{\csim}\\
\midrule
\texttt{1}  $15$ &$0$& $2.85$& $0.28$ &$0.04$	&$6.60$	&$29.94$	&$0.83$&$0.06$	&$0.45$	&$0.35$	&$27.46$	&$0.72$	&$6.33$	&$7.33$ \\
\texttt{2}  $15$ &$45$&$2.87$&$0.67$&$0.21$	&$6.91$	&$32.12$ &$3.25$ &$0.49$ &$5.03$ &$25.22$ &$56.69$&	$8.78$&$12.93$&$14.51$ \\
\texttt{3}  $15$ &$60$&$3.98$	&$0.00$	&$0.73$	&$5.19$	&$35.85$	&$0.66$	&$0.12$	&$4.95$	&$3.56$	&$57.51$	&$1.21$	&$10.34$	&$12.06$ \\
\rowcolor{blue!12}
\texttt{4}  $15$ &$90$&$3.74$&$1.45$&$2.51$	&$6.67$	&$46.72$	&$8.19$	&$2.21$	&$39.71$	&$45.89$	&$79.84$	&$6.04$	&$\bm{22.09}$	&$\bm{25.91}$   \\
\midrule
\texttt{5}  $35$ &$0$&$1.63$ &$0.00$	&$0.06$	&$4.86$	&$28.84$	&$2.77$	&$0.07$&	$0.04$	&$0.30$	&$29.70$	&$6.96$	&$6.84$	&$7.40$  \\
\texttt{6}  $35$ &$45$&$4.37$	&$0.39$	&$1.60$	&$4.90$	&$25.33$	&$4.98$	&$0.10$	&$1.35$	&$0.29$	&$68.37$	&$29.51$	&$12.83$	&$11.92$  \\
\texttt{7}  $35$ &$60$&$2.06$ &$0.01$	&$0.04$	&$6.15$	&$30.05$	&$9.43$	&$0.02$	&$0.23$	&$0.11$	&$54.99$	&$13.43$	&$10.59$	&$11.23$  \\
\texttt{8}  $35$ &$90$&$2.13$&$0.04$	&$0.64$	&$7.34$	&$28.07$	&$7.26$	&$0.09$	&$0.30$	&$0.11$	&$50.27$	&$11.11$	&$9.76$	&$9.63$ \\ 
\midrule
\texttt{9}  $60$ &$0$&$2.23$	&$0.00$	&$0.00$	&$8.30$	&$32.07$	&$1.72$	&$0.10$	&$0.03$	&$0.17$	&$11.58$	&$13.30$	&$6.32$	&$6.00$\\ 
\texttt{10} $60$ &$45$&$1.07$&$0.00$	&$0.02$&$7.71$&$30.03$&$4.73$&$0.17$	&$0.00$	&$0.22$	&$31.62$	&$8.87$	&$7.68$&	$8.28$\\ 
\texttt{11} $60$ & $60$&$0.81$	&$0.00$	&$0.02$	&$4.84$	&$26.59$	&$7.21$	&$0.16$	&$0.01$	&$0.16$& $51.34$	&$8.39$	&$9.05$	&$9.11$  \\ 
\texttt{12} $60$ & $90$&$1.32$	&$0.00$	&$0.25$	&$7.90$	&$27.64$	&$3.21$	&$0.04$	&$0.26$	&$0.22$	&$21.36$	&$6.65$	&$6.26$	&$6.76$ \\ 
\midrule
\textbf{\textbf{\syn}}\\
\midrule
\texttt{13} $20$ &$30$&$6.50$	&$0.10$	&$1.47$	&$5.47$	&$27.68$	&$16.07$	&$0.29$	&$11.89$	&$0.30$	&$62.47$	&$32.97$	&$15.02$	&$13.97$\\ 
\texttt{14} $50$ & $60$& $5.88$	&$0.05$	&$0.45$	&$4.70$	&$36.85$	&$31.48$	&$0.09$	&$0.38$	&$0.44$	&$64.83$	&$29.13$	&$15.84$	&$15.47$ \\
\rowcolor{blue!12}
\texttt{15} $80$ & $90$&$4.17$ &$0.01$	&$0.25$&	$0.67$	&$37.34$	&$36.11$	&$0.04$	&$0.24$&	$0.38$	&$68.21$	&$41.85$&	$\underline{17.75}$	&$\underline{15.92}$ \\ 
\bottomrule
\end{tabular}
}

\caption{\footnotesize\textbf{Models trained on \icg aligned \textit{(h,$\bm{\theta}$)}} display better generalization performance . We have trained both \csim and \syn on every subset of \textit{(h,$\theta$)} provided by the respective datasets. Avg$^{*}$ - Average IoU reported over all classes excluding \textit{other} and \textit{terrain}(both numbers are reported since a discrepancy was observed in other and terrain class from \csim which resulted in overlapping cases across these classes)}
\label{table:icg_expert}
\end{center}

\end{table}
\subsection{Advantage of \csim over Real Datasets}
Curating densely annotated real data under diverse conditions in a controlled / uncontrolled manner is prohibitively expensive. As a synthetic alternative, \csim is well-suited for evaluating model sensitivity to changing conditions. Further, due to the lack of diverse annotated data, we observe that real aerial datasets are somewhat homogenous. Consequently, models trained on real data under specific conditions (unlike \csim) struggle to generalize to differing conditions -- Rein DINOv2~\cite{wei2024stronger} models trained on \csim generalize better to \icg~\cite{drone-dataset} ($64.15$ mIoU) compared to ones trained on \aero~\cite{aeroscapes} ($28.29$ mIoU) and \ua~\cite{lyu2020uavid} ($45.57$ mIoU)\footnote{A common merging scheme was used as detailed in Table.~\ref{table:common_sky_class}}

\subsection{Synthetic$\to$Real Aligned Data Selection}
\label{sec:syn2real}
As stated in the main paper, for our Synthetic$\to$Real experiments, we train  
models on $(h, \theta)$ subsets of synthetic datasets that are 
aligned with corresponding real data $(h, \theta)$ conditions. In case of \ua and \aero 
we find that $(h=35m, \theta=45^{\circ})$ viewpoints in \csim and $(h=20m, \theta=30^{\circ})$ viewpoints in \syn are best aligned with \ua conditions (see Table.~\ref{table:uavid_expert} and Table.~\ref{table:aero_expert}) and provide best transfer performance. Similarly, for \icg we observe that $(h=15m,\theta=90^{\circ})$ \csim conditions are best aligned with the low-altitude, nadir perspective imagery in \icg and lead to best transfer performance (see Table.~\ref{table:icg_expert}). However, for \syn, we find that the model trained on $(h=80m,\theta=90^{\circ})$ has best transfer performance, indicating that model performance is more sensitive to pitch alignment than height alignment.

\subsection{Synthetic$\to$Real Additional Experiments}
\label{sec:syn2real add}

\begin{table}[ht!]

\centering
\centering
\setlength{\tabcolsep}{2.5pt}
\begin{tabular}{lccccc}
\toprule

\textbf{(Source)} & \multicolumn{3}{c}{\textbf{(Target) Real-World mIoU ($\uparrow$)}}\\ 
\textbf{Synthetic}& \ua & \aero & \icg \\ 
\midrule
\texttt{1} \synth & $39.56$ &$29.69$ & $9.12$\\
\texttt{2} \valid & $45.03$ &$32.80$ & $27.64$\\
\texttt{3} \syn & $\textbf{54.92}$ &$40.28$ & $20.01$\\
\texttt{4} \csim & $54.19$ &$\textbf{43.96}$ & $\textbf{28.10}$ \\

\bottomrule
\end{tabular}

\caption{\textbf{Syn$\to$Real Rein Semantic Segmentation}}
\label{table:synthetic}

\end{table}

\par\noindent
A proper comparison with other synthetic datasets on a recognition task would require similar classes, (which a subset of synthetic alternatives, such as MidAir~\cite{midAir}, do not satisfy).  Considering other desired attributes (outlined in Table.1 in the paper), we chose \syn~\cite{rizzoli2023syndrone} as the most aligned alternative for our experiments. For comprehensive analysis and additional comparisons with other synthetic alternatives across various Synthetic$\to$Real settings, we included comparisons with \valid~\cite{valid}, which features only nadir perspective, and \synth~\cite{synthaer}, which lacks human classes, in Table~\ref{table:synthetic}. From the results in Table~\ref{table:synthetic} , it is evident that models trained on \csim generalize better to real datasets compared to other synthetic alternatives. 

\subsection{\csim$+$ Real Data Experiments}
\label{sec:real+syn}
\begin{figure}
    \centering
    \includegraphics[width=\columnwidth]{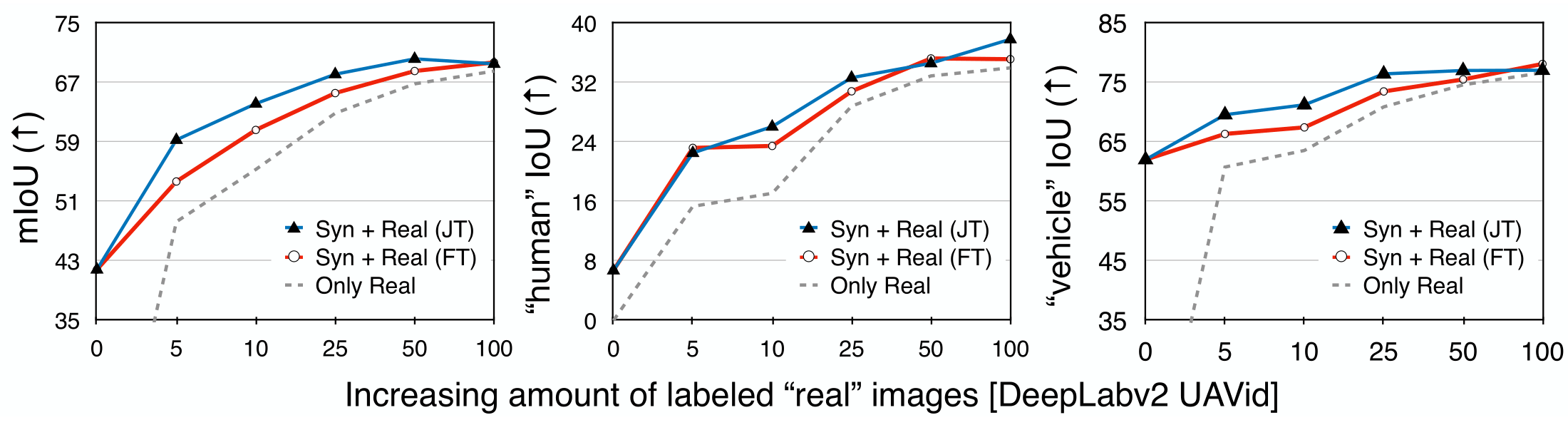}
    \caption{\footnotesize\textbf{[DeepLabv2 UAVid] \csim can augment ``real'' training data.} We show how \csim can additionally augment real (UAVid~\cite{lyu2020uavid}) training data. We compare DeepLabv2~\cite{chen2017deeplab} models trained using only $5\%,10\%,25\%,50\%,100\%$ of labeled UAVid~\cite{lyu2020uavid} images with counterparts that were either (1) pretrained on \csim, and finetuned on UAVid~\cite{lyu2020uavid} (FT) or (2) trained jointly on \csim and UAVid~\cite{lyu2020uavid} (JT). We find that \textbf{[Left]} additionally augmenting training data with \csim and help improve real-world generalization in low-shot regimes, \textbf{[Middle, Right]} especially for under-represented classes.}
    \label{fig:fig_1}
\end{figure}
\begin{figure}
    \centering
    \includegraphics[width=\columnwidth]{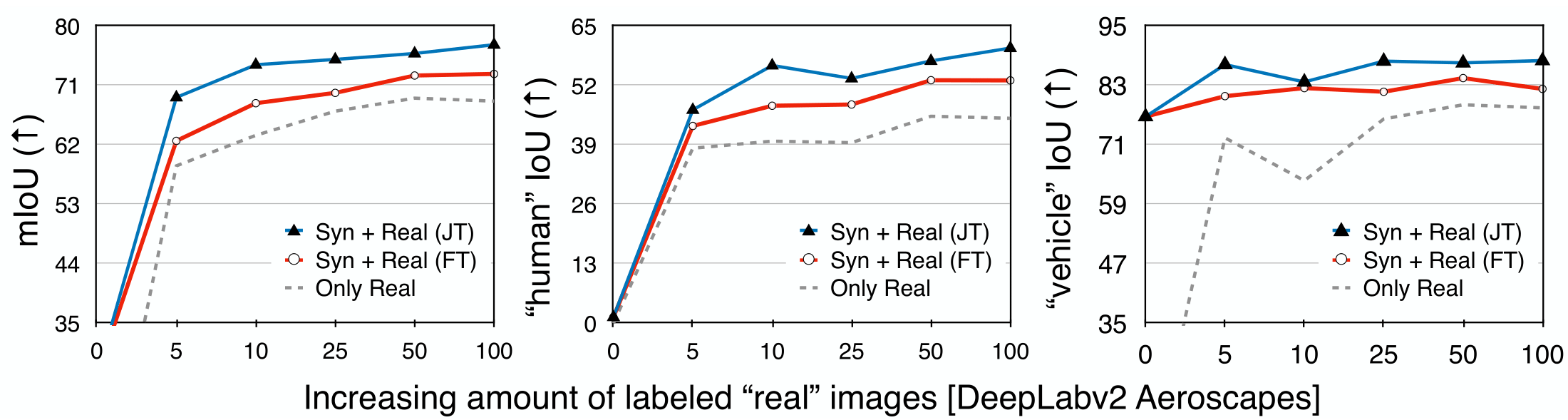}
    \caption{\footnotesize\textbf{[DeepLabv2 Aeroscapes] \csim can augment ``real'' training data.} We show how \csim can additionally augment real (Aeroscapes~\cite{aeroscapes}) training data. We compare DeepLabv2~\cite{chen2017deeplab} models trained using only $5\%,10\%,25\%,50\%,100\%$ of labeled Aeroscapes~\cite{aeroscapes} images with counterparts that were either (1) pretrained on \csim, and finetuned on Aeroscapes~\cite{aeroscapes} (FT) or (2) trained jointly on \csim and Aeroscapes~\cite{aeroscapes} (JT). We find that \textbf{[Left]} additionally augmenting training data with \csim and help improve real-world generalization in low-shot regimes, \textbf{[Middle, Right]} especially for under-represented classes.}
    \label{fig:fig_2}
\end{figure}
\begin{figure}
    \centering
    \includegraphics[width=\columnwidth]{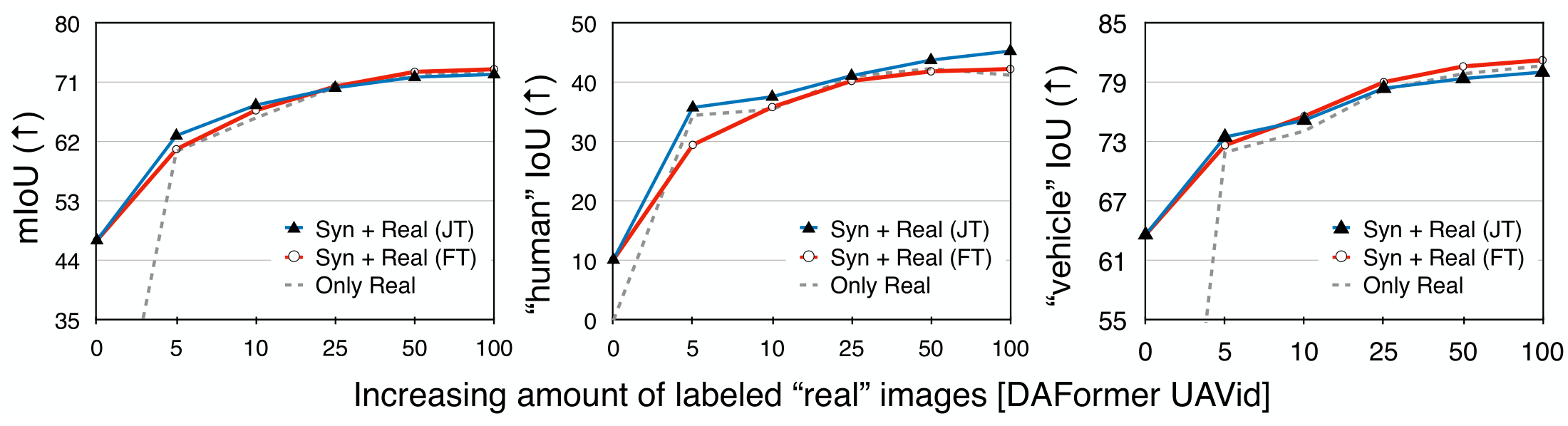}
    \caption{\footnotesize\textbf{[DAFormer UAVid] \csim can augment ``real'' training data.} We show how \csim can additionally augment real (UAVid~\cite{lyu2020uavid}) training data. We compare DAFormer~\cite{hoyer2022daformer} models trained using only $5\%,10\%,25\%,50\%,100\%$ of labeled UAVid~\cite{lyu2020uavid} images with counterparts that were either (1) pretrained on \csim, and finetuned on UAVid~\cite{lyu2020uavid} (FT) or (2) trained jointly on \csim and UAVid~\cite{lyu2020uavid} (JT). We find that \textbf{[Left]} additionally augmenting training data with \csim and help improve real-world generalization in low-shot regimes, \textbf{[Middle, Right]} especially for under-represented classes.}
    \label{fig:fig_3}
\end{figure}
\begin{figure}
    \centering
    \includegraphics[width=\columnwidth]{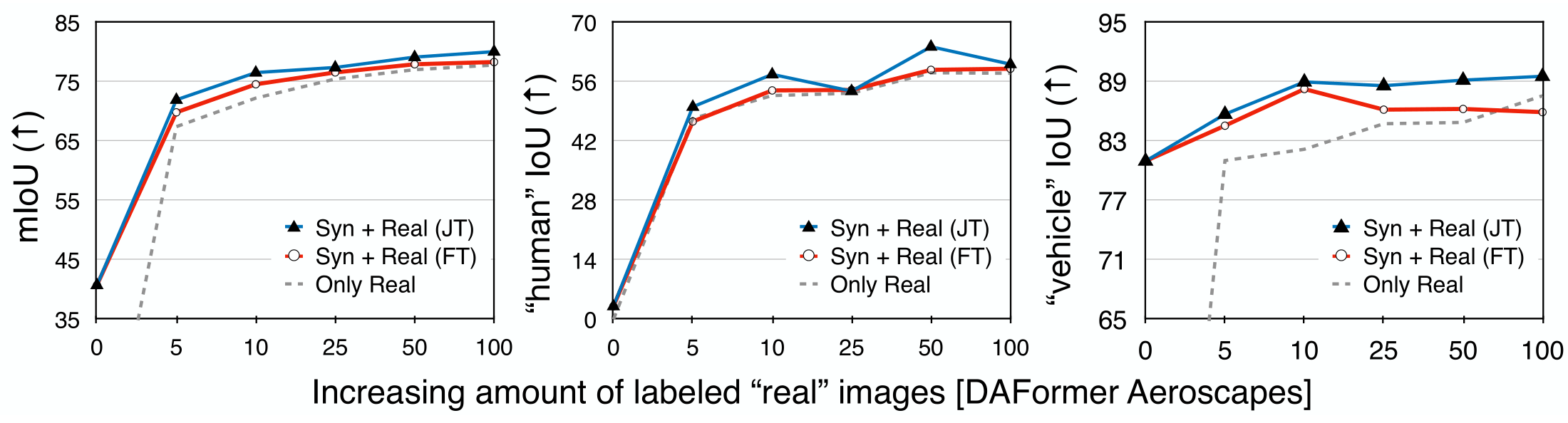}
    \caption{\footnotesize\textbf{[DAFormer Aeroscapes] \csim can augment ``real'' training data.} We show how \csim can additionally augment real (Aeroscapes~\cite{aeroscapes}) training data. We compare DAFormer~\cite{hoyer2022daformer} models trained using only $5\%,10\%,25\%,50\%,100\%$ of labeled Aeroscapes~\cite{aeroscapes} images with counterparts that were either (1) pretrained on \csim, and finetuned on Aeroscapes~\cite{aeroscapes} (FT) or (2) trained jointly on \csim and Aeroscapes~\cite{aeroscapes} (JT). We find that \textbf{[Left]} additionally augmenting training data with \csim and help improve real-world generalization in low-shot regimes, \textbf{[Middle, Right]} especially for under-represented classes.}
    \label{fig:fig_4}
\end{figure}
\begin{figure}
    \centering
    \includegraphics[width=\columnwidth]{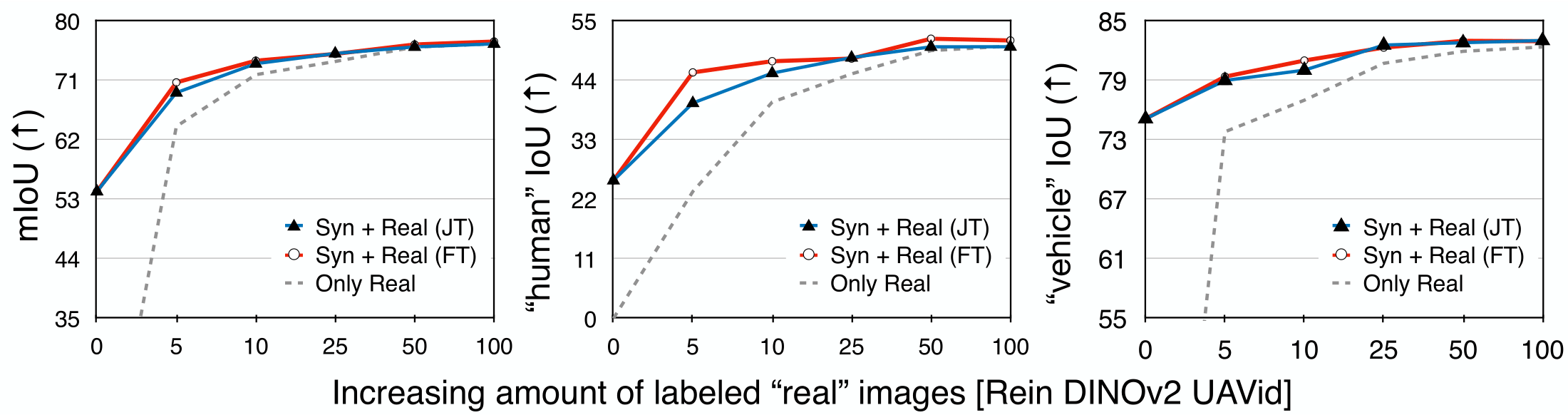}
    \caption{\footnotesize\textbf{[Rein DINOv2 UAVid] \csim can augment ``real'' training data.} We show how \csim can additionally augment real (UAVid~\cite{lyu2020uavid}) training data. We compare Rein DINOv2~\cite{wei2024stronger} models trained using only $5\%,10\%,25\%,50\%,100\%$ of labeled UAVid~\cite{lyu2020uavid} images with counterparts that were either (1) pretrained on \csim, and finetuned on UAVid~\cite{lyu2020uavid} (FT) or (2) trained jointly on \csim and UAVid~\cite{lyu2020uavid} (JT). We find that \textbf{[Left]} additionally augmenting training data with \csim and help improve real-world generalization in low-shot regimes, \textbf{[Middle, Right]} especially for under-represented classes.}
    \label{fig:dino_v2_budget}
\end{figure}

\begin{table}[htbp]
\footnotesize
\centering
\setlength{\tabcolsep}{2.5pt}
\begin{center}
\resizebox{\columnwidth}{!}{
\begin{tabular}{lcccccccccc}
\toprule
{\textbf{Training}} &{\textbf{Data Size}} & \multicolumn{8}{c}{\textbf{Synthetic$\rightarrow\ua$ mIoU ($\uparrow$)}}\\ 
& &Clutter &Building &Road &Tree &Low Vegetation &Human &Vehicle & \textbf{Avg}  \\
\midrule
\textbf{\csim}\\
\midrule
\texttt{1}  FT &$5\%$& $47.60$ & $83.73$ & $62.08$ & $58.98$ & $33.70$ & $23.21$ & $66.36$ & $\textbf{53.67}$ \\
\texttt{2}  FT &$10\%$& $50.06$ & $86.93$ & $67.70$ & $70.58$ & $58.09$ & $23.47$ & $67.46$ & $\textbf{60.61}$ \\
\texttt{3}  FT &$25\%$& $57.61$ & $89.37$ & $73.28$ & $73.30$ & $61.15$ & $30.82$ & $73.50$ & $\textbf{65.57}$ \\
\texttt{4}  FT &$50\%$& $62.44$ & $90.87$ & $75.84$ & $74.75$ & $65.16$ & $35.24$ & $75.52$ & $\textbf{68.54}$ \\
\texttt{5}  FT &$100\%$& $63.56$ & $91.23$ & $77.03$ & $76.01$ & $66.81$ & $35.14$ & $78.12$ & $\textbf{69.70}$ \\
\midrule
\texttt{6}  JT &$5\%$& $52.09$ & $86.83$ & $65.35$ & $68.45$ & $50.06$ & $22.53$ & $69.59$ & $\textbf{59.27}$ \\
\texttt{7}  JT &$10\%$& $56.23$ & $89.06$ & $71.84$ & $73.23$ & $61.35$ & $26.07$ & $71.25$ & $\textbf{64.15}$ \\
\texttt{8}  JT &$25\%$& $61.50$ & $90.45$ & $75.98$ & $75.18$ & $64.56$ & $32.65$ & $76.45$ & $\textbf{68.11}$ \\
\texttt{9}  JT &$50\%$& $65.70$ & $91.68$ & $78.14$ & $76.84$ & $67.29$ & $34.59$ & $77.02$ & $\textbf{70.18}$ \\
\texttt{10}  JT &$100\%$& $64.19$ & $89.06$ & $74.52$ & $76.59$ & $67.34$ & $37.82$ & $77.06$ & $\underline{69.51}$ \\
\midrule

\textbf{\textbf{\syn}}\\
\midrule
\texttt{11}  FT &$5\%$& $46.58$ & $82.81$ & $59.69$ & $58.95$ & $37.53$ & $21.50$ & $63.78$ & $52.97$ \\
\texttt{12}  FT &$10\%$& $48.08$ & $85.76$ & $64.56$ & $69.14$ & $54.29$ & $21.91$ & $65.94$ & $58.53$ \\
\texttt{13}  FT &$25\%$& $56.46$ & $88.78$ & $70.97$ & $72.06$ & $61.14$ & $28.59$ & $73.12$ & $64.45$ \\
\texttt{14}  FT &$50\%$& $61.70$ & $90.45$ & $73.83$ & $75.48$ & $64.67$ & $33.45$ & $74.98$ & $67.79$ \\
\texttt{15}  FT &$100\%$& $63.42$ & $91.16$ & $75.75$ & $76.56$ & $66.63$ & $33.53$ & $76.59$ & $69.10$ \\
\midrule
\texttt{16}  JT &$5\%$& $49.72$ & $81.39$ & $63.63$ & $65.52$ & $49.00$ & $6.70$ & $70.19$ & $55.17$ \\
\texttt{17}  JT &$10\%$& $52.94$ & $85.41$ & $67.25$ & $71.47$ & $59.05$ & $23.32$ & $72.43$ & $61.70$ \\
\texttt{18}  JT &$25\%$& $59.99$ & $87.79$ & $72.26$ & $74.00$ & $65.59$ & $36.32$ & $75.91$ & $67.40$ \\
\texttt{19}  JT &$50\%$& $63.38$ & $88.88$ & $74.09$ & $74.97$ & $66.85$ & $38.47$ & $77.45$ & $69.16$ \\
\texttt{20}  JT &$100\%$& $64.06$ & $89.09$ & $74.61$ & $75.98$ & $67.31$ & $40.59$ & $78.40$ & $\textbf{70.01}$ \\
\midrule

\textbf{\textbf{Target}}\\
\midrule
\texttt{21}  Target &$5\%$& $41.36$ & $80.24$ & $57.07$ & $57.20$ & $25.72$ & $15.35$ & $60.79$ & $48.25$ \\
\texttt{22}  Target &$10\%$& $42.94$ & $81.50$ & $61.06$ & $67.51$ & $53.32$ & $17.10$ & $63.58$ & $55.29$ \\
\texttt{23}  Target &$25\%$& $53.57$ & $86.89$ & $69.70$ & $70.95$ & $59.22$ & $28.82$ & $70.91$ & $62.86$ \\
\texttt{24}  Target &$50\%$& $59.75$ & $89.42$ & $72.58$ & $74.28$ & $64.09$ & $32.89$ & $74.61$ & $66.81$ \\
\texttt{25}  Target &$100\%$& $62.45$ & $90.76$ & $74.41$ & $75.83$ & $65.65$ & $33.97$ & $76.65$ & $68.53$ \\
\bottomrule
\end{tabular}}

\caption{\footnotesize\textbf{[DeepLabv2 UAVid] \csim can augment ``real'' training data.} We compare \csim against \syn for their ability to additionally augment real (UAVid~\cite{lyu2020uavid}) training data. We compare DeepLabv2~\cite{chen2017deeplab} models trained using only $5\%,10\%,25\%,50\%,100\%$ of labeled UAVid~\cite{lyu2020uavid} images with counterparts that were either (1) pretrained on \csim/\syn, and finetuned on UAVid~\cite{lyu2020uavid} (FT) or (2) trained jointly on \csim/\syn and UAVid~\cite{lyu2020uavid} (JT). We find that both FT and JT with \csim outperforms \syn in almost all of the different labeled data splits.}
\label{table:budget_uavid_dlv2}
\end{center}
\end{table}

\begin{table}[htbp]
\footnotesize
\centering
\setlength{\tabcolsep}{2.5pt}
\begin{center}
\resizebox{\columnwidth}{!}{
\begin{tabular}{lccccccccccc}
\toprule
{\textbf{Training}} &{\textbf{Data Size}} & \multicolumn{9}{c}{\textbf{Synthetic$\rightarrow\aero$ mIoU ($\uparrow$)}}\\ 
& &Background & Bicycle &Person&Vehicle &Vegetation &Building &Road & Sky & Avg  \\ 
\midrule
\textbf{\csim}\\
\midrule
\texttt{1}  FT &$5\%$& $59.25$ & $4.69$ & $43.11$ & $80.81$ & $91.79$ & $59.32$ & $72.43$ & $89.13$ & $\textbf{62.57}$ \\
\texttt{2}  FT &$10\%$& $71.45$ & $14.76$ & $47.53$ & $82.43$ & $92.96$ & $60.08$ & $83.92$ & $93.09$ & $\textbf{68.28}$ \\
\texttt{3}  FT &$25\%$& $72.93$ & $21.78$ & $47.80$ & $81.69$ & $93.42$ & $62.25$ & $84.51$ & $94.30$ & $\textbf{69.84}$ \\
\texttt{4}  FT &$50\%$& $76.15$ & $23.69$ & $53.11$ & $84.45$ & $93.58$ & $67.00$ & $87.73$ & $94.06$ & $\textbf{72.47}$ \\
\texttt{5}  FT &$100\%$& $76.91$ & $25.30$ & $53.07$ & $82.27$ & $94.05$ & $67.81$ & $87.78$ & $94.51$ & $\textbf{72.71}$ \\
\midrule
\texttt{6}  JT &$5\%$& $67.86$ &	$22.42$ &	$46.63$ &	$87.18$ &	$92.11$ &	$59.69$ &	$85.04$ &	$92.31$ &	$\textbf{69.16}$ \\
\texttt{7}  JT &$10\%$& $76.24$ &	$30.44$ &	$56.33$ &	$83.67$ &	$92.15$ &	$69.39$ &	$89.70$ &	$94.86$ &	$\textbf{74.10}$ \\
\texttt{8}  JT &$25\%$& $76.33$ &	$34.14$ &	$53.50$ &	$87.87$ &	$93.12$ &	$68.89$ &	$90.23$ &	$95.23$ &	$\textbf{74.91}$ \\
\texttt{9}  JT &$50\%$& $78.13$ &	$32.06$ &	$57.32$ &	$87.52$ &	$93.23$ &	$73.14$ &	$90.42$ &	$94.69$ &	$\underline{75.81}$ \\
\texttt{10}  JT &$100\%$& $80.35$ &	$32.43$ &	$60.14$ &	$87.98$ &	$94.00$ &	$75.20$ &	$91.89$ &	$95.06$ &	$\textbf{77.13}$ \\
\midrule

\textbf{\textbf{\syn}}\\
\midrule
\texttt{11}  FT &$5\%$& $57.20$ & $3.79$ & $39.07$ & $77.44$ & $90.79$ & $58.60$ & $69.09$ & $88.48$ & $60.56$ \\
\texttt{12}  FT &$10\%$& $69.18$ & $14.77$ & $47.07$ & $64.87$ & $92.29$ & $60.42$ & $81.09$ & $91.94$ & $65.20$ \\
\texttt{13}  FT &$25\%$& $72.21$ & $23.27$ & $44.35$ & $79.72$ & $92.83$ & $63.94$ & $83.31$ & $92.31$ & $68.99$ \\
\texttt{14}  FT &$50\%$& $74.28$ & $22.32$ & $49.59$ & $81.59$ & $93.01$ & $64.70$ & $85.56$ & $92.81$ & $70.49$ \\
\texttt{15}  FT &$100\%$& $74.61$ & $22.45$ & $49.13$ & $81.42$ & $93.15$ & $64.18$ & $86.15$ & $92.64$ & $70.47$ \\
\midrule
\texttt{16}  JT &$5\%$& $63.05$ & $23.34$ & $46.50$ & $83.30$ & $92.24$ & $64.15$ & $78.81$ & $93.64$ & $68.12$ \\
\texttt{17}  JT &$10\%$& $74.54$ & $26.14$ & $53.81$ & $81.26$ & $93.01$ & $66.44$ & $88.83$ & $94.62$ & $72.32$ \\
\texttt{18}  JT &$25\%$& $76.96$ & $33.69$ & $55.34$ & $83.53$ & $93.67$ & $69.00$ & $90.37$ & $95.26$ & $74.73$ \\
\texttt{19}  JT &$50\%$& $79.35$ & $31.45$ & $59.18$ & $85.80$ & $93.66$ & $74.00$ & $91.27$ & $95.05$ & $\textbf{76.22}$ \\
\texttt{20}  JT &$100\%$& $79.66$ & $31.82$ & $58.71$ & $86.87$ & $93.75$ & $74.65$ & $91.95$ & $94.20$ & $76.45$ \\
\midrule

\textbf{\textbf{Target}}\\
\midrule
\texttt{21}  Target &$5\%$& $53.92$ & $7.33$ & $38.25$ & $72.41$ & $90.15$ & $53.49$ & $66.79$ & $87.95$ & $58.79$ \\
\texttt{22}  Target &$10\%$& $67.08$ & $14.89$ & $39.79$ & $63.76$ & $91.65$ & $58.98$ & $79.45$ & $91.70$ & $63.41$ \\
\texttt{23}  Target &$25\%$& $70.51$ & $20.69$ & $39.46$ & $76.20$ & $92.34$ & $62.25$ & $82.51$ & $92.56$ & $67.07$ \\
\texttt{24}  Target &$50\%$& $73.10$ & $20.95$ & $45.23$ & $79.07$ & $92.68$ & $64.70$ & $84.48$ & $92.11$ & $69.04$ \\
\texttt{25}  Target &$100\%$& $73.21$ & $19.21$ & $44.79$ & $78.44$ & $92.95$ & $63.49$ & $85.34$ & $91.27$ & $68.59$ \\
\bottomrule
\end{tabular}}

\caption{\footnotesize\textbf{[DeepLabv2 Aeroscapes] \csim can augment ``real'' training data.} We compare \csim against \syn for their ability to additionally augment real (Aeroscapes~\cite{aeroscapes}) training data. We compare DeepLabv2~\cite{chen2017deeplab} models trained using only $5\%,10\%,25\%,50\%,100\%$ of labeled Aeroscapes~\cite{aeroscapes} images with counterparts that were either (1) pretrained on \csim/\syn, and finetuned on Aeroscapes~\cite{aeroscapes} (FT) or (2) trained jointly on \csim/\syn and Aeroscapes~\cite{aeroscapes} (JT). We find that both FT and JT with \csim outperforms \syn in almost all of the different labeled data splits.}

\label{table:budget_aero_dlv2}
\end{center}

\end{table}
    
\begin{table}[htbp]
\footnotesize
\centering
\setlength{\tabcolsep}{2.5pt}
\begin{center}
\resizebox{\columnwidth}{!}{
\begin{tabular}{lcccccccccc}
\toprule
{\textbf{Training}} &{\textbf{Data Size}} & \multicolumn{8}{c}{\textbf{Synthetic$\rightarrow\ua$ mIoU ($\uparrow$)}}\\ 
& &Clutter &Building &Road &Tree &Low Vegetation &Human &Vehicle & Avg  \\
\midrule
\textbf{\csim}\\
\midrule
\texttt{1}  FT &$5\%$& $52.63$ & $88.02$ & $67.74$ & $66.56$ & $49.14$ & $29.53$ & $72.68$ & $60.90$ \\
\texttt{2}  FT &$10\%$& $57.43$ & $89.48$ & $72.93$ & $74.26$ & $61.96$ & $35.89$ & $75.61$ & $66.79$ \\
\texttt{3}  FT &$25\%$& $63.43$ & $90.87$ & $78.66$ & $75.59$ & $64.94$ & $40.29$ & $79.05$ & $70.41$ \\
\texttt{4}  FT &$50\%$& $67.06$ & $92.22$ & $79.76$ & $77.86$ & $68.97$ & $41.90$ & $80.64$ & $72.63$ \\
\texttt{5}  FT &$100\%$& $67.67$ & $92.51$ & $79.66$ & $78.68$ & $69.03$ & $42.29$ & $81.27$ & $73.02$ \\
\midrule
\texttt{6}  JT &$5\%$& $53.72$ & $85.99$ & $66.35$ & $69.02$ & $56.45$ & $35.82$ & $73.51$ & $\textbf{62.97}$ \\
\texttt{7}  JT &$10\%$& $59.77$ & $87.82$ & $72.97$ & $74.68$ & $64.95$ & $37.63$ & $75.21$ & $\textbf{67.58}$ \\
\texttt{8}  JT &$25\%$& $63.58$ & $88.68$ & $76.22$ & $75.80$ & $67.49$ & $41.17$ & $78.43$ & $\textbf{70.20}$ \\
\texttt{9}  JT &$50\%$& $66.83$ & $89.63$ & $77.76$ & $76.90$ & $68.45$ & $43.79$ & $79.41$ & $\textbf{71.83}$ \\
\texttt{10}  JT &$100\%$& $66.92$ & $89.74$ & $76.89$ & $77.98$ & $68.81$ & $45.32$ & $80.06$ & $72.25$ \\
\midrule

\textbf{\textbf{\syn}}\\
\midrule
\texttt{11}  FT &$5\%$& $53.53$ & $87.81$ & $67.31$ & $67.51$ & $55.40$ & $33.94$ & $73.40$ & $62.70$ \\
\texttt{12}  FT &$10\%$& $56.80$ & $89.64$ & $72.31$ & $74.22$ & $64.12$ & $36.77$ & $75.15$ & $67.00$ \\
\texttt{13}  FT &$25\%$& $63.77$ & $91.11$ & $78.18$ & $76.17$ & $66.63$ & $41.27$ & $78.90$ & $70.86$ \\
\texttt{14}  FT &$50\%$& $66.89$ & $92.06$ & $79.14$ & $77.79$ & $69.22$ & $43.32$ & $80.47$ & $72.70$ \\
\texttt{15}  FT &$100\%$& $67.48$ & $92.48$ & $79.48$ & $78.76$ & $69.73$ & $42.82$ & $81.57$ & $73.19$ \\
\midrule
\texttt{16}  JT &$5\%$& $53.54$ & $85.58$ & $64.30$ & $70.07$ & $58.64$ & $29.86$ & $73.21$ & $62.17$ \\
\texttt{17}  JT &$10\%$& $58.00$ & $87.26$ & $71.23$ & $74.57$ & $64.65$ & $35.58$ & $75.08$ & $66.62$ \\
\texttt{18}  JT &$25\%$& $61.82$ & $88.35$ & $74.09$ & $75.38$ & $66.43$ & $40.85$ & $78.21$ & $69.31$ \\
\texttt{19}  JT &$50\%$& $65.10$ & $89.19$ & $75.80$ & $76.95$ & $68.15$ & $40.65$ & $78.52$ & $70.62$ \\
\texttt{20}  JT &$100\%$& $66.78$ & $89.81$ & $76.56$ & $77.91$ & $69.38$ & $45.34$ & $80.45$ & $72.32$ \\
\midrule

\textbf{\textbf{Target}}\\
\midrule
\texttt{21}  Target &$5\%$& $51.20$ & $86.59$ & $62.68$ & $66.73$ & $50.46$ & $34.50$ & $71.97$ & $60.59$ \\
\texttt{22}  Target &$10\%$& $53.95$ & $88.35$ & $70.56$ & $73.65$ & $63.29$ & $35.50$ & $74.12$ & $65.63$ \\
\texttt{23}  Target &$25\%$& $62.86$ & $90.48$ & $76.89$ & $76.02$ & $66.66$ & $41.03$ & $78.31$ & $70.31$ \\
\texttt{24}  Target &$50\%$& $66.23$ & $91.84$ & $78.75$ & $77.61$ & $68.44$ & $42.36$ & $79.90$ & $72.16$ \\
\texttt{25}  Target &$100\%$& $66.67$ & $92.20$ & $79.16$ & $78.35$ & $68.93$ & $41.25$ & $80.70$ & $72.47$ \\
\bottomrule
\end{tabular}}

\caption{\footnotesize\textbf{[DAFormer UAVid] \csim can augment ``real'' training data.} We compare \csim against \syn for their ability to additionally augment real (UAVid~\cite{lyu2020uavid}) training data. We compare DAFormer~\cite{hoyer2022daformer} models trained using only $5\%,10\%,25\%,50\%,100\%$ of labeled UAVid~\cite{lyu2020uavid} images with counterparts that were either (1) pretrained on \csim/\syn, and finetuned on UAVid~\cite{lyu2020uavid} (FT) or (2) trained jointly on \csim/\syn and UAVid~\cite{lyu2020uavid} (JT). We find that both FT and JT with \csim outperforms or is on par with \syn in almost all of the different labeled data splits.}

\label{table:budget_uavid_daf}
\end{center}
\end{table}

\begin{table}[htbp]
\footnotesize
\centering
\setlength{\tabcolsep}{2.5pt}
\begin{center}
\resizebox{\columnwidth}{!}{
\begin{tabular}{lccccccccccc}
\toprule
{\textbf{Training}} &{\textbf{Data Size}} & \multicolumn{9}{c}{\textbf{Synthetic$\rightarrow\aero$ mIoU ($\uparrow$)}}\\ 
& &Background & Bicycle &Person&Vehicle &Vegetation &Building &Road & Sky & Avg  \\ 
\midrule
\textbf{\csim}\\
\midrule
\texttt{1}  FT &$5\%$& $66.93$ & $28.05$ & $46.63$ & $84.56$ & $93.29$ & $64.03$ & $79.76$ & $95.41$ & $69.83$ \\
\texttt{2}  FT &$10\%$& $76.11$ & $31.60$ & $53.95$ & $88.26$ & $93.92$ & $67.90$ & $88.45$ & $96.25$ & $74.56$ \\
\texttt{3}  FT &$25\%$& $79.68$ & $39.95$ & $54.07$ & $86.18$ & $94.22$ & $71.31$ & $91.53$ & $95.66$ & $\textbf{76.56}$ \\
\texttt{4}  FT &$50\%$& $81.09$ & $39.03$ & $58.80$ & $86.25$ & $94.42$ & $76.20$ & $91.15$ & $96.45$ & $\textbf{77.93}$ \\
\texttt{5}  FT &$100\%$& $81.78$ & $41.22$ & $59.06$ & $85.93$ & $94.62$ & $76.06$ & $91.47$ & $96.40$ & $\textbf{78.31}$ \\
\midrule
\texttt{6}  JT &$5\%$& $69.98$ &	$33.78$ &	$50.09$ &	$85.72$ &	$93.94$ &	$64.09$ &	$82.32$ &	$95.64$ &	$\textbf{71.94}$ \\
\texttt{7}  JT &$10\%$& $78.47$ &	$38.45$ &	$57.73$ &	$88.98$ &	$94.39$ &	$67.87$ &	$90.23$ &	$96.14$ &	$\textbf{76.53}$ \\
\texttt{8}  JT &$25\%$& $81.07$ &	$40.95$ &	$53.79$ &	$88.60$ &	$94.30$ &	$71.92$ &	$92.76$ &	$95.77$ &	$77.39$ \\
\texttt{9}  JT &$50\%$& $81.55$ &	$39.43$ &	$64.25$ &	$89.16$ &	$94.30$ &	$76.81$ &	$91.34$ &	$96.30$ &	$79.14$ \\
\texttt{10}  JT &$100\%$& $83.13$ &	$44.90$ &	$60.14$ &	$89.55$ &	$94.93$ &	$79.55$ &	$92.49$ &	$95.83$ &	$\textbf{80.06}$ \\
\midrule

\textbf{\textbf{\syn}}\\
\midrule
\texttt{11}  FT &$5\%$& $70.94$ & $29.02$ & $47.17$ & $82.25$ & $93.23$ & $64.19$ & $83.31$ & $93.27$ & $70.42$ \\
\texttt{12}  FT &$10\%$& $76.62$ & $29.13$ & $60.31$ & $86.91$ & $93.80$ & $66.29$ & $89.80$ & $95.65$ & $74.81$ \\
\texttt{13}  FT &$25\%$& $78.97$ & $34.86$ & $57.69$ & $86.06$ & $93.83$ & $69.73$ & $92.03$ & $95.33$ & $76.06$ \\
\texttt{14}  FT &$50\%$& $80.28$ & $34.42$ & $63.33$ & $85.34$ & $94.31$ & $73.71$ & $91.55$ & $95.69$ & $77.33$ \\
\texttt{15}  FT &$100\%$& $80.90$ & $35.47$ & $62.18$ & $85.43$ & $94.45$ & $74.45$ & $92.12$ & $95.76$ & $77.59$ \\
\midrule
\texttt{16}  JT &$5\%$& $69.12$ &	$30.82$ &	$53.16$ &	$81.33$ &	$93.19$ &	$61.11$ &	$83.80$ &	$95.36$ &	$70.99$ \\
\texttt{17}  JT &$10\%$& $77.38$ &	$40.42$ &	$57.80$ &	$88.32$ &	$94.14$ &	$68.06$ &	$90.24$ &	$95.60$ &	$76.47$ \\
\texttt{18}  JT &$25\%$& $80.64$ &	$44.19$ &	$53.95$ &	$89.80$ &	$94.28$ &	$71.89$ &	$92.58$ &	$96.14$ &	$77.94$ \\
\texttt{19}  JT &$50\%$& $81.82$ &	$40.86$ &	$63.15$ &	$89.20$ &	$94.38$ &	$77.72$ &	$92.29$ &	$96.43$ &	$79.42$ \\
\texttt{20}  JT &$100\%$& $79.38$ &	$35.85$ &	$50.91$ &	$86.01$ &	$94.08$ &	$70.88$ &	$90.73$ &	$94.32$ &	$75.27$ \\
\midrule

\textbf{\textbf{Target}}\\
\midrule
\texttt{21}  Target &$5\%$& $66.26$ &	$24.03$ &	$47.34$ &	$81.04$ &	$92.88$ &	$58.85$ &	$78.11$ &	$90.85$ &	$67.42$\\
\texttt{22}  Target &$10\%$& $74.91$ &	$32.00$ &	$52.74$ &	$82.18$ &	$93.62$ &	$59.63$ &	$88.73$ &	$94.00$ &	$72.23$\\
\texttt{23}  Target &$25\%$& $78.16$ &	$41.22$ &	$53.29$ &	$84.76$ &	$93.95$ &	$66.04$ &	$90.92$ &	$95.28$ &	$75.45$\\
\texttt{24}  Target &$50\%$& $80.09$ &	$38.73$ &	$58.11$ &	$84.89$ &	$94.14$ &	$73.72$ &	$90.85$ &	$95.61$ &	$77.02$\\
\texttt{25}  Target &$100\%$& $80.35$ &	$42.65$ &	$57.99$ &	$87.58$ &	$94.48$ &	$72.65$ &	$91.20$ &	$95.51$ &	$77.80$\\
\bottomrule
\end{tabular}}

\caption{\footnotesize\textbf{[DAFormer Aeroscapes] \csim can augment ``real'' training data.} We compare \csim against \syn for their ability to additionally augment real (Aeroscapes~\cite{aeroscapes}) training data. We compare DAFormer~\cite{hoyer2022daformer} models trained using only $5\%,10\%,25\%,50\%,100\%$ of labeled Aeroscapes~\cite{aeroscapes} images with counterparts that were either (1) pretrained on \csim/\syn, and finetuned on Aeroscapes~\cite{aeroscapes} (FT) or (2) trained jointly on \csim/\syn and Aeroscapes~\cite{aeroscapes} (JT). We find that both FT and JT with \csim outperforms or is on par with \syn in almost all of the different labeled data splits.}
\label{table:budget_aero_daf}
\end{center}
\end{table}

\begin{table}[htbp]
\footnotesize
\centering
\setlength{\tabcolsep}{2.5pt}
\begin{center}
\resizebox{\columnwidth}{!}{
\begin{tabular}{lcccccccccc}
\toprule
{\textbf{Training}} &{\textbf{Data Size}} & \multicolumn{8}{c}{\textbf{Synthetic$\rightarrow\ua$ mIoU ($\uparrow$)}}\\ 
& &Clutter &Building &Road &Tree &Low Vegetation &Human &Vehicle & Avg  \\
\midrule
\textbf{\csim}\\
\midrule
\texttt{1}  FT &$5\%$& $61.91$ & $91.38$ & $77.16$ & $74.79$ & $64.8$ & $45.49$ & $79.39$ & $\textbf{70.7}$ \\
\texttt{2}  FT &$10\%$& $67.41$ & $92.21$ & $81.14$ & $78.79$ & $69.82$ & $47.56$ & $81.01$ & $\textbf{73.99}$ \\
\texttt{3}  FT &$25\%$& $69.32$ & $93.39$ & $83.75$ & $78.88$ & $69.4$ & $48.05$ & $82.32$ & $75.01$ \\
\texttt{4}  FT &$50\%$& $71.37$ & $93.57$ & $84.15$ & $80.14$ & $71.14$ & $51.72$ & $83$ & $\textbf{76.44}$ \\
\texttt{5}  FT &$100\%$& $72.96$ & $94.06$ & $84.33$ & $80.63$ & $71.89$ & $51.4$ & $82.96$ & $\textbf{76.89}$ \\
\midrule
\texttt{6}  JT &$5\%$& $61.47$ & $90.65$ & $76.46$ & $73.78$ & $62.87$ & $39.83$ & $79$ & ${69.15}$ \\
\texttt{7}  JT &$10\%$& $67.55$ & $92.12$ & $82.54$ & $78.21$ & $68.94$ & $45.36$ & $80.04$ & ${73.54}$ \\
\texttt{8}  JT &$25\%$& $69.07$ & $93.41$ & $83.34$ & $79.07$ & $69.89$ & $48.23$ & $82.57$ & $\textbf{75.08}$ \\
\texttt{9}  JT &$50\%$& $71.1$ & $93.5$ & $84.11$ & $79.9$ & $70.9$ & $50.22$ & $82.81$ & ${76.08}$ \\
\texttt{10} JT &$100\%$& $72.16$ & $93.93$ & $84.14$ & $80.32$ & $71.99$ & $50.24$ & $83.03$ & $76.54$ \\
\midrule

\textbf{\textbf{Target}}\\
\midrule
\texttt{21}  Target &$5\%$& $58.19$ & $89.25$ & $71.01$ & $71.86$ & $60.65$ & $23.46$ & $73.83$ & $64.04$ \\
\texttt{22}  Target &$10\%$& $66.39$ & $92.22$ & $80.62$ & $78.11$ & $68.78$ & $40.00$ & $77.01$ & $71.87$ \\
\texttt{23}  Target &$25\%$& $68.57$ & $93.14$ & $82.59$ & $78.34$ & $68.52$ & $45.22$ & $80.73$ & $73.87$ \\
\texttt{24}  Target &$50\%$& $71.52$ & $93.93$ & $84.27$ & $80.07$ & $71.12$ & $49.53$ & $81.94$ & $76.05$ \\
\texttt{25}  Target &$100\%$& $72.38$ & $93.98$ & $84.63$ & $80.63$ & $71.52$ & $50.36$ & $82.38$ & $76.55$ \\
\bottomrule
\end{tabular}}

\caption{\footnotesize\textbf{[Rein DINOv2 UAVid] \csim can augment ``real'' training data.} We show \csim's ability to additionally augment real (UAVid~\cite{lyu2020uavid}) training data. We compare Rein DINOv2~\cite{wei2024stronger} models trained using only $5\%,10\%,25\%,50\%,100\%$ of labeled UAVid~\cite{lyu2020uavid} images with counterparts that were either (1) pretrained on \csim, and finetuned on UAVid~\cite{lyu2020uavid} (FT) or (2) trained jointly on \csim and UAVid~\cite{lyu2020uavid} (JT). We find that both FT and JT with \csim improve the performance on Target only.}

\label{table:budget_uavid_dinov2}
\end{center}
\end{table}

\par\noindent
In addition to zero-shot transfer to real data, 
we also show how \csim is useful as additional training data when labeled real-world data is available. In Fig.~\ref{fig:fig_1} and Fig.~\ref{fig:fig_2}, we compare the performance of DeepLabv2 \cite{chen2017deeplab} for \csim$\to$\ua~\cite{lyu2020uavid} and for \csim$\to$\aero ~\cite{aeroscapes} trained only using $5\%,\\ 10\%,25\%,50\%,100\%$ of \ua~\cite{lyu2020uavid} and \aero~\cite{aeroscapes} training images respectively with counterparts that were either pretrained using \csim data or additionally supplemented with \csim data at training time. In Fig.~\ref{fig:fig_3} and Fig.~\ref{fig:fig_4} we make a similar comparison with the DAFormer \cite{hoyer2022daformer} architecture, and in Fig.~\ref{fig:dino_v2_budget} with the Rein DINOv2 \cite{wei2024stronger} architecture. In low-shot regimes (when little ``real'' world data is available), \csim data (either explicitly via joint training or implicitly via finetuning) is beneficial in improving recognition performance. We find this to be especially beneficial for under-represented classes in aerial imagery (such as \texttt{humans} and \texttt{vehicles}). 
In Table.~\ref{table:budget_uavid_dlv2} and Table.~\ref{table:budget_aero_dlv2} we 
present similar fine-grained (per-class) comparison of \csim with \syn for a DeepLabv2 model when real data is available for training -- via Target-Only, Finetuning or Joint-Training.Tables.~\ref{table:budget_uavid_daf} and~\ref{table:budget_aero_daf} show similar comparisons using the DAFormer \cite{hoyer2022daformer} model. Table.~\ref{table:budget_uavid_dinov2} shows similar comparison using the Rein DINOv2 \cite{wei2024stronger} model.

\subsection{\csim Diagnostic Experiments}

\begin{table*}[htbp]

\parbox{.2\textwidth}{
\centering
\setlength{\tabcolsep}{2.5pt}
\resizebox{0.25\textwidth}{!}{
\begin{tabular}{lccccc}
\toprule
&\multicolumn{4}{c}{\textbf{Test mIoU ($\uparrow$)}} \\
\multirow{2}{*}{\textbf{Height}}& \multicolumn{4}{c}{\textbf{Pitch}} \\
& $\theta=0^{\circ}$ & $\theta=45^{\circ}$ & $\theta=60^{\circ}$ & $\theta=90^{\circ}$\\
\midrule
\texttt{1} $h=15$m &\textcolor{blue}{72.57}&66.38&57&39.46\\
\texttt{2} $h=35$m &\textcolor{blue}{59.07} &55.55 &54.43 &39.23\\
\texttt{3} $h=60$m &\textcolor{blue}{48.58} &44.94 &43.85 &31.71\\
\bottomrule
\end{tabular}}

\caption*{(a) \textbf{Height $15$ \& Pitch $0^{\circ}$}}
\label{table:s2s_h15p0}
}
\hfill
\parbox{.2\textwidth}{
\centering
\setlength{\tabcolsep}{2.5pt}
\resizebox{0.25\textwidth}{!}{
\begin{tabular}{lccccc}
\toprule
&\multicolumn{4}{c}{\textbf{Test mIoU ($\uparrow$)}} \\
\multirow{2}{*}{\textbf{Height}}& \multicolumn{4}{c}{\textbf{Pitch}} \\
& $\theta=0^{\circ}$ & $\theta=45^{\circ}$ & $\theta=60^{\circ}$ & $\theta=90^{\circ}$\\
\midrule
\texttt{1} $h=15$m &59.03 &\textcolor{blue}{66.98} &63.07	&61.85\\
\texttt{2} $h=35$m &43.8 &\textcolor{blue}{49.41} &\textcolor{blue}{49.91} &45.19\\
\texttt{3} $h=60$m &34.27 &\textcolor{blue}{39.86} &\textcolor{blue}{40.1} &35.42\\
\bottomrule
\end{tabular}}

\caption*{(b) \textbf{Height $15$ \& Pitch $45^{\circ}$}}
\label{table:s2s_h15p45}
}
\hfill
\parbox{.2\textwidth}{
\centering
\setlength{\tabcolsep}{2.5pt}
\resizebox{0.25\textwidth}{!}{
\begin{tabular}{lccccc}
\toprule
&\multicolumn{4}{c}{\textbf{Test mIoU ($\uparrow$)}} \\
\multirow{2}{*}{\textbf{Height}}& \multicolumn{4}{c}{\textbf{Pitch}} \\
& $\theta=0^{\circ}$ & $\theta=45^{\circ}$ & $\theta=60^{\circ}$ & $\theta=90^{\circ}$\\
\midrule
\texttt{1} $h=15$m & 45.36	&64.13	&\textcolor{blue}{68.88}	&\textcolor{blue}{69.28}\\
\texttt{2} $h=35$m &30.49 &45.05 &\textcolor{blue}{50.39}	&\textcolor{blue}{53.44}\\
\texttt{3} $h=60$m &21.86 &31.21 &\textcolor{blue}{36.51} &\textcolor{blue}{38.43}\\
\bottomrule
\end{tabular}}

\caption*{(c) \textbf{Height $15$ \& Pitch $60^{\circ}$}}
\label{table:s2s_hp}
}
\hfill
\parbox{.2\textwidth}{
\centering
\setlength{\tabcolsep}{2.5pt}
\resizebox{0.25\textwidth}{!}{
\begin{tabular}{lccccc}
\toprule
&\multicolumn{4}{c}{\textbf{Test mIoU ($\uparrow$)}} \\
\multirow{2}{*}{\textbf{Height}}& \multicolumn{4}{c}{\textbf{Pitch}} \\
& $\theta=0^{\circ}$ & $\theta=45^{\circ}$ & $\theta=60^{\circ}$ & $\theta=90^{\circ}$\\
\midrule
\texttt{1} $h=15$m &30.61 &55.79 &64.98 &\textcolor{blue}{71.93}\\
\texttt{2} $h=35$m &16.84 &33.2&37.89 &\textcolor{blue}{48.63}\\
\texttt{3} $h=60$m &11.26 &21.12 &26.38 &\textcolor{blue}{31.22}\\
\bottomrule
\end{tabular}}

\caption*{(d) \textbf{Height $15$ \& Pitch $90^{\circ}$}}
\label{table:s2s_h15p90}
}
\parbox{.2\textwidth}{
\centering
\setlength{\tabcolsep}{2.5pt}
\resizebox{0.25\textwidth}{!}{
\begin{tabular}{lccccc}
\toprule
&\multicolumn{4}{c}{\textbf{Test mIoU ($\uparrow$)}} \\
\multirow{2}{*}{\textbf{Height}}& \multicolumn{4}{c}{\textbf{Pitch}} \\
& $\theta=0^{\circ}$ &  $\theta=45^{\circ}$ & $\theta=60^{\circ}$ & $\theta=90^{\circ}$\\
\midrule
\texttt{1} $h=15$m & \textcolor{blue}{52.14} &47.62 &39.04 &29.31\\
\texttt{2} $h=35$m &\textcolor{blue}{58.03} &55.61 &55.07 &45.54\\
\texttt{3} $h=60$m &\textcolor{blue}{53.31} &50.38 &50.23	&46.65\\
\bottomrule
\end{tabular}}

\caption*{(e) \textbf{Height $35$ \& Pitch $0^{\circ}$}}
\label{table:s2s_h35p0}
}
\hfill
\parbox{.2\textwidth}{
\centering
\setlength{\tabcolsep}{2.5pt}
\resizebox{0.25\textwidth}{!}{
\begin{tabular}{lccccc}
\toprule
&\multicolumn{4}{c}{\textbf{Test mIoU ($\uparrow$)}} \\
\multirow{2}{*}{\textbf{Height}}& \multicolumn{4}{c}{\textbf{Pitch}} \\
& $\theta=0^{\circ}$ & $\theta=45^{\circ}$ & $\theta=60^{\circ}$ & $\theta=90^{\circ}$\\
\midrule
\texttt{1} $h=15$m &\textcolor{blue}{48.50} & \textcolor{blue}{$50.71$} & $45.22$ & $42.21$\\
\texttt{2} $h=35$m &$50.49$ & \textcolor{blue}{$55.74$} & \textcolor{blue}{$57.11$} & $52.19$\\
\texttt{3} $h=60$m &$45.33$ & \textcolor{blue}{$49.79$} & \textcolor{blue}{$50.37$} & $44.62$\\
\bottomrule
\end{tabular}}

\caption*{(f) \textbf{Height $35$ \& Pitch $45^{\circ}$}}
\label{table:s2s_h35p45}
}
\hfill
\parbox{.2\textwidth}{
\centering
\setlength{\tabcolsep}{2.5pt}
\resizebox{0.25\textwidth}{!}{
\begin{tabular}{lccccc}
\toprule
&\multicolumn{4}{c}{\textbf{Test mIoU ($\uparrow$)}} \\
\multirow{2}{*}{\textbf{Height}}& \multicolumn{4}{c}{\textbf{Pitch}} \\
& $\theta=0^{\circ}$ & $\theta=45^{\circ}$ & $\theta=60^{\circ}$ & $\theta=90^{\circ}$\\
\midrule
\texttt{1} $h=15$m & 37.21 &\textcolor{blue}{49.04}&\textcolor{blue}{47.42}	&44.93\\
\texttt{2} $h=35$m &34.37 &52.67	&\textcolor{blue}{57.52} &\textcolor{blue}{54.14}\\
\texttt{3} $h=60$m &29.82 &\textcolor{blue}{44.36} &\textcolor{blue}{48.33}	&\textcolor{blue}{44.71}\\
\bottomrule
\end{tabular}}

\caption*{(g) \textbf{Height $35$ \& Pitch $60^{\circ}$}}
\label{table:s2s_h35p60}
}
\hfill
\parbox{.2\textwidth}{
\centering
\setlength{\tabcolsep}{2.5pt}
\resizebox{0.25\textwidth}{!}{
\begin{tabular}{lccccc}
\toprule
&\multicolumn{4}{c}{\textbf{Test mIoU ($\uparrow$)}} \\
\multirow{2}{*}{\textbf{Height}}& \multicolumn{4}{c}{\textbf{Pitch}} \\
& $\theta=0^{\circ}$ & $\theta=45^{\circ}$ & $\theta=60^{\circ}$ & $\theta=90^{\circ}$\\
\midrule
\texttt{1} $h=15$m &28.73 &40.11 &\textcolor{blue}{44.9} &\textcolor{blue}{46}\\
\texttt{2} $h=35$m &25.89 &38.98 &46.26	&\textcolor{blue}{54.02}\\
\texttt{3} $h=60$m &20.36 &29.95 &36.1 &\textcolor{blue}{43.16}\\
\bottomrule
\end{tabular}}

\caption*{(h) \textbf{Height $35$ \& Pitch $90^{\circ}$}}
\label{table:s2s_h35p90}
}
\hfill
\parbox{.2\textwidth}{
\centering
\setlength{\tabcolsep}{2.5pt}
\resizebox{0.25\textwidth}{!}{
\begin{tabular}{lccccc}
\toprule
&\multicolumn{4}{c}{\textbf{Test mIoU ($\uparrow$)}} \\
\multirow{2}{*}{\textbf{Height}}& \multicolumn{4}{c}{\textbf{Pitch}} \\
& $\theta=0^{\circ}$ & $\theta=45^{\circ}$ & $\theta=60^{\circ}$ & $\theta=90^{\circ}$\\
\midrule
\texttt{1} $h=15$m & \textcolor{blue}{37.59} &32.89 &24.53 &18.35\\
\texttt{2} $h=35$m &\textcolor{blue}{48.42} &44.31 &42.82	&32.04\\
\texttt{3} $h=60$m &\textcolor{blue}{51.53} &48.38&	47.71 &39.45\\
\bottomrule
\end{tabular}}

\caption*{(i) \textbf{Height $60$ \& Pitch $0^{\circ}$}}
\label{table:s2s_h60p0}
}
\hfill
\parbox{.2\textwidth}{
\centering
\setlength{\tabcolsep}{2.5pt}
\resizebox{0.25\textwidth}{!}{
\begin{tabular}{lccccc}
\toprule
&\multicolumn{4}{c}{\textbf{Test mIoU ($\uparrow$)}} \\
\multirow{2}{*}{\textbf{Height}}& \multicolumn{4}{c}{\textbf{Pitch}} \\
& $\theta=0^{\circ}$ & $\theta=45^{\circ}$ & $\theta=60^{\circ}$ & $\theta=90^{\circ}$\\
\midrule
\texttt{1} $h=15$m &\textcolor{blue}{43.01} &\textcolor{blue}{43.33}	&35.6 &31.05\\
\texttt{2} $h=35$m & 52.13 &\textcolor{blue}{57.6} &	\textcolor{blue}{58.84} &51.95\\
\texttt{3} $h=60$m &51.89 &\textcolor{blue}{56.83}&	\textcolor{blue}{56.43} &50.07\\
\bottomrule
\end{tabular}}

\caption*{(j) \textbf{Height $60$ \& Pitch $45^{\circ}$}}
\label{table:s2s_h60p45}
}
\hfill
\parbox{.2\textwidth}{
\centering
\setlength{\tabcolsep}{2.5pt}
\resizebox{0.25\textwidth}{!}{
\begin{tabular}{lccccc}
\toprule
&\multicolumn{4}{c}{\textbf{Test mIoU ($\uparrow$)}} \\
\multirow{2}{*}{\textbf{Height}}& \multicolumn{4}{c}{\textbf{Pitch}} \\
& $\theta=0^{\circ}$ & $\theta=45^{\circ}$ & $\theta=60^{\circ}$ & $\theta=90^{\circ}$\\
\midrule
\texttt{1} $h=15$m &38.58 &\textcolor{blue}{43.4} &36.96	&32.22\\
\texttt{2} $h=35$m &44.69 &\textcolor{blue}{57.52}	&\textcolor{blue}{60.21} &54.3\\
\texttt{3} $h=60$m &43.24 &\textcolor{blue}{56.68} &\textcolor{blue}{59.27}	&49.18\\
\bottomrule
\end{tabular}}

\caption*{(k) \textbf{Height $60$ \& Pitch $60^{\circ}$}}
\label{table:s2s_h60p60}
}
\hfill
\parbox{.2\textwidth}{
\centering
\setlength{\tabcolsep}{2.5pt}
\resizebox{0.25\textwidth}{!}{
\begin{tabular}{lccccc}
\toprule
&\multicolumn{4}{c}{\textbf{Test mIoU ($\uparrow$)}} \\
\multirow{2}{*}{\textbf{Height}}& \multicolumn{4}{c}{\textbf{Pitch}} \\
& $\theta=0^{\circ}$ & $\theta=45^{\circ}$ & $\theta=60^{\circ}$ & $\theta=90^{\circ}$\\
\midrule
\texttt{1} $h=15$m &28.12 &\textcolor{blue}{34.9} &\textcolor{blue}{33.6} &\textcolor{blue}{33.28}\\
\texttt{2} $h=35$m &32.76 &45.32	&\textcolor{blue}{53.71} &\textcolor{blue}{55.27}\\
\texttt{3} $h=60$m &29.58 &42.59	&50.73	&\textcolor{blue}{57.07}\\
\bottomrule
\end{tabular}}

\caption*{(l) \textbf{Height $60$ \& Pitch $90^{\circ}$}}
\label{table:s2s_h60p90}
}
\caption{\footnotesize\textbf{Model Sensitivity to changing Height and Pitch.} We evaluate a model trained on one $h,\theta$ variation (indicated by sub-table caption) across all other  $h,\theta$ variations. Performant conditions are highlighted in \textcolor{blue}{blue}.}
\label{table:hp_all}
\end{table*}
\label{sec:diagnostic_exp}
\par\noindent
Similar to Table 6 (d) in the main paper, in Table~\ref{table:hp_all}, we assess broader $(h, \theta)$ sensitivity of models by training DAFormer~\cite{hoyer2022daformer} models across all (total $12$) $(h, \theta)$ settings and evaluate them across the same conditions in \csim.Table ~\ref{table:hp_all} (a) models trained on $(h=15m, \theta=0^{\circ})$ are representative of one extreme of the range (both lowest height and pitch values) -- we notice that this model is \textit{extremely sensitive} to height variations.Tables~\ref{table:hp_all} (a), (e) and (i),
we can deduce that due to the high variability in perspective between $\theta=0^{\circ}$ and other $\theta\neq0^{\circ}$ conditions, models trained on $\theta=0^{\circ}$ do not generalize well to other $\theta$ values. On the other hand, Tables.~\ref{table:hp_all} (b), (c), (d), (f), (g) and (h) models trained on oblique perspectives are better at generalizing to other pitch conditions.






\begin{figure}[htbp]
    \begin{minipage}[b]{\textwidth}
    \centering
    \begin{tabular}{lcccccc}
        \toprule
        \textbf{Daytime}& \multicolumn{5}{c}{\textbf{Test mIoU ($\uparrow$)}} \\ 
        \textbf{Conditions} & \textbf{Vanilla} & \multicolumn{2}{c}{\textbf{Motion Blur}} & \multicolumn{2}{c}{\textbf{Spatter}} \\
        \cmidrule(lr){2-2} \cmidrule(lr){3-4} \cmidrule(lr){5-6}
        & & \textbf{sev. = 3} & \textbf{sev. = 5} & \textbf{sev. = 3} & \textbf{sev. = 5} \\
        \midrule
        Noon & 83.54 & 74.31 (9.23$\downarrow$) & 67.49 (16.05$\downarrow$) & 78.42 (5.11$\downarrow$) & 63.03 (20.51$\downarrow$) \\
        Sunset & 84.32 & 74.62 (9.70$\downarrow$) & 68.08 (16.24$\downarrow$) & 78.10 (6.22$\downarrow$) & 68.62 (15.69$\downarrow$) \\
        \bottomrule
    \end{tabular}

        \captionsetup[sub]{labelformat=parens} 
        \subcaption{\textbf{Degraded Performance of Rein models under Visual Corruptions.}}
        \label{tab:miou_scores}
    \end{minipage}%
    \hfill
    \begin{minipage}[b]{\textwidth}
        \centering
    \includegraphics[width=\linewidth]{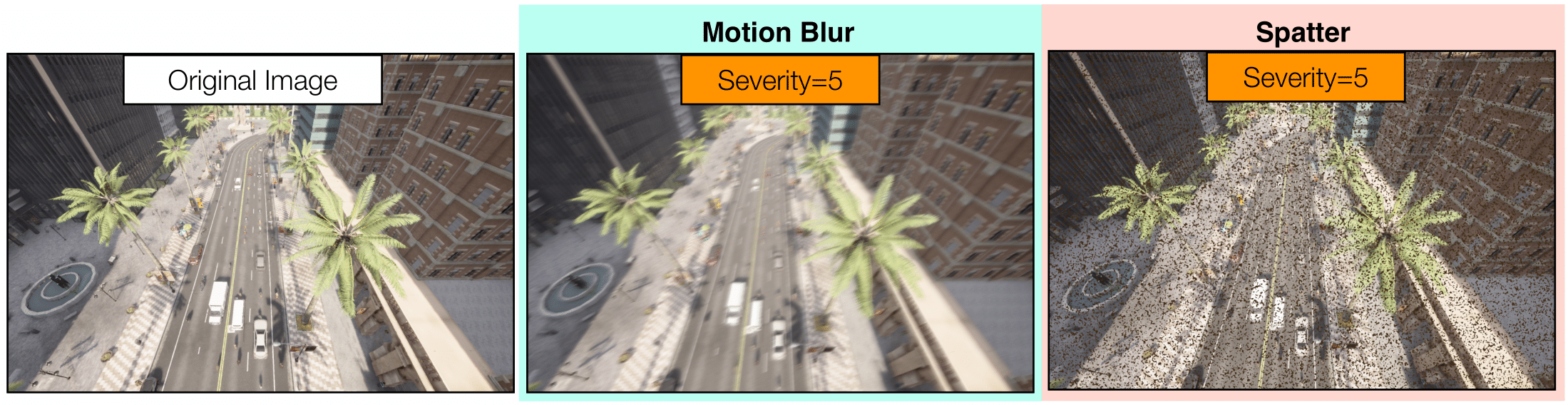}
        \captionsetup[sub]{labelformat=parens} 
        \subcaption{\textbf{Simulating Visual Artifacts.}}
        \label{fig:corruption}
    \end{minipage}
    \label{fig:corruption_all}
\end{figure}

\subsection{Visual Artifacts}
Real-world images can have sensory artifacts beyond 
weather/daytime variations (UAV123 \cite{10.1007/978-3-319-46448-0_27}, VIRAT \cite{virat}, UG$^{2}$ \cite{ug2}). Prior work (RobustNav \cite{2021RobustNav}) shows it's possible to simulate such corruptions with varying severities. Fig.\ref{fig:corruption} provides examples of motion blur and spatter, highlighting how models trained on clean data degrade under these conditions (see Table.~\ref{tab:miou_scores}).

\subsection{Auto labeling for Real Datasets}
\begin{figure}[ht!]
    \centering
    \includegraphics[width=\linewidth]{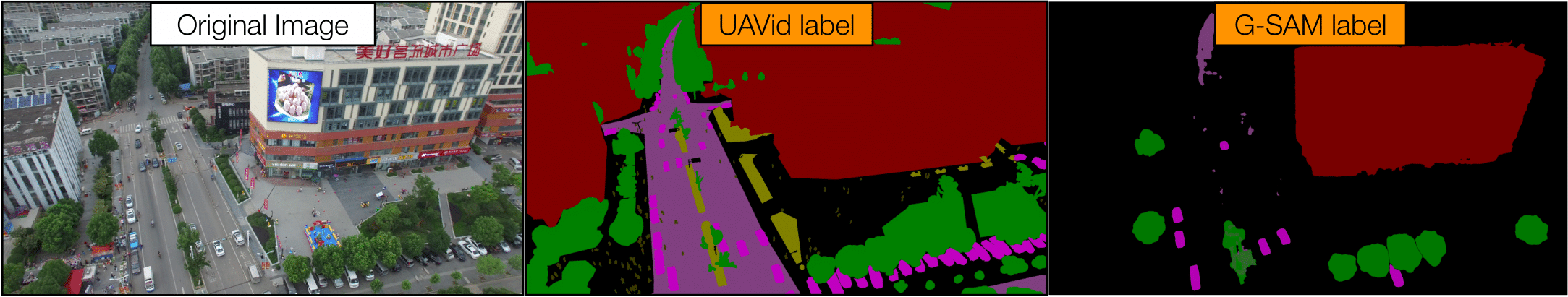}
    \caption{\textbf{Grounded-SAM predictions on a UAVid image.}}
    \label{fig:gsam}
\end{figure}
\par\noindent
Fig.\ref{fig:gsam} shows a sample semantic segmentation prediction of a random \ua image using Grounded-SAM \cite{gsam}. 
We find that most of the roads are absent from the prediction, no humans are predicted, and only a few instances of vehicles, buildings, and trees are identified correctly. 
While auto-labeling using foundation models has potential, we believe there is still room for improvement in specific applications (such as aerial).
\begin{figure*}[!htbp]
    \centering
    \includegraphics[width=\linewidth]{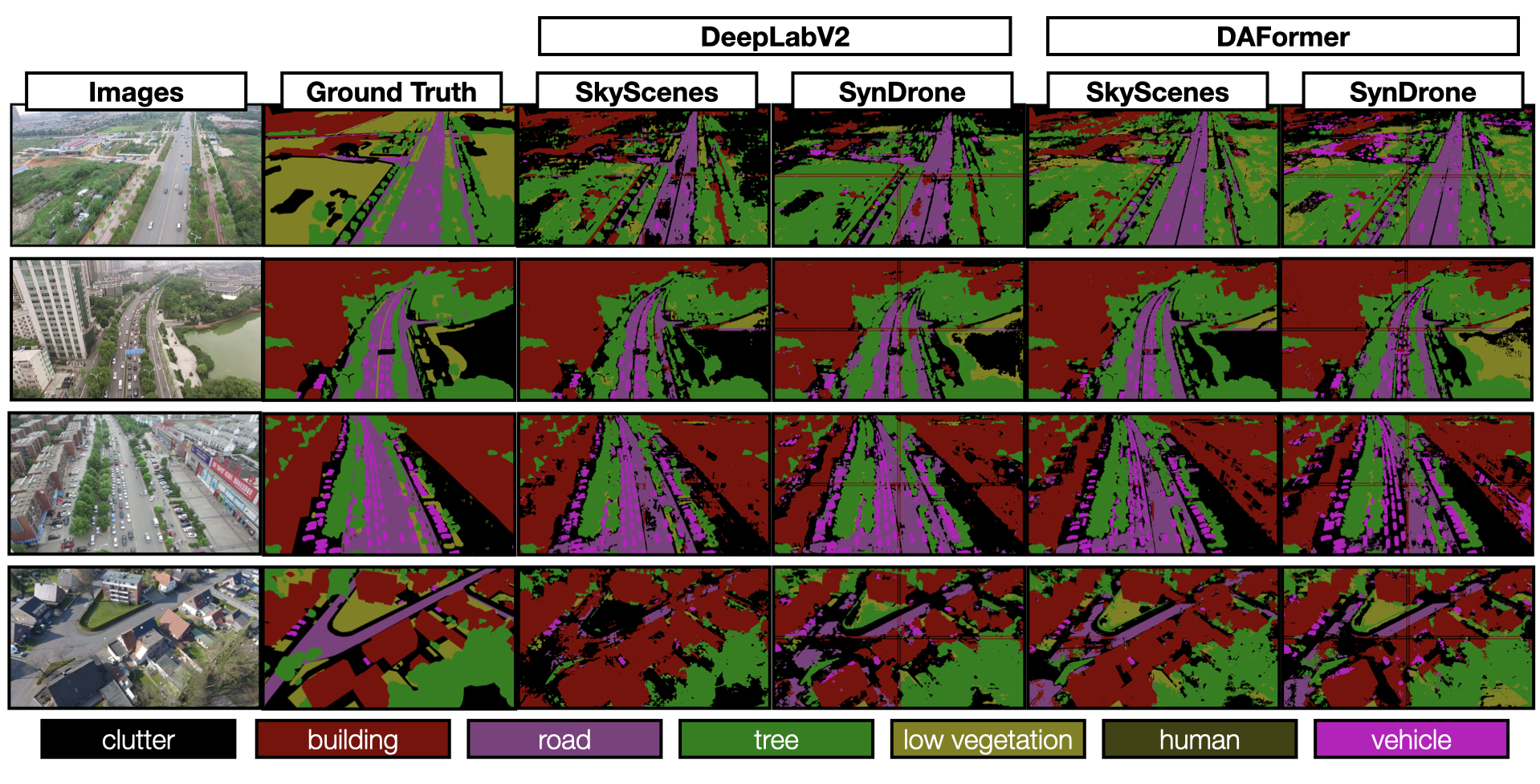}
\captionof{figure}{
\footnotesize\textbf{Synthetic $\rightarrow$\ua Semantic Segmentation Predictions} Out-of-the-box semantic segmentation predictions made on randomly selected \ua~\cite{lyu2020uavid} validation
images by models trained on \csim and \syn. The first two columns indicate the original image and the associated ground truth respectively, columns $3$ (\csim) and $4$ (\syn) indicate predictions by DeepLabv2~\cite{chen2017deeplab} models and columns $5$ (\csim) and $6$ (\syn) indicate predictions made by DAFormer~\cite{hoyer2022daformer} models. 
}
\label{fig:uavid}
\end{figure*}

\begin{figure*}[!htbp]
    \centering
    \includegraphics[width=\linewidth]{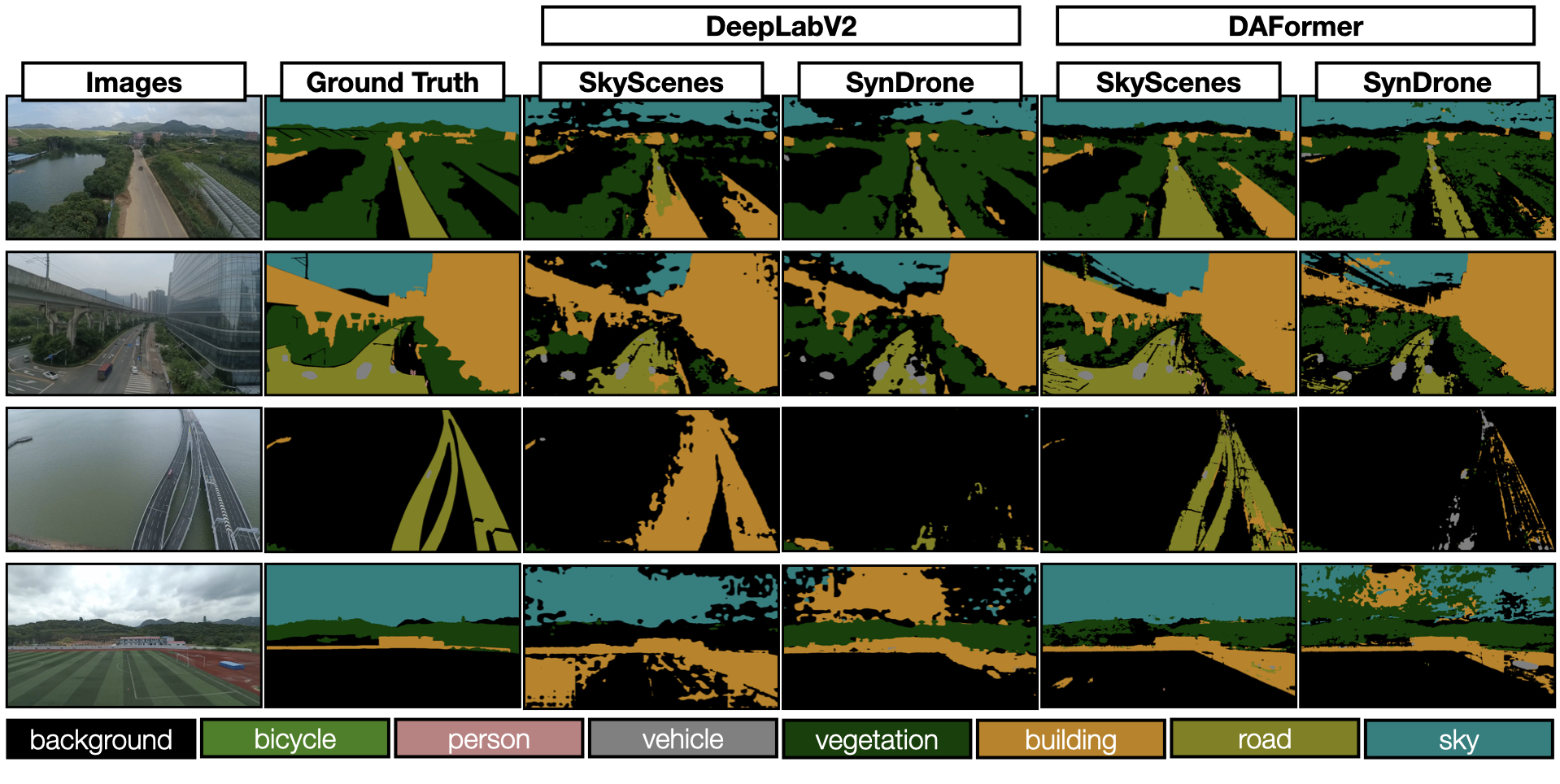}
    \captionof{figure}{
\footnotesize\textbf{Synthetic $\rightarrow$\aero Out of the box semantic Segmentation Predictions}
Out-of-the-box semantic segmentation predictions made on randomly selected \aero~\cite{aeroscapes} validation
images by models trained on \csim and \syn. The first two columns indicate the original image and the associated ground truth respectively, columns $3$ (\csim) and $4$ (\syn) indicate predictions by DeepLabv2~\cite{chen2017deeplab} models and columns $5$ (\csim) and $6$ (\syn) indicate predictions made by DAFormer~\cite{hoyer2022daformer} models. 
}
\label{fig:aero}
\end{figure*}

\begin{figure*}[!htbp]
    \centering
    \includegraphics[width=\linewidth]{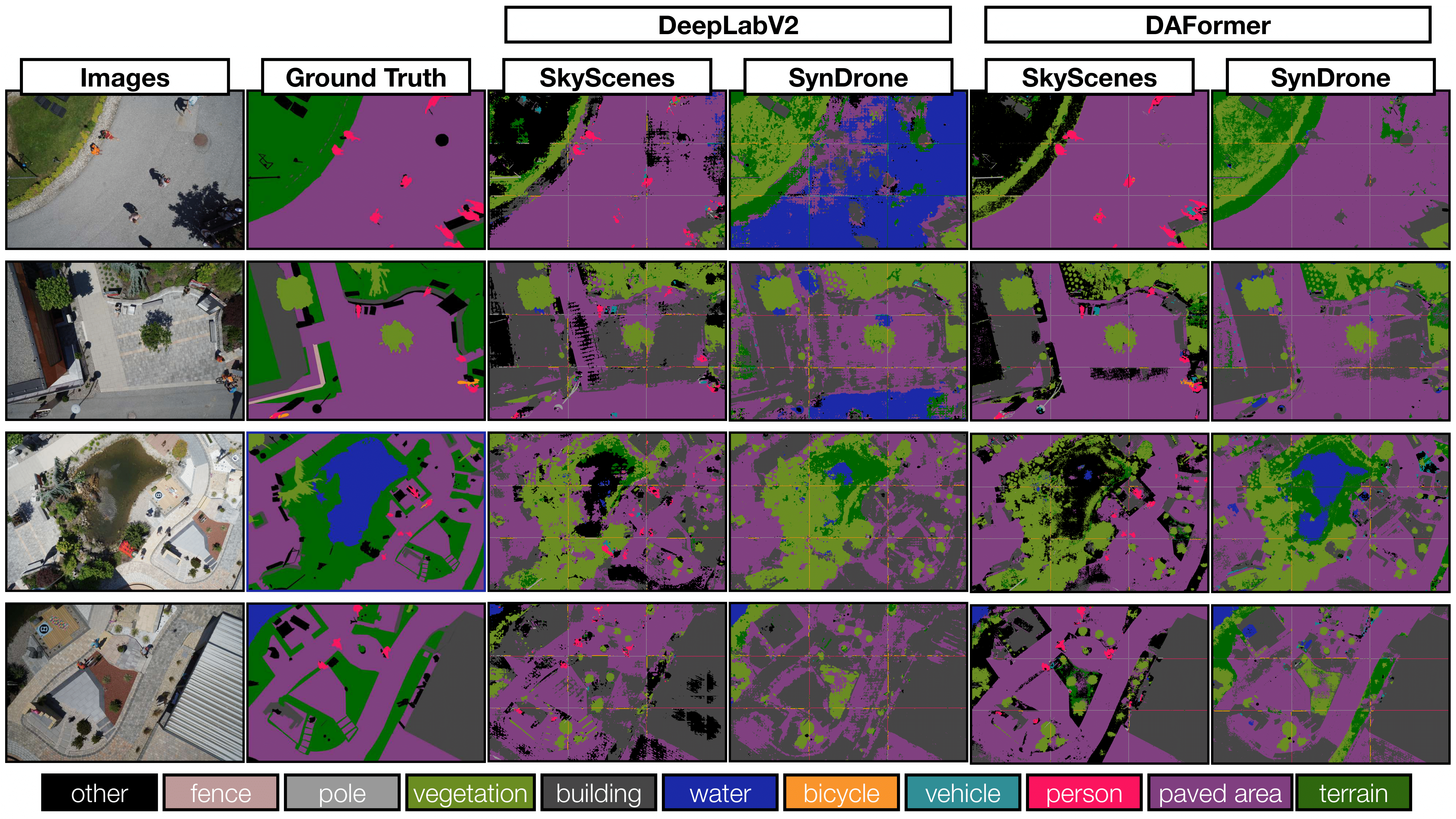}
\captionof{figure}{
\footnotesize\textbf{Synthetic $\rightarrow$\icg Out of the box semantic Segmentation Predictions}  
Out-of-the-box semantic segmentation predictions made on randomly selected \icg~\cite{drone-dataset} validation
images by models trained on \csim and \syn. The first two columns indicate the original image and the associated ground truth respectively, columns $3$ (\csim) and $4$ (\syn) indicate predictions by DeepLabv2~\cite{chen2017deeplab} models and columns $5$ (\csim) and $6$ (\syn) indicate predictions made by DAFormer~\cite{hoyer2022daformer} models. 
}
\label{fig:icg}
\end{figure*}

\begin{figure*}[!htbp]
    \centering
    \includegraphics[width=\linewidth]{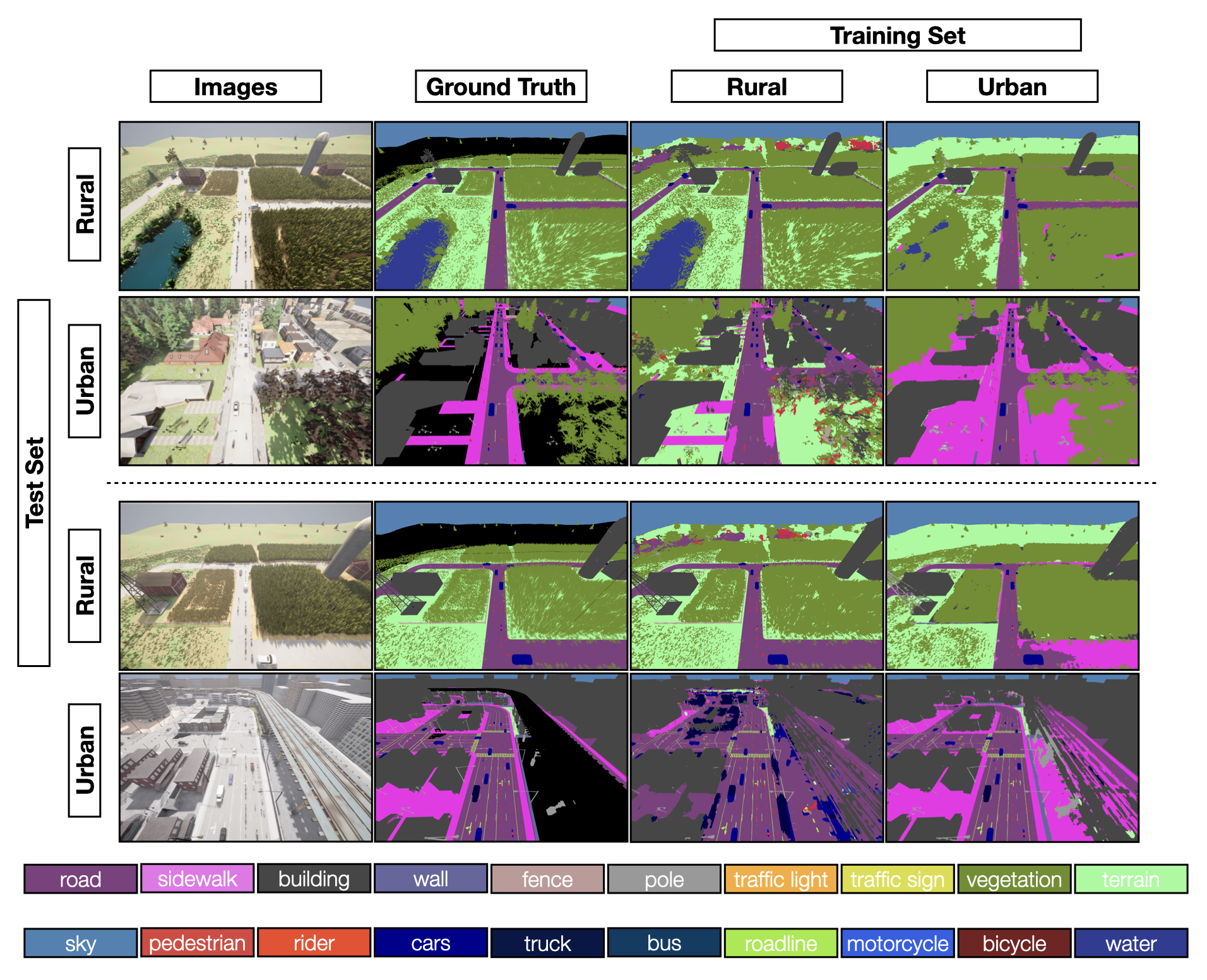}
\captionof{figure}{
\footnotesize\textbf{Semantic Segmentation predictions across rural and urban town variations} Semantic segmentation predictions made on 
on held-out \csim images
across rural and urban scenes by a DAFormer~\cite{hoyer2022daformer} model trained on rural and urban scenes respectively. The first two columns indicate the original image and the associated ground truth, column 3 is predictions by model trained on rural scenes subset, column 4 is predictions by models trained on urban scenes subset.
}
\label{fig:layout_qual}
\end{figure*}

\begin{figure}
\centering

\includegraphics[width=\linewidth]{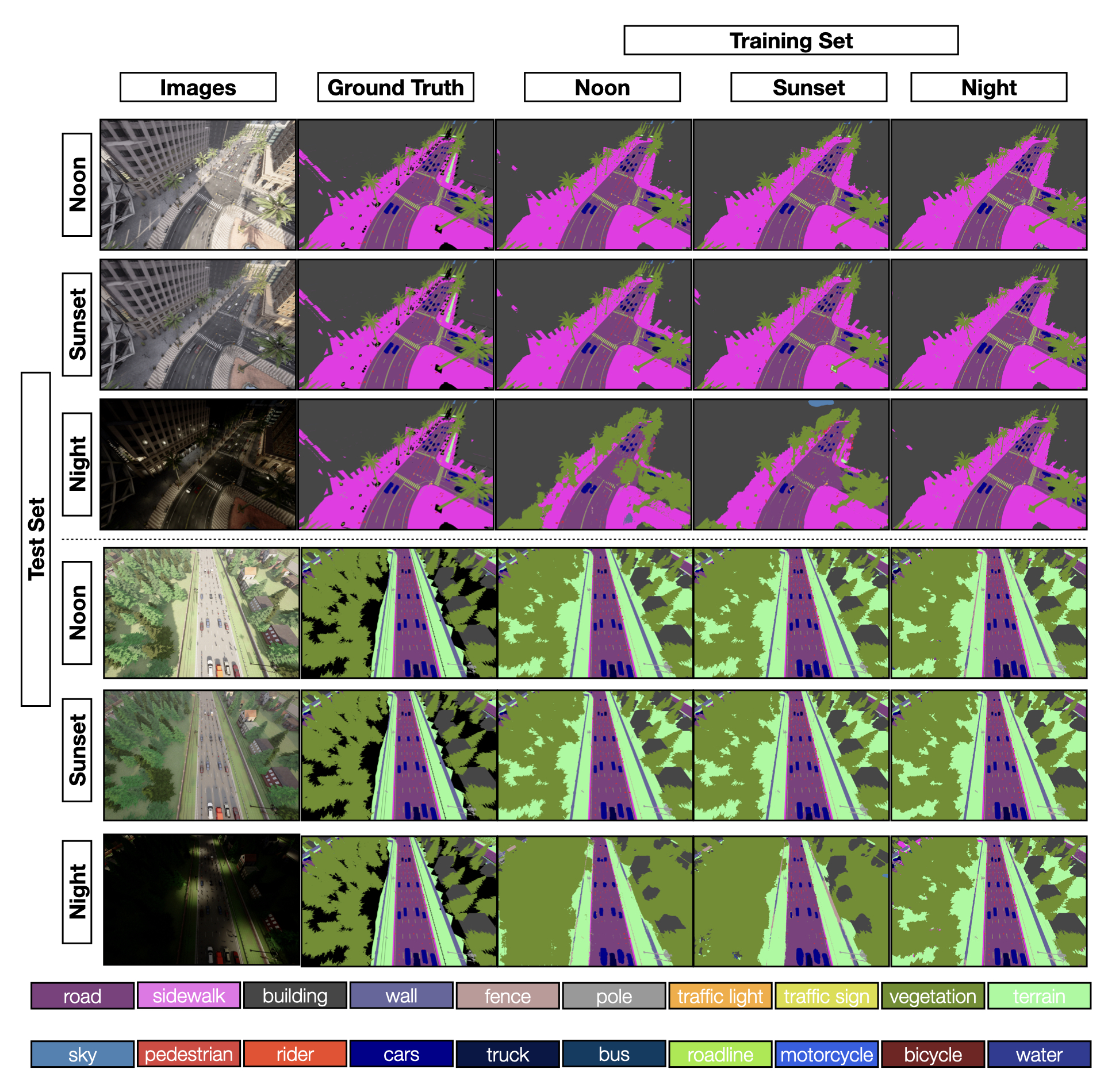}

\captionof{figure}{
\footnotesize\textbf{Semantic Segmentation predictions across different daytime variations} Semantic segmentation predictions made on 
on held-out \csim images
across all daytime variations by a DAFormer~\cite{hoyer2022daformer} model trained on select daytime variations. The first two columns indicate the original image and the associated ground truth, column 3 is predictions by model trained on Noon subset, column 4 is predictions by models trained on Sunset subset and column 5 is predictions by model trained on Night subset. 
}
\label{fig:daytime_qual}
\end{figure}



\subsection{Qualitative examples}
In Fig.~\ref{fig:uavid},~\ref{fig:aero} and~\ref{fig:icg}, we show qualitative examples of predictions made by \csim and \syn trained DeepLabv2 and DAFormer models on \ua, \aero, and \icg respectively. In Fig.~\ref{fig:daytime_qual} and~\ref{fig:layout_qual}, we show how predictions are impacted by changing \csim conditions.

\section{Limitations}

Depth of distant objects in \csim is imperfect, akin to real-world conditions. As evident from Fig. 5 in the paper, depth observations struggle to discern finer details of high-altitude viewpoints. We hypothesize that improvement from depth is helpful only for objects close to the camera or lower altitude viewpoints. 

\section{Future Work}
\label{sec:future work}
\begin{figure}
    \centering
    \includegraphics[width=\linewidth]{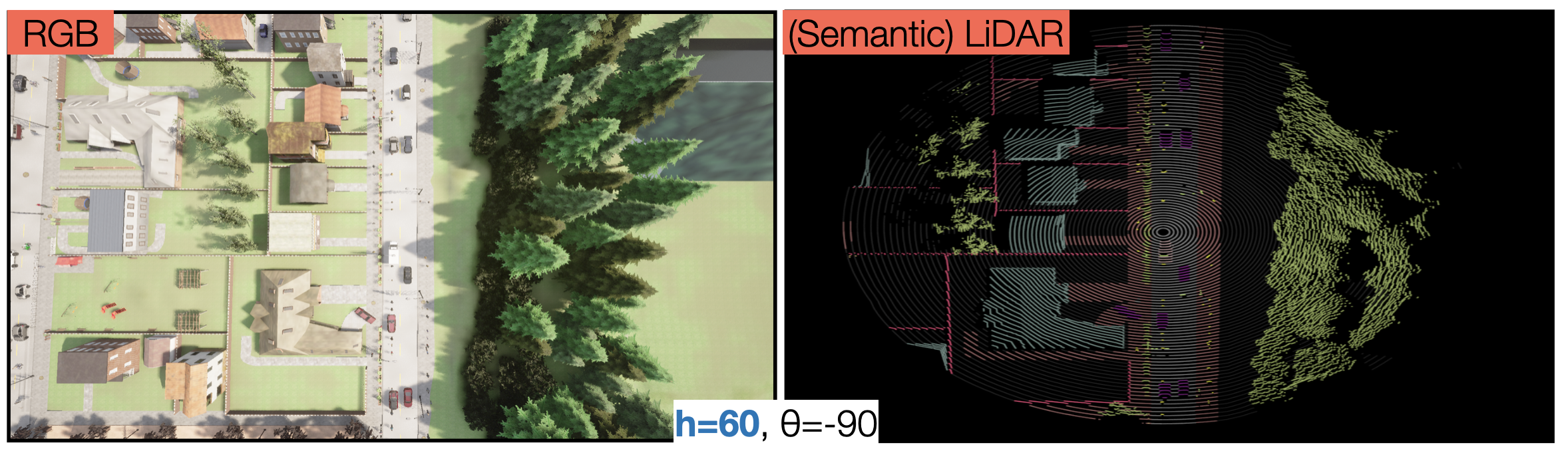}
    \caption{\textbf{Demonstrating support for 3D perception tasks in \csim in future iterations} \csim RGB and corresponding Semantic LiDAR image generated for ($h =60m$, $\theta = 90^{\circ}$) and ClearNoon setting.}
    \label{fig:liadar}

\end{figure}
\par\noindent
We plan on updating \csim with evolving considerations for real-world aerial scene-understanding -- improved realism, additional anticipated edge cases -- as more and more features are supported in the underlying simulator and provide additional support for 3D perception tasks (see Fig.~\ref{fig:liadar}.)

\end{document}